\documentclass{article}
\usepackage{hyperref}
\usepackage{url}
\usepackage{titletoc}
\usepackage{todonotes}
\usepackage{subcaption}
\usepackage{fullpage}

\usepackage{natbib}
\usepackage{graphicx}
\usepackage{wrapfig}

\usepackage{notation}

\usepackage{authblk}

\newcommand{\arxivversion}{}

\author[1]{Yuxin Ma}
\author[2]{Adir Dayan}
\author[2]{Yam Eitan}
\author[2,3]{Haggai Maron}
\author[1]{Soledad Villar}

\affil[1]{Department of Applied Mathematics and Statistics, Johns Hopkins University}
\affil[2]{Technion}
\affil[3]{NVIDIA}

\title{Transferable Graph Metanetworks}
\date{}

\begin{document}

\maketitle

\begin{abstract}
A \emph{weight space network} (or \emph{metanetwork}) takes the weights of another neural network as input and predicts properties of it.
Most prior work trains such models on input networks of one or a few fixed sizes and evaluates them in-distribution.
The few attempts at out-of-distribution size generalization remain limited in scope and have achieved only modest success.
Consequently, the potential efficiency gains of training on small networks and evaluating on much larger ones remain largely unrealized.
We propose \emph{Transferable Graph Metanetworks}, which extend the graph metanetwork paradigm with a set of modifications that make performance transferable across input networks of different \emph{widths}.
The modifications follow two principles: invariance to the ways in which networks of different widths represent the same function, and continuity, such that weights representing similar functions receive similar predictions.
We further study whether size generalization is possible for input networks trained independently from random initialization.
Empirically, our modifications significantly improve size generalization on every task we consider. 
Performance is strongest on input networks trained under the maximal-update parameterization ($\mu$P), where it remains robust up to $42\times$ the training width.
Theoretically, we explain these observations with infinite-width limit theory: we prove size-generalization guarantees for our model on $\mu$P-trained inputs, and explain why it can fail under other parameterizations.
\end{abstract}
\newpage
\setcounter{tocdepth}{2}
\tableofcontents
\newpage

\section{Introduction}
\label{sec:intro}
The parameters of a trained neural network are increasingly treated as a data modality in their own right \citep{han2026survey}.
Such data is plentiful and informative: public model repositories such as Hugging Face hold millions of checkpoints, and each one determines the function its network represents while carrying traces of its task, its learned representations, and the training process that produced it.
What makes this data hard to work with is that it arrives as a long list of numbers in which no individual coordinate means anything on its own and in which many hidden symmetries relate different parameter settings to the same underlying function.
As a result, the map from those numbers to the quantities one cares about is neither explicit nor simple.
The approach that has emerged is to learn that map from data by feeding the parameters as input to a second neural network that exploits the underlying symmetries of the parameter space.
This second network is trained either to predict the property of interest or to output new weights or weight updates.
Such a model is called a \emph{weight space network}, or \emph{metanetwork}.
This recipe supports a wide range of applications.
For example, metanetworks can predict accuracy or the generalization gap from the weights without ever evaluating the network on data, making model selection and hyperparameter search cheaper~\citep{unterthiner2020predicting,schurholt2021self,mehta2024improving}.
Another use case is the implicit neural representation (INR): each data point \emph{is} a small network representing an image, a shape, or a scene, so that classifying or editing the data becomes an operation on parameters~\citep{dupont2022data,deluigi2023deep,navon2023equivariant}.

We ask a question that prior work largely leaves open:
\begin{center}
Can we construct metanetworks that generalize across input networks of different sizes, and in particular that are trained on small networks and perform well on large ones?
\end{center}

For graph neural networks (GNNs), there is a vast literature on size-generalization \cite{yehudai2021local, bevilacqua2021size} and transferability \cite{ruiz2020graphon, levie2024graphon, maskey2023transferability, keriven2020convergence, velasco2024graph}, but those results are not directly applicable to the metanetwork setting due to the particular structure and symmetries of their inputs. 
Size generalization is useful in this context for two related reasons.
First, real model populations are heterogeneous, spanning many widths and depths~\citep{touvron2023llama}, but checkpoints of the largest models are the rarest, precisely because they are the most expensive to train.
Second, the cost of training a metanetwork grows with the size of its input networks.
A metanetwork that trains on the small networks that are cheap and plentiful, and then applies to larger sizes it has never seen, avoids both obstacles.

Many existing metanetworks based on equivariant layers~\citep{navon2023equivariant,navon2023align,zhou2023permutation,zhou2023neural,zhou2024universal,tran2024monomial} are designed specifically for a fixed weight space (on specific architecture with fixed width and depth).
Graph metanetworks (GMNs) that construct graphs out of input networks and apply GNNs or attention~\citep{lim2023graph,kofinas2024graph,kalogeropoulos2024scale,bar2026graph} do accept input networks of arbitrary size, and some are tested on networks a small factor wider than those seen in training, with only limited success reported~\citep{lim2023graph,kofinas2024graph}.
Indeed, size generalization was not a design goal for these models, and nothing in their architecture constrains their predictions at unseen sizes.

As our \textbf{first contribution} (\Cref{sec:transferable_gmn}), motivated by the theory of any-dimensional learning and transferability~\citep{levin2025transferring}, we propose the framework of \emph{Transferable Graph Metanetworks}: a set of modifications that make the performance of an existing GMN transferable across input networks of different \emph{widths} (leaving depth to future work).
The modifications come in two levels, and both are generic enough to apply to any standard GMN variants taking MLP and CNN input networks.
All claimed properties are proved in \Cref{sec:transferable_gmn_proofs} and verified numerically in \Cref{app:verify-compatibility}.

\begin{figure}[tbp]
\centering
\newlength{\duplicationpanelheight}
\newlength{\blockwisepanelheight}
\newlength{\panellabelheight}
\setlength{\duplicationpanelheight}{0.2\textwidth}
\setlength{\blockwisepanelheight}{0.2\textwidth}
\setlength{\panellabelheight}{1.2\baselineskip}
\newcommand{\panellabel}[1]{%
\par\vspace{2pt}%
\parbox[b][\panellabelheight][t]{\linewidth}{\centering\scriptsize #1}}
\begin{subfigure}[b]{0.495\textwidth}
\centering
\includegraphics[trim=12pt 12pt 12pt 12pt, clip,width=\linewidth]{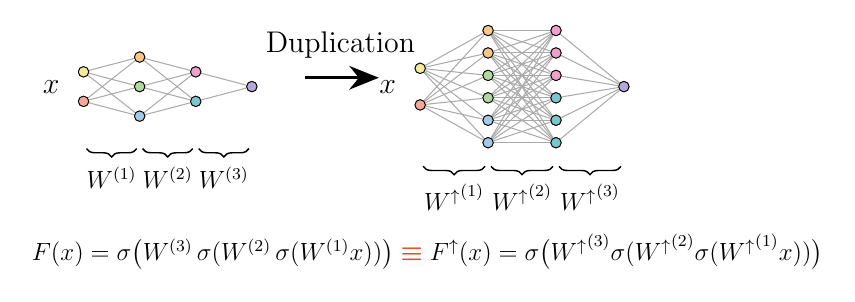}
\panellabel{(a) Duplicating hidden neurons leaves the function unchanged}
\end{subfigure}\hspace{4pt}%
\begin{subfigure}[b]{0.26\textwidth}
\centering
\includegraphics[trim=12pt 12pt 12pt 12pt, clip,height=\duplicationpanelheight]{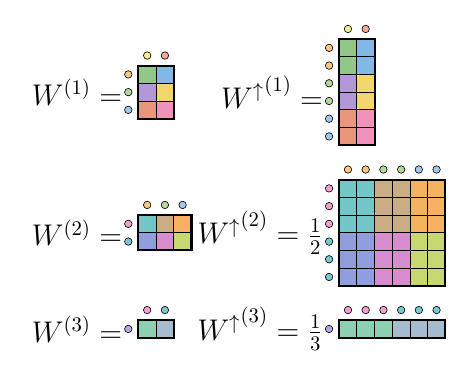}
\panellabel{(b) Duplication widening \eqref{eq:dup}}
\end{subfigure}\hspace{4pt}%
\begin{subfigure}[b]{0.22\textwidth}
\centering
\includegraphics[trim=12pt 12pt 12pt 12pt, clip,height=\blockwisepanelheight]{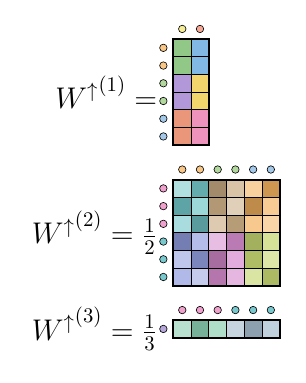}
\panellabel{(c) Block-wise widening \eqref{eq:blockwise}}
\end{subfigure}
\captionsetup{font=small,skip=2pt}
\caption{Function-preserving widenings of an MLP.
(b) Copying and rescaling weights duplicates hidden neurons, as in (a), so the function is unchanged; duplication-compatible GMN is invariant to this transformation.
(c) More generally, the copied weights can be redistributed across rows of the block, which still amounts to the duplication of neurons in (a) and preserves the function; matrix-product GMN is invariant to this transformation.}
\label{fig:widening_illustration}
\end{figure}
\begin{itemize}[leftmargin=*,topsep=0pt,itemsep=0pt]
\item \textbf{Level 1}.
Networks of different widths can represent the same function, and a metanetwork should be invariant to the transformations that relate them.
We begin with the simplest such transformation, copying and rescaling the weights accordingly (see Figure \ref{fig:widening_illustration}(b)), and show that three simple modifications already make a GMN exactly invariant to it: fan-in rescaling of the input weights, mean aggregation, and a layer-wise mean readout.
We call the result a \emph{duplication-compatible GMN}. This modification does not affect the expressivity at a fixed width.
\item \textbf{Level 2}.
The GMNs from Level 1 have two shortcomings.
First, they are not invariant to the more general transformations that preserve the represented function (see Figure \ref{fig:widening_illustration}(c)).
Second, they are not continuous, which may affect generalization behavior: a small perturbation of the input weights that barely changes the represented function can change the metanetwork's output substantially.
We modify the GMNs to address both issues and call the result a \emph{matrix-product GMN}.
This modification might restrict expressivity, but we show that it improves size generalization.
\end{itemize}

Our \textbf{second contribution} (\Cref{sec:experiments}) is an empirical study of whether our proposed modifications improve size generalization in practical settings: can the GMNs generalize to input networks of a much larger width than the one it was trained on?
In practice, the input networks of different widths are trained independently from random initializations, so how their training dynamics relate to each other across widths depends on both the \emph{optimization} (the algorithm and its hyperparameters) and the \emph{parameterization} (how the initialization and those optimization hyperparameters scale with width).
We fix the optimizer and vary only the parameterization, considering two choices: the standard PyTorch default, called standard parameterization (SP), and maximal-update parameterization ($\mu$P)~\citep{yang2020feature}, a popular choice that was shown to align training dynamics across widths and to enable zero-shot hyperparameter transfer~\citep{yang2021tuning} (see~\Cref{fig:mup-hp-transfer}).
Our experiments show that (i) both Level 1 and Level 2 modifications significantly improve size generalization in all cases, under both parameterizations, (ii) size generalization is much better on $\mu$P-trained input networks than on SP-trained ones, and (iii) the combination of the Level 2 matrix-product GMN with $\mu$P-trained input performs well across the full range of test widths.

Our \textbf{third contribution} (\Cref{sec:sizegen}) is a theoretical analysis that explains these empirical findings.
Using results from the infinite-width limit literature~\citep{yang2023tensorprogramsivb} to characterize the distribution of trained weights, we prove size generalization guarantees for the matrix-product GMN on $\mu$P-trained inputs, formalizing observation (iii), and identify the obstructions that explain why size generalization may otherwise fail.
$\mu$P constrains no individual width; it is a rule for how the initialization and optimization hyperparameters change with width. So our guarantee places no restriction on the input networks used for training, but instead specifies the wider networks it provably generalizes to: those trained with the same optimizer under the hyperparameters this scaling prescribes.
Some such specification is unavoidable, since if the wide networks' training dynamics bear no relation to the narrow ones' then nothing learned at the training width can transfer, whatever the architecture; $\mu$P makes the relation precise.
Beyond this regime, we still expect our modifications to help,albeit with weaker performance and no theoretical guarantee, as the SP experiments indicate.

Our theory focuses on MLP input networks, but the modification recipe itself is architecture-agnostic, and our experiments apply it to both MLP and CNN input networks.
We expect the theory to extend beyond MLPs and leave this to future work.

Our code is available at \url{https://github.com/yuxinma98/transferable-graph-metanetworks}.

\section{Preliminaries and related work}
\label{sec:related}

\subsection{Weight-space symmetries, metanetworks, and graph metanetworks}\label{sec:related-weightsym}
Most architectures are overparameterized with many different weights computing the same function.
Whenever the prediction target depends only on the \emph{function} the input network expresses, the metanetwork should be invariant to the transformations that leave the represented function unchanged.
State-of-the-art metanetworks are designed to exploit this.
The simplest such transformation is \textbf{permutation symmetry}: permuting the corresponding rows and columns of two adjacent weight matrices permutes the hidden neurons between them and leaves the function unchanged.
Another is \textbf{scale symmetry}: scaling a hidden neuron's incoming weights and bias while inversely scaling its outgoing weights leaves the function unchanged, with the admissible rescalings depending on the activation.
Both symmetries act within a single weight space, at one fixed width.

There are three main approaches to designing metanetworks that respect these symmetries.
The first builds equivariant layers with the standard toolkit of equivariant machine learning, as in DWSNets~\citep{navon2023equivariant}, NFNs~\citep{zhou2023permutation}, and NFTs~\citep{zhou2023neural}, which impose permutation symmetry, and Monomial-NFNs~\citep{tran2024monomial} and MAGEP-NFN~\citep{vo2024equivariant}, which further impose scale symmetry.
Such metanetworks are defined only for a fixed input width and depth, which prevents their application to networks of varying size.
The second strategy views the input network as a graph, with neurons as nodes and weights as edges, and applies a message-passing GNN or transformer to it, as in GMNs~\citep{lim2023graph} and neural graphs~\citep{kofinas2024graph}.
ScaleGMN \citep{kalogeropoulos2024scale} extends GMNs to also impose scale symmetry.
These models naturally apply to input networks of varying size, just as GNNs and transformers apply to graphs of varying size.
A third strategy avoids operating on the full parameter space and instead represents networks through probes, gradients, or other derived quantities~\citep{herrmann2024learning,kahana2025deep,heo2026linear,gelberg2026gradmetanet}, which reduces the need to model weight-space symmetries directly but requires suitable probe data and may miss information not exposed by the chosen inputs or derived signals.
For a comprehensive review of weight-space symmetries and metanetworks, see \citet{han2026survey}.

In this work, we focus on the second approach.
Here we concretely describe a simple GMN on MLP input networks, so that we can specify the proposed modifications in~\Cref{sec:transferable_gmn}.
\footnote{The extensions to other GMN variants from the literature are described in \Cref{app:gmns-variants}.
They involve some technicalities, but the underlying idea is the same throughout.}

We first fix the notation for MLP input networks, that we will use throughout the paper.
An $L$-layer MLP parameterizes a function from $\RR^{\din}$ to $\RR^{\dout}$, mapping $a^{(0)}=u$ to $x^{(L+1)}$ by recursing
\begin{equation}
\label{eq:mlp}
x^{(\ell)} = \Wl a^{(\ell-1)} + b^{(\ell)} \in \RR^{n_\ell},
\qquad
a^{(\ell)} = \sigma\bigl(x^{(\ell)}\bigr) \in \RR^{n_\ell},
\qquad \ell=1,\dots,L+1,
\end{equation}
where $n_0=\din$ and $n_{L+1}=\dout$, $(\Wl\in\RR^{n_\ell\times n_{\ell-1}})_{\ell=1}^{L+1}$ and $(b^{(\ell)}\in\RR^{n_\ell})_{\ell=1}^{L+1}$ are the layer-wise weights and biases, and $\sigma$ is the activation function applied elementwise.
We write $\theta=\bigl((\Wl,b^{(\ell)})\bigr)_{\ell=1}^{L+1}$ for the weights.

A GMN acting on an MLP input network is a message-passing GNN on the MLP's computation graph: given the weights $\theta$, the graph has one node $v^{(\ell)}_i$ per neuron ($\ell=0,\dots,L+1$, $i\in[n_\ell]$) with initial feature $h^{(\ell),[0]}_i=\psi_v\bigl(b^{(\ell)}_i\bigr)$ for $\ell\ge1$ and a constant for $\ell=0$, and one directed edge $v^{(\ell-1)}_j\to v^{(\ell)}_i$ with initial feature $e^{(\ell),[0]}_{ji}=\psi_e\bigl(W^{(\ell)}_{ij}\bigr)$.
The model then runs $T$ rounds of message passing: for $t=0,\dots,T-1$, and $\ell=0,\dots, L+1$,
\begin{equation}
\label{eq:gmn}
\begin{aligned}
m^{(\ell),[t]}_{j\to i}
&=\psi_m\bigl(e^{(\ell),[t]}_{ji},h^{(\ell-1),[t]}_j\bigr),
\qquad
m^{(\ell),[t]}_{i\leftarrow j}
=\psi_m^{\leftarrow}\bigl(e^{(\ell+1),[t]}_{ij},h^{(\ell+1),[t]}_j\bigr),
\\
h^{(\ell),[t+1]}_i
&=\psi_h\bigl(h^{(\ell),[t]}_i,\sum_{j\in[n_{\ell-1}]} m^{(\ell),[t]}_{j\to i}, \sum_{j\in[n_{\ell+1}]} m^{(\ell),[t]}_{i\leftarrow j}\bigr),\\
e^{(\ell),[t+1]}_{ji}
&=\psi_u\bigl(e^{(\ell),[t]}_{ji},h^{(\ell-1),[t]}_j,h^{(\ell),[t]}_i\bigr).
\end{aligned}
\end{equation}
The first aggregate is set to $0$ at $\ell=0$, and the second is set to $0$ at $\ell=L+1$.
Finally, the GMN reads out a graph-level prediction by pooling the final node features,
\begin{equation}
\label{eq:gmn-readout}
f^\Theta(\theta)
=\rho\Bigl(\sum_{\ell=0}^{L+1}\sum_{i\in[n_\ell]}\psi_r\bigl(h^{(\ell),[T]}_i\bigr)\Bigr).
\end{equation}
Each node $v^{(\ell)}_i$ aggregates a forward message $m^{(\ell),[t]}_{j\to i}$ along incoming edges and a backward message $m^{(\ell),[t]}_{i\leftarrow j}$ along outgoing ones.
We call this the \emph{bidirectional} GMN, and the restriction $\psi_m^{\leftarrow}=0$, in which messages pass only along edge direction, gives the \emph{forward} GMN.
Components $\psi_v,\psi_e,\psi_m,\psi_m^{\leftarrow},\psi_h,\psi_u,\psi_r,\rho$ are all shared MLPs, with parameters collected in $\Theta$. Since $\Theta$ is independent of the width $(n_\ell)_{\ell=0}^{L+1}$, the GMN applies to MLPs of any width.


\subsection{Any-dimensional learning, transferability, and size generalization}\label{sec:related-anydim}
Our proposal of transferable GMNs is guided by the theory of any-dimensional learning and transferability in \citet{levin2025transferring}.
This theory studies models that accept inputs of any size, such as DeepSets, GNNs, and transformers.
It asks when such a model keeps working well at sizes larger than those it was trained on.
The theory identifies two properties a model can have, each defined relative to some structure on the input space that the target function itself respects.
The first, \emph{compatibility}, says the model must give the same output on two inputs that are equivalent, in the sense that the target function itself gives the same output on both.
The second, \emph{continuity}, says the model should vary continuously with respect to a prescribed metric that compares inputs across sizes, a metric under which the target itself varies continuously.
Continuity is the stronger property: compatibility only transfers a prediction to larger inputs exactly equivalent to a training input.
Continuity transfers it to every larger input merely \emph{close} to one---the situation that arises in practice.

These are formalized through \emph{consistent sequences}, which we briefly review since our later analysis relies on this formulation.
Consider a family of vector spaces $V_n$, where $V_n$ collects the objects of size $n$, together with linear embeddings $\varphi_{N,n}: V_n\hookrightarrow V_N$ for $n\leq N$ that send a size-$n$ object to its size-$N$ counterpart and thereby declare which objects of different sizes are equivalent.\footnote{Each $V_n$ also carries an action of a group $\msf G_n$ where $\msf G_n \hookrightarrow G_N$, and the embedding is group equivariant. For weight space applications, this models the permutation symmetries that prior works consider.}
Identifying each object with its larger counterparts gives the direct limit $\Vinf\colonequals\varinjlim_n V_n=\bigcup_n V_n$, a single space holding all sizes at once. 
Moreover, if each $V_n$ carries a norm that respects the equivalence across sizes, that is, $\norm[V_N]{\varphi_{N,n}(x)}=\norm[V_n]{x}$ for $x\in V_n$, these norms glue into a norm $\norm[\Vinf]{\cdot}$ on $\Vinf$ measuring distances across sizes.
A sequence of functions $f_n\colon V_n\to\RR$ is \emph{compatible} if $f_N|_{V_n}=f_n$ for $n\leq N$, equivalently if the $f_n$ glue into a single $f_\infty\colon\Vinf\to\RR$. The sequence $f_n$ is \emph{continuous} (or transferable) if it is compatible and $f_\infty$ is \emph{continuous} in $\norm[\Vinf]{\cdot}$.
This formalizes the compatibility and continuity of models and targets mentioned above.

Weight spaces of varying width fit this template.
Here $V_n$ is the weight space of MLPs of width $n$: two weights are equivalent (respectively, close) when they represent the same (respectively, similar) functions.
A GMN is an any-dimensional neural network accepting weights of different widths.
In \Cref{sec:transferable_gmn}, we make the equivalence and the norm on weight space precise, and design a GMN that is compatible and continuous with respect to them.
We give the full consistent-sequence formulation of the weight space in \Cref{sec:consistent-sequence-formulation}.

\subsection{$\mu$P, infinite-width limits, and tensor programs}
The relationship between the distribution of trained weights of different widths depends on the choice of parameterization (initialization and hyperparameter scalings with width).
Table~\ref{tab:width-scalings} summarizes two popular choices: the standard PyTorch default (SP) and $\mu$P.
We emphasize that this choice affects only the \emph{relationship} between widths, not the behavior at any single fixed width: each width-scaling in Table~\ref{tab:width-scalings} carries a free multiplicative constant, which can always be chosen to make SP and $\mu$P agree at one chosen width.

The distributions of trained weights in the infinite-width limit, that is, as all hidden widths tend to infinity, are well studied.
In our theoretical analysis in~\Cref{sec:sizegen}, we use the tensor-programs machinery of \citet{yang2023tensorprogramsivb}, which treats the i.i.d.\ random initialization as the sole source of randomness, entering through the law of large numbers and the central limit theorem, to rigorously characterize the limiting distribution of activations and outputs at each training step.
This analysis is what led to the derivation of $\mu$P~\citep{yang2020feature}, the unique parameterization achieving ``optimal'' feature-learning behavior.
Under $\mu$P, networks of every width approximate this optimal infinite-width limit with error $O(1/\sqrt n)$, which is why training dynamics look so similar across widths in practice.
Separately, $\mu$P has become a popular choice because it enables zero-shot hyperparameter transfer across widths~\citep{yang2021tuning}: hyperparameters tuned on a small network can be used directly on a larger one, since their optimal values stay close across widths (see~\Cref{fig:mup-hp-transfer}).

\begin{center}
\captionsetup{type=table,hypcap=false, font=small,skip=2pt}
\caption{Initialization variance and example hyperparameters for different MLP parameters under SP and {\color{purple}$\mu$P} trained with SGD and Adam~\citep{yang2021tuning}.
$\mu$P is also defined for any standard architecture and prescribe scalings for weight-decay coefficients. See~\cite{yang2023tensorprogramsivb} for details. In both cases, the output bias $b^{(L+1)}$ is initialized to zero with constant learning rate. For a weight matrix $W$ taking inputs $x$ as $Wx$, fan\_in and fan\_out are the number of columns and rows respectively.}
\label{tab:width-scalings}
\small
\newcommand{\paramcell}[2]{${\color{black}#1}\quad{\color{purple}#2}$}
\resizebox{\textwidth}{!}{%
\begin{tabular}{@{}c c c c@{}}
\toprule
& \shortstack[c]{Init variance}
& \shortstack[c]{SGD learning rate}
& \shortstack[c]{AdamW learning rate}\\
\midrule
Input weights $W^{(1)}$; Biases $(b^{(\ell)})_{\ell=1}^{L}$
& \paramcell{1/\text{fan\_in}}{1/\text{fan\_in}}
& \paramcell{1}{\text{fan\_out}}
& \paramcell{1}{1}\\
Hidden weights $(\Wl)_{\ell=2}^{L}$
& \paramcell{1/\text{fan\_in}}{1/\text{fan\_in}}
& \paramcell{1}{1}
& \paramcell{1}{1/\text{fan\_in}}\\
Output weights $W^{(L+1)}$
& \paramcell{1/\text{fan\_in}}{1/\text{fan\_in}^2}
& \paramcell{1}{1/\text{fan\_in}}
& \paramcell{1}{1/\text{fan\_in}}\\
\bottomrule
\end{tabular}%
}
\end{center}

\section{Transferable Graph Metanetworks}
\label{sec:transferable_gmn}
\subsection{Duplication-compatible Graph Metanetworks}
As mentioned in \Cref{sec:related-weightsym}, a metanetwork should be invariant to transformations of its input networks that leave the represented function unchanged.
Prior work has taken permutation and scale symmetries, both acting within a single width, as its guiding design principle.
To handle input networks of any width, we instead exploit a structure relating \emph{different} widths that prior work has not used: weights of different widths can represent the same function.

\paragraph{Widening by duplicating weight entries.}
The simplest such transformation duplicates and rescales the weight entries (See~\Cref{fig:widening_illustration}(b)).
Recall the MLP weights $\theta=\bigl((\Wl,b^{(\ell)})\bigr)_{\ell=1}^{L+1}$ from~\eqref{eq:mlp}.
Fix width multipliers $k_1,\dots,k_{L}\in\NN$ and set $k_0=k_{L+1}=1$, so that only hidden layers grow, and define $\theta^\uparrow=\bigl((\Wlup,\blup)\bigr)_{\ell=1}^{L+1}$ by
\begin{equation}
\label{eq:dup}
\Wlup \colonequals \Wl\otimes k_{\ell-1}^{-1}\one_{k_\ell}\one_{k_{\ell-1}}^\top,
\qquad
\blup \colonequals b^{(\ell)}\otimes\one_{k_\ell},
\end{equation}
where $\otimes$ is the Kronecker product and $\one_k\in\RR^k$ is the vector of ones.\footnote{In \textsc{PyTorch}: \texttt{W\_up = W.repeat\_interleave(k0, dim=0).repeat\_interleave(k1, dim=1) / k1}, and \texttt{b\_up = b.repeat\_interleave(k0)}, where \texttt{k0} and \texttt{k1} are $k_\ell$ and $k_{\ell-1}$.}
Then $\theta^\uparrow$, of widths $(n_\ell k_\ell)_{\ell=0}^{L+1}$, represents the same function as $\theta$.
We call this the \emph{duplication widening}.
Indeed, by induction on $\ell$ the widened activations are $a^{(\ell)\uparrow}=a^{(\ell)}\otimes\one_{k_\ell}$ (the hidden neurons are duplicated), and at $\ell=L+1$, where $k_{L+1}=1$, the two outputs agree.
This weight-space structure has been studied in prior work on optimization landscapes~\citep{fukumizu2000local,fukumizu2019semi} and on initializing wide networks from narrow ones~\citep{chen2015net2net,ma2026mupscalingsmallmodelsprincipled}, but to our knowledge it has not been exploited in the design of metanetworks.
Also, this is not specific to simple MLPs but generalizes to almost all standard architectures (see Theorem 2.5 of~\cite{ma2026mupscalingsmallmodelsprincipled}).

\paragraph{Duplication-compatible GMNs.}
Following the theoretical framework reviewed in \Cref{sec:related-anydim}, we call a metanetwork \emph{duplication compatible} if it is invariant to this duplication widening, that is, if $f^\Theta(\theta^\uparrow)=f^\Theta(\theta)$ for every duplication widening \eqref{eq:dup} and every GMN parameter $\Theta$.
We show that three simple modifications of the GMN of \Cref{sec:related-weightsym} suffice to enforce it.
\begin{enumerate}[leftmargin=*,topsep=0pt,itemsep=0pt]
\item \emph{Fan-in rescaled edge features}: initialize edge features using the rescaled weights $\Wlbar\colonequals n_{\ell-1}\Wl$ rather than from $\Wl$.
This rescaling cancels the $k_{\ell-1}^{-1}$ factor in the duplication widening \eqref{eq:dup}, so that the features exactly get duplicated when the GMN applies to widened weights.
\item \emph{Mean aggregation}: replace the two sum-aggregations in \eqref{eq:gmn} by mean-aggregations, $\frac{1}{n_{\ell-1}}\sum_{j\in[n_{\ell-1}]} m^{(\ell),[t]}_{j\to i}$ and $\frac{1}{n_{\ell+1}}\sum_{j\in[n_{\ell+1}]} m^{(\ell),[t]}_{i\leftarrow j}$.
This is because summing over a wider neighborhood would scale up the incoming messages by the multiplicity of the duplication, whereas averaging returns the same value as before.
\item \emph{Layer-wise mean readout}: replace the readout \eqref{eq:gmn-readout} by 
\[f^\Theta(\theta)=\rho\bigl(r_1,\dots,r_{L},(h^{(0),[T]}_i)_{i\in[n_0]},(h^{(L+1),[T]}_i)_{i\in[n_{L+1}]}\bigr),\quad r_\ell\colonequals\frac{1}{n_\ell}\sum_{i\in[n_\ell]}\psi_r\bigl(h^{(\ell),[T]}_i\bigr).\]
In the hidden layers, taking layer-wise mean over node features is invariant to duplication.
Input and output node features may be listed coordinatewise because $n_0$ and $n_{L+1}$ do not grow.
\end{enumerate}

\begin{proposition}[Duplication compatibility]
\label{prop:dup-invariance}
The GMN with the three modifications above is duplication compatible.
\end{proposition}

None of the three modifications reduces expressivity at a fixed width.
Rescaling the edge features and averaging instead of summing only rescale intermediate quantities by constants depending on the width, which the learned maps can absorb.
The layer-wise mean readout retains strictly more information than the single global sum it replaces.

\subsection{Matrix-product Graph Metanetworks}
The duplication-compatible GMN has two shortcomings that hinder size generalization.

\paragraph{Problem 1: continuity in the wrong norm.}
Compatibility only ensures generalization to wide weights \emph{exactly} equivalent to a training weight.
As discussed in \Cref{sec:related-anydim}, generalizing to weights that are merely \emph{nearby} additionally requires the metanetwork to be continuous, that is, robust to small perturbations: weights representing nearly the same function should receive nearly the same output.
Which perturbations count as small is not a free choice, it is dictated by the target.

To make this precise, let $V_n\colonequals\bigoplus_{\ell=1}^{L+1}(\RR^{n_\ell\times n_{\ell-1}}\oplus\RR^{n_\ell})$ be the weight space of width $n=(n_1,\dots,n_{L})$, embedded in $V_N$ for $n_\ell\mid N_\ell$ by the duplication widening $\theta\mapsto\theta^\uparrow$~\eqref{eq:dup}.
This makes $(V_n)$ a consistent sequence, in which equivalent weights of different widths represent the same function.
On each $V_n$, we consider two norms: the \emph{normalized spectral norm} $\norm[V_n,\op]{\cdot}$ and the \emph{normalized $\ell_1$ norm} $\norm[V_n,1]{\cdot}$
\begin{equation}
\label{eq:norms}
\norm[V_n,\op]{\theta}
=\sum_{\ell=1}^{L+1}\Bigl(\sqrt{\tfrac{n_{\ell-1}}{n_\ell}}\,\norm{\Wl}+\tfrac{1}{\sqrt{n_\ell}}\norm[2]{b^{(\ell)}}\Bigr),
\quad
\norm[V_n,1]{\theta}
=\sum_{\ell=1}^{L+1}\Bigl(\tfrac{\norm[1]{\Wlbar}}{n_\ell n_{\ell-1}}+\tfrac{\norm[1]{b^{(\ell)}}}{n_\ell}\Bigr),
\end{equation}
where $\norm{\cdot}$ is the spectral norm, $\norm[1]{\cdot}$ the entrywise $\ell_1$ norm, and $\norm[2]{\cdot}$ the Euclidean norm.
Both norms satisfy $\norm[V_n]{\theta}=\norm[V_N]{\theta^\uparrow}$ for every duplication widening \eqref{eq:dup}, so they are compatible with the equivalence across widths.
Hence, they induce norms $\norm[V_\infty,\op]{\cdot}$ and $\norm[V_\infty,1]{\cdot}$ on the limit space $\Vinf$, defining two different notions of distance between weights of different widths.
\Cref{sec:consistent-sequence-formulation} carries this construction out in full and explains what each norm measures.

The target and the duplication-compatible GMN turn out to respect different ones of the two.
\begin{proposition}[Misalignment]
\label{prop:misalignment}
For the input MLP, let its input space $X\subseteq\RR^{\din}$ be nonempty and compact, and let its activation function $\sigma$ be Lipschitz.
Equip the space of continuous functions $\mc{C}(X,\RR^{\dout})$ with the uniform norm.
\begin{enumerate}[leftmargin=*,topsep=0pt,itemsep=0pt]
\item Let $\mc{R}\colon\Vinf\to\mc{C}(X,\RR^{\dout})$ be the realization map that sends the weights to the function they represent. Then $\mc{R}$ is Lipschitz on every bounded ball of $\norm[V_\infty,\op]{\cdot}$ with a width-independent constant. However, if $\sigma$ is unbounded, then $\mc{R}$ can be discontinuous in $\norm[V_\infty,1]{\cdot}$.
\item The duplication-compatible GMN with Lipschitz MLP components is Lipschitz in $\norm[V_\infty,1]{\cdot}$. 
However, if the input MLP has more than one layer, there exist duplication-compatible GMNs that are discontinuous in $\norm[V_\infty,\op]{\cdot}$.
\end{enumerate}
\end{proposition}

The implication of this misalignment is the following:
Weights close in $\norm[V_\infty,\op]{\cdot}$ represent similar functions, hence have similar targets, yet the duplication-compatible GMN may map them to very different outputs; weights close in $\norm[V_\infty,1]{\cdot}$ do receive similar outputs from it, but may represent very different functions.
Either way, the architecture itself puts no constraint on its predictions at unseen widths.
Such misalignment was observed empirically to hinder size generalization across a range of tasks for any-dimensional models~\citep{levin2025transferring}, so we expect the same to happen here.

\paragraph{Problem 2: invariance to the simplest widening only.}
The duplication widening~\eqref{eq:dup} is not the only function-preserving widening.
Partition $\Wlup$ into blocks $\Wlup_{ij}\in\RR^{k_\ell\times k_{\ell-1}}$ for $i\in [n_\ell], j\in [n_{\ell-1}]$, and ask only that these blocks satisfy, for each $\ell$,
\begin{equation}
\label{eq:blockwise}
\Wlup_{ij}\one_{k_{\ell-1}}=W^{(\ell)}_{ij}\one_{k_\ell}
\quad
\blup = b^{(\ell)}\otimes\one_{k_\ell}.
\end{equation}
In words, every row of the block $\Wlup_{ij}$ sums to $W^{(\ell)}_{ij}$: the copied weights can be redistributed across rows of the block, as illustrated in Figure~\ref{fig:widening_illustration}(c).
Then $\theta^\uparrow$ again represents the same function as $\theta$. 
We call these \emph{block-wise widenings}, and the previous duplication widening~\eqref{eq:dup} is the special case in which every block is constant.
Indeed, each copy of neuron $i$ still collects the whole of the original weight from the copies of neuron $j$, but is free to collect it from them in any proportion.
\Cref{sec:duplication-equivalence} derives~\eqref{eq:blockwise} as exactly the class of widenings whose hidden neurons are copies of the narrow network's (Figure~\ref{fig:widening_illustration}(a)), so it is the most general of its kind, and \Cref{sec:blockwise-compat} proves that the duplication-compatible GMN is not invariant to it.

\paragraph{Matrix-product GMNs.}
Both shortcomings share one cause: the hidden weights are processed entrywise by $\psi_m,\psi_m^{\leftarrow}$ in \eqref{eq:gmn}.
Removing this freedom remedies both at once.
We modify the duplication-compatible GMN to define the \emph{bidirectional matrix-product GMN}: drop all edge features beyond $\Wlbar_{ij}$, and let each weight act on the node feature by plain multiplication:
\begin{equation}\label{eq:matprod}
m^{(\ell),[t]}_{j\to i}=\Wlbar_{ij}\,A^{\mathrm{in}}_\ell h^{(\ell-1),[t]}_j,
\qquad
m^{(\ell),[t]}_{i\leftarrow j}=\overline W^{(\ell+1)}_{ji}\,A^{\mathrm{out}}_\ell h^{(\ell+1),[t]}_j,
\end{equation}
for learnable matrices $A^{\mathrm{in}}_\ell,A^{\mathrm{out}}_\ell\in\RR^{d_h\times d_h}$.
The \emph{forward matrix-product GMN} is the case $A^{\mathrm{out}}_\ell=0$.
Equivalently, stacking the layer-$\ell$ node features as the rows of $H^{(\ell),[t]}\in\RR^{n_\ell\times d_h}$, the mean aggregates of \eqref{eq:matprod} are $\Wl H^{(\ell-1),[t]}A^{\mathrm{in}}_\ell$ and $\frac{n_\ell}{n_{\ell+1}}(W^{(\ell+1)})^\top H^{(\ell+1),[t]}A^{\mathrm{out}}_\ell$: the weights $(\Wl)$ now enter only via matrix multiplication.
This one change resolves both shortcomings: matrix multiplication is exactly the operation controlled by the spectral norm, resolving Problem~1's misalignment, and exactly the operation left invariant by block-wise widenings~\eqref{eq:blockwise}, resolving Problem~2.

\begin{proposition}
\label{prop:matprod}
\begin{enumerate}[leftmargin=*,topsep=0pt,itemsep=0pt]
\item The matrix-product GMN with Lipschitz MLP components is Lipschitz on every bounded ball of $\norm[V_\infty,\op]{\cdot}$ with a width-independent constant.
\item The forward matrix-product GMN satisfies $f^\Theta(\theta^\uparrow)=f^\Theta(\theta)$ for every block-wise widening \eqref{eq:blockwise}.
The bidirectional one satisfies it for those block-wise widenings whose blocks additionally have equal column sums, $\one_{k_\ell}^\top\Wlup_{ij}=\frac{k_\ell}{k_{\ell-1}}W^{(\ell)}_{ij}\one_{k_{\ell-1}}^\top$ for all $i,j,\ell$.
\end{enumerate}
\end{proposition}

The matrix-product GMN is a strict restriction of the duplication-compatible GMN. 
It is presumably less expressive at a fixed width, but we do not observe any performance loss in all experiments.

\section{Experiments}
\label{sec:experiments}
\begin{figure}[t]
\begin{subfigure}{\textwidth}
\centering
\includegraphics[width=\textwidth]{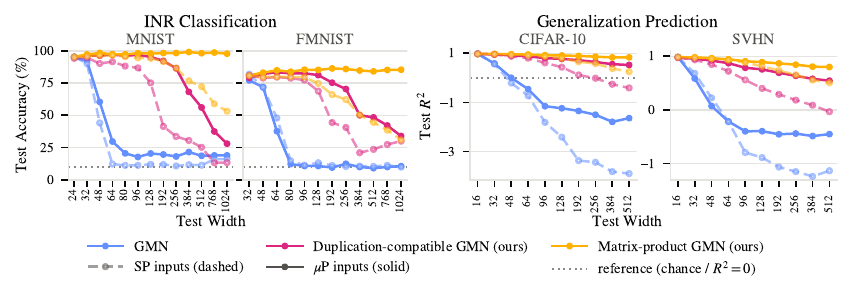}
\caption{GMN (bidirectional).}
\label{fig:sizegen-gmn-bd}
\end{subfigure}
\\[2pt]
\begin{subfigure}{\textwidth}
\centering
\includegraphics[width=\textwidth]{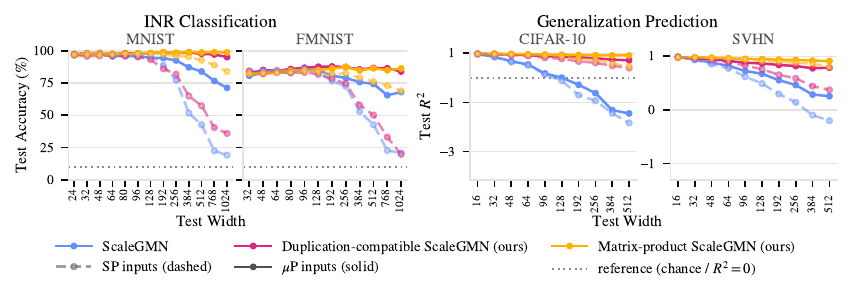}
\caption{ScaleGMN (forward).}
\label{fig:sizegen-scalegmn-fw}
\end{subfigure}
\captionsetup*{font=small}
\caption{Size generalization with independently trained weights, comparing the baseline GMN, the duplication-compatible GMN, and the matrix-product GMN on SP- and $\mu$P-trained input networks, for two baseline architectures.
Each panel trains at the smallest width and evaluates at the sequence of larger widths shown.
Columns 1--2: test accuracy on the INR classification tasks, with the random-guessing baseline.
Columns 3--4: test $R^2$ against ground-truth accuracies on the CNN generalization prediction task, with $R^2=0$ (mean-accuracy prediction) as baseline.}
\label{fig:sizegen}
\end{figure}
We empirically study the size generalization of our transferable GMNs, comparing three models: a baseline GMN with positional encoding and residual connections, using the implementation from~\citep{kalogeropoulos2024scale}; our duplication-compatible GMN; and our matrix-product GMN, the latter two obtained from that baseline via the proposed modifications.
We report two baseline architectures: the bidirectional GMN, and the forward ScaleGMN, which is additionally equivariant to the scaling symmetries of the input network~\citep{kalogeropoulos2024scale}.
The two remaining architectures, the forward GMN and the bidirectional ScaleGMN, together with all other experiment details, are deferred to~\Cref{appen:experiments}.

We experiment on two tasks, each with two datasets of different difficulty, all adapted from~\citep{kalogeropoulos2024scale} and used repeatedly in prior work~\citep{zhou2023permutation, zhou2024universal, unterthiner2020predicting, lim2023graph, kofinas2024graph}:
(1) \emph{INR classification}: given the weights of an MLP fitted to an MNIST~\citep{lecun1998gradient} or FMNIST~\citep{xiao2017fashion} image, predict the image class and report accuracy; and (2) \emph{generalization prediction}: given the weights of a CNN trained on CIFAR-10~\citep{krizhevsky2009learning} or SVHN~\citep{Netzer2011}, predict its test accuracy and report the $R^2$ score.
We further report other metrics, such as Kendall's $\tau$ correlation~\citep{kendall1938new}, in~\Cref{appen:experiments}.
In all experiments the GMN is trained at one small width, with hyperparameters selected on the validation metric at that width only, and evaluated at a range of larger widths, constructed in two ways that we discuss separately below.

\paragraph{Size generalization on hand-crafted widenings.}
As a sanity check, we widen the training-set networks by hand via the duplication widening~\eqref{eq:dup} and the block-wise widening~\eqref{eq:blockwise}, both of which preserve the represented function exactly.
This is a controlled setting, since our modified GMNs are designed to capture these equivalences and the baseline GMN is not.
The results match the theory exactly and are given in \Cref{app:hand-crafted-widenings}.

\paragraph{Size generalization on independently trained weights.}
The more realistic scenario is when the weights at each width are trained independently from random initialization, where it is unclear a priori whether they are even approximately a block-wise widening~\eqref{eq:blockwise} of the training-width networks, and hence whether our modifications help at all.
\Cref{fig:sizegen} shows the results, and \Cref{fig:sizegen-appendix} those for the other baseline GMN models.
The observations below are consistent across all tasks, datasets, and GMN models:
\begin{enumerate}[leftmargin=*,topsep=0pt,itemsep=0pt]
\item On every dataset, performance improves in the order GMN, duplication-compatible GMN, matrix-product GMN, confirming that both modifications help, and echoing the examples of~\cite{levin2025transferring} in which both compatibility and continuity matter.
\item For every model, size generalization is much better on $\mu$P-trained inputs than on SP-trained ones, even though $\mu$P is calibrated to coincide with SP at the training width, so the two share the same training distribution.
Intuitively, $\mu$P-trained networks have similar training dynamics across widths, so small widths inform predictions at large ones; under SP the trained weights vary more, making size generalization inherently harder.
\item Combining the matrix-product GMN with $\mu$P inputs gives an almost flat test score across all larger widths, which we justify theoretically next.
\end{enumerate}

\section{Theory of size generalization on trained weights}
\label{sec:sizegen}

In \Cref{fig:sizegen}, the test score stays nearly constant across widths only for the matrix-product GMN on $\mu$P-trained inputs, degrading visibly as soon as either ingredient is removed.
We now prove size-generalization guarantees that explain this.

We first specify the data distribution: one law $\mu_n$ per width $n=(n_\ell)_{\ell=1}^L$, all produced by a single training rule, with $\mu_n$ the law of the trained weights $\theta_n=\mathsf{Train}_n(\Gamma,S,\xi_n)\in V_n$.
Here $\mathsf{Train}_n$ is the training procedure encoding the optimization algorithm and the parameterization, $\Gamma$ collects the hyperparameters, $\xi_n$ is the random initialization, and $S$ is the remaining randomness coming from mini-batching, etc.
We assume $\Gamma,S,\xi_n$ are mutually independent random variables, and take $\Gamma$ and $S$ to be independent of width, so $\mu_n$ depends on the width only through $\mathsf{Train}_n$ and $\xi_n$.

Now fix a GMN $f^\Theta$, a target function $f^*$, and a loss $\ell$, and write $\mc{R}_m(f^\Theta)=\Ex_{\theta\sim\mu_m}\ell(f^\Theta(\theta),f^*(\theta))$ for the expected loss at width $m$.
If $f^\Theta$ is trained at width $n$ and tested on all widths $m\ge n$, its \emph{size-generalization gap} $\sup_{m\geq n}\abs{\mc{R}_n(f^\Theta) - \mc{R}_m(f^\Theta)}$ is the worst discrepancy between its in-distribution and out-of-distribution test error, exactly what the flatness of a curve in \Cref{fig:sizegen} shows.
We say that $f^\Theta$ \emph{size generalizes} if the gap vanishes as $n\to\infty$.
We will give two guarantees below.
Both control this gap through the Tensor Programs framework of~\citet{yang2023tensorprogramsivb}, which rigorously characterizes the infinite-width limit of the training dynamics, and hence of $\mu_n$.
We state them informally here, with full details given in \Cref{appen:size-generalization-theory}.

The first guarantee covers 1-layer input MLPs only, but comes with a clean geometric interpretation.
We have shown in~\Cref{prop:matprod}(i) that the matrix-product GMN parameterizes a function on $\Vinf$ that is continuous with respect to $d_\infty^{\op}$, the distance induced by the normalized spectral norm $\norm[\Vinf,\op]{\cdot}$~\eqref{eq:norms} after quotienting by hidden-unit permutations.
So if the trained weights $\theta_n$ of different widths are close in $d_\infty^{\op}$ with high probability, performance transfers across widths: intuitively, the model generalizes to larger inputs that are ``similar'' to the training-width inputs, with $d_\infty^{\op}$ capturing the notion of similarity that the GMN respects.
This is exactly what $\mu$P delivers: throughout training, the weights attached to the $n$ hidden neurons behave like $n$ i.i.d.\ draws from a single width-independent law, forcing the trained weights of different widths to be close in $d_\infty^{\op}$.

\begin{theorem}[Informal]\label{thm:guarantee-1}
Let $\mu_n$ be the law of the trained weights of a 1-layer MLP under $\mu$P.
Then $\sup_{m\ge n}W_1(\mu_n,\mu_m;d_\infty^{\op})\to0$ as $n\to\infty$, where $W_1$ is the Wasserstein-1 distance with respect to $d_\infty^{\op}$.
Consequently, if the target is a Lipschitz functional of the represented function and the loss is Lipschitz, then $\sup_{m\ge n}\abs{\mc R_n(f^\Theta)-\mc R_m(f^\Theta)}\to0$ for every duplication-compatible GMN $f^\Theta$ with Lipschitz components, matrix-product ones included.
\end{theorem}
This covers all duplication-compatible GMNs because for 1-layer MLP $\norm[\Vinf,1]{\cdot}$ is dominated by the $\norm[\Vinf,\op]{\cdot}$ (\Cref{prop:one-layer-op-domination}), so \Cref{prop:misalignment}(ii) already gives Lipschitzness in $d_\infty^{\op}$: for 1-layer inputs there is no misalignment, and the weaker model suffices.
The theorem also explains why SP fails: it initializes the weights so that $\norm[V_n,\op]{\theta_n}=\Theta(\sqrt n)$, which already at initialization forces $W_1(\mu_n,\mu_{2n};d_\infty^{\op})\to\infty$.
The weights drift apart as width increases, rather than settling to a common limit, leaving the GMN's predictions at unseen widths unconstrained.

Beyond one hidden layer, we give a second guarantee that works at any depth.
Here we instead compare the laws of the trained weights of different widths, $\mu_n$ and $\mu_m$, through the integral probability metric generated by the GMN losses themselves, $d_{\mc A}(\mu_n,\mu_m)\colonequals\sup_{\Theta\in\mc A}\abs{\mc R_n(f^\Theta)-\mc R_m(f^\Theta)}$ over a compact set $\mc A$ of GMN parameters, which dominates the gap by construction.
This lets us bypass the limitation of the $d_\infty^{\op}$-distance, which does not fully capture the notion of similarity that the GMN respects at depth $L>1$.
However, this guarantee makes a stronger assumption on the target: we state it here for targets exactly representable by the GMN, but it also holds under less restrictive assumptions, as discussed in \Cref{app:guarantee-2}.

\begin{theorem}[Informal]
\label{thm:guarantee-2}
Let $\mu_n$ be the law of the trained weights of a fixed-depth MLP under $\mu$P, and let $f^\Theta$ be the matrix-product GMN.
Suppose the loss is Lipschitz and the target is representable by such a GMN.
Then $\sup_{m\ge n}d_{\mc A}(\mu_n,\mu_m)\to0$ as $n\to\infty$, and hence the size-generalization gap vanishes uniformly over $\Theta\in\mc A$.
\end{theorem}
Taken together, the two guarantees capture the interplay of the GMN architecture and the data distribution in size generalization.
Under SP our modifications still help empirically, so we expect the mechanism to reach beyond the regime our theory covers.

\section{Conclusion}
\label{sec:conclusion}
We show theoretically and empirically that strong size generalization in weight space learning is possible, but that it requires the metanetwork architecture to be matched to the distribution of input weights.
On the architectural side, any-dimensional learning and transferability point to two design requirements.
A metanetwork should be \emph{compatible} with function-preserving widenings, so that functionally equivalent weights at different widths receive the same prediction.
It should also be \emph{continuous} in a metric that reflects functional similarity.
GMNs are a natural fit for these requirements because simple modifications of their message-passing structure can enforce compatibility and continuity without sacrificing much expressivity or scalability.
On the data side, we seek to understand which independently trained, wider weights are sufficiently ``similar'' to the trained narrower weights, so that what was learned on the narrower weights can transfer meaningfully.
Our answer is: wider weights trained with the same optimizer on the same data for the same number of steps, with their initialization and hyperparameters related to those of the narrower weights as prescribed by $\mu$P.
This is intuitive because, under $\mu$P, weights of different widths have very similar training dynamics: they are close to the same nontrivial limit, with $O(n^{-1/2})$ finite approximation error.
We make this intuition rigorous.

Several directions remain open.
Our modifications and analysis address only the \emph{width} of the input networks.
Depth generalization remains untouched: although function-preserving transformations that change depth do exist, it is unclear whether comparably simple GMN modifications can make a metanetwork invariant to them.
Our theory also covers only MLP input networks.
For some of our results, extensions to standard architectures may be accessible through the tensor-programs machinery of \citet{yang2023tensorprogramsivb}.
Finally, it would be valuable to develop an expressivity theory for metanetworks that accept inputs of arbitrary width, complementing the fixed-width analyses currently available~\citep{dayan2026expressive}.
Such a theory would clarify the trade-off between transferability and expressivity in passing from the duplication-compatible GMN to the more restrictive matrix-product GMN.
It would also sharpen our understanding of \Cref{thm:guarantee-2} by characterizing the target functions that are uniformly representable by matrix-product GMNs.

\subsection*{AI use statement}

Generative AI tools were used extensively in this work, under the authors' direction and review throughout.
Their use was iterative rather than one-shot: nothing was adopted as returned from a single prompt, and every AI-assisted artifact went through repeated rounds of author-specified goals, assistant drafts, and author correction.

We used generative AI to \emph{implement methods}, to \emph{generate synthetic datasets}, to \emph{clean and reformat datasets}, to \emph{design or provide feedback on research methodology or experiments}, to \emph{provide critical ingredients for proving mathematical claims}, to \emph{assist in the writing of proofs}, to \emph{propose or refine hypotheses}, and, in the limited sense described below, to \emph{formulate mathematical claims}.
We did not use generative AI to \emph{help develop theoretical models or conceptual frameworks}, or to \emph{interpret results}.
The remaining required-disclosure tasks (\emph{assisting with translation} and \emph{support for qualitative and thematic data analysis}) are not applicable to this work.
We also used generative AI to \emph{edit the paper to improve readability} and to \emph{summarize or analyze existing literature}.
We have reviewed all AI-assisted work.

The main text was written by the authors; the generative AI polished, proofread, and typeset it.
In the theoretical part the main ideas are the authors' own, including but not limited to, the application of the transferability framework to the GMN setting, the weight-space equivalences and metrics, the design of transferable GMN modifications, and the use of Tensor Programs to analyze the distribution of trained weights.
The generative AI wrote up arguments the authors had already settled on, searched them for errors and missing hypotheses, and checked notation and cross-references, over several rounds of author correction per argument.
Most theorem statements and proof strategies are likewise the authors' own, with the AI assistant extending them to more general settings and carrying out mundane calculations.
We do highlight that the idea behind \Cref{thm:guarantee-2} was proposed by the generative AI, and reached its final form through repeated exchanges in which the authors fixed the setting and corrected the argument.
All AI-generated statements and proofs were substantially revised and reorganized by the authors for readability, clarity and correctness. They were verified independently of the AI assistant's own checks and, wherever possible, validated numerically.
The experimental codebase was written with the AI coding assistant under the same process of author specification, inspection, and debugging; all results are logged and audited by the authors, with the AI tools helping to tabulate and visualize them.

We take responsibility for the final content of this work, including any text, claims, or artifacts produced with the aid of generative AI.


\subsubsection*{Acknowledgments}
YM was partially funded by an Amazon Fellowship and by NSF BSF 2430292. HM is supported by the Israel Science Foundation through a personal grant (ISF 264/23)
and an equipment grant (ISF 532/23), and by the Career Advancement Chairs in Artificial
Intelligence – Schmidt Futures.
SV was partially funded by NSF–Simons
Research Collaboration on the Mathematical and Scientific Foundations of Deep Learning (MoDL)
(NSF DMS 2031985), NSF CAREER 2339682, NSF BSF 2430292, and a Sloan Fellowship.

\bibliographystyle{plainnat}
\bibliography{bibliography}

\appendix
\crefalias{section}{appendix}
\crefalias{subsection}{appendix}
\crefalias{subsubsection}{appendix}
\ifdefined\arxivversion\else
\newpage
\section*{Contents}
\printcontents[appendixtoc]{l}{1}{}
\clearpage
\fi

\section{Structure of the weight space}
\label{app:weight-space-structure}
\Cref{sec:transferable_gmn} builds the transferable GMN to respect the following two weight space structures: (1) invariance to transformations that form wider weights representing the same function, and (2) continuity with respect to a distance that compares weights of different widths.
We now discuss these structures in more detail, showing that they are not arbitrary choices but rather the natural ones to consider.

We first discuss in~\Cref{sec:duplication-equivalence} the duplication widening~\eqref{eq:dup} and the block-wise widening~\eqref{eq:blockwise} from the main text.
Then, in~\Cref{sec:zero-padding-equivalence}, we explore another natural transformation that also produces a wider network representing the same function.
We further explain why we do not consider this in our design of the transferable GMN.
Finally, in~\Cref{sec:consistent-sequence-formulation}, we use the consistent-sequence framework of \citet{levin2025transferring} (reviewed in \Cref{sec:related-anydim}) to model the weight space structure across different widths, and give the details of the norms defined on the weight space discussed in~\Cref{sec:transferable_gmn}.

\paragraph{A common derivation.}
Both \Cref{sec:duplication-equivalence} and \Cref{sec:zero-padding-equivalence} instantiate one recipe, which we set up once here.
Fix an MLP \eqref{eq:mlp} with weights $\theta$ and widths $(n_\ell)_{\ell=0}^{L+1}$, and read it as a chain of maps,
\[
\RR^{n_0}
\xrightarrow{W^{(1)}}
\RR^{n_1}
\xrightarrow{+b^{(1)}}
\RR^{n_1}
\xrightarrow{\sigma}
\RR^{n_1}
\xrightarrow{W^{(2)}}
\cdots
\xrightarrow{W^{(L+1)}}
\RR^{n_{L+1}}
\xrightarrow{+b^{(L+1)}}
\RR^{n_{L+1}}.
\]
Fix target widths $(N_\ell)_{\ell=0}^{L+1}$ with $N_0=n_0$ and $N_{L+1}=n_{L+1}$, so that only the hidden layers grow, together with injective linear maps
\[
\varphi_\ell\colon \RR^{n_\ell}\to\RR^{N_\ell},
\qquad \ell=0,\dots,L+1,
\qquad
\varphi_0=I_{n_0},
\quad
\varphi_{L+1}=I_{n_{L+1}},
\]
one per layer.
The maps $\varphi_\ell$ give the prescription for how a width-$n_\ell$ activation vector is to be written at width $N_\ell$, and one can think of them as defining a structure-preserving embedding of the entire narrow network into the wide one.
We ask: what are the wide weights $\theta^\uparrow= (\Wlup, \blup)$ of widths $(N_\ell)_{\ell=0}^{L+1}$ so that the diagram
\[
\begin{tikzcd}[column sep=3em]
\RR^{n_{\ell-1}} \arrow[r, "W^{(\ell)}"] \arrow[d, "\varphi_{\ell-1}"'] &
\RR^{n_\ell} \arrow[r, "+b^{(\ell)}"] \arrow[d, "\varphi_\ell"] &
\RR^{n_\ell} \arrow[r, "\sigma"] \arrow[d, "\varphi_\ell"] &
\RR^{n_\ell} \arrow[d, "\varphi_\ell"] \\
\RR^{N_{\ell-1}} \arrow[r, "\Wlup"'] &
\RR^{N_\ell} \arrow[r, "+\blup"'] &
\RR^{N_\ell} \arrow[r, "\sigma"'] &
\RR^{N_\ell}
\end{tikzcd}
\]
commutes for every $\ell$.
That is,
\begin{equation}
\label{eq:phi-squares}
\Wlup\varphi_{\ell-1}=\varphi_\ell W^{(\ell)},
\qquad
\blup=\varphi_\ell b^{(\ell)},
\qquad
\sigma\circ\varphi_\ell=\varphi_\ell\circ\sigma .
\end{equation}
This asks for the wide network to match the narrow one at each layer $\ell$ (via the map $\varphi_\ell$), that is, $\xlup = \varphi_\ell x^{(\ell)}$ and $\alup = \varphi_\ell a^{(\ell)}$ for every $\ell$.
In particular, this asks for more than $\mc{R}(\theta^\uparrow)=\mc{R}(\theta)$, which is given by the case $\ell=L+1$.

We consider two natural choices of $\varphi_\ell$.
In \Cref{sec:duplication-equivalence} we take $\varphi_\ell$ to duplicate the coordinates; in \Cref{sec:zero-padding-equivalence} we take it to zero-pad them. 

\subsection{Widening by duplicating neurons}
\label{sec:duplication-equivalence}
We first consider the case where the wide network reproduces the narrow network's forward pass with each hidden neuron replaced by several copies of itself.
In other words, let
\begin{equation}
\label{eq:phi-dup}
\varphi_\ell
\colon \RR^{n_\ell}\to \RR^{N_\ell},
\qquad
x\mapsto x\otimes \one_{k_\ell}.
\end{equation}
In words, $\varphi_\ell$ duplicates the $i$th coordinate of $x$ exactly $k_\ell$ times.
We ask what the corresponding conditions on the wide weights $\theta^\uparrow$ are so that \eqref{eq:phi-squares} holds.

Two of the three conditions of \eqref{eq:phi-squares} are immediate.
The bias condition reads $\blup=b^{(\ell)}\otimes\one_{k_\ell}$.
The activation condition holds for \emph{every} elementwise activation $\sigma$.
The weight condition is the content of the following proposition.

\begin{proposition}
\label{prop:dup-characterization}
Let $\varphi_\ell$ be as in \eqref{eq:phi-dup} and partition $\Wlup\in\RR^{N_\ell\times N_{\ell-1}}$ into blocks $\Wlup_{ij}\in\RR^{k_\ell\times k_{\ell-1}}$, $i\in[n_\ell]$ and $j\in[n_{\ell-1}]$.
Then, for each $\ell$, $\Wlup\varphi_{\ell-1}=\varphi_\ell W^{(\ell)}$ if and only if
\[
\Wlup_{ij}\one_{k_{\ell-1}}
=
W^{(\ell)}_{ij}\one_{k_\ell}
\qquad\text{for all }i,j.
\]
\end{proposition}

\begin{proof}
For the $\Leftarrow$ direction, let $x\in\RR^{n_{\ell-1}}$.
The $j$th block of $\varphi_{\ell-1}x$ is $x_j\one_{k_{\ell-1}}$, so the $i$th block of $\Wlup\varphi_{\ell-1}x$ is, by block matrix multiplication,
\[
\sum_{j\in[n_{\ell-1}]}\Wlup_{ij}\left(x_j\one_{k_{\ell-1}}\right)
=\sum_{j\in[n_{\ell-1}]}x_jW^{(\ell)}_{ij}\one_{k_\ell}
=\left(W^{(\ell)}x\right)_i\one_{k_\ell},
\]
where the first equality is the hypothesis and the second is the definition of the matrix--vector product.
This is the $i$th block of $\varphi_\ell W^{(\ell)}x$, and $i$ was arbitrary.

For the $\Rightarrow$ direction, evaluate the identity at the standard basis vector $x=e_j\in\RR^{n_{\ell-1}}$, for which $\varphi_{\ell-1}e_j$ has $\one_{k_{\ell-1}}$ in its $j$th block and zeros elsewhere.
Reading off the $i$th output block of both sides gives $\Wlup_{ij}\one_{k_{\ell-1}}=W^{(\ell)}_{ij}\one_{k_\ell}$, for every pair $(i,j)$.
\end{proof}

Together with the bias condition, \Cref{prop:dup-characterization} proves that the block-wise widening~\eqref{eq:blockwise} of the main text preserves the represented function.
Moreover, since it is proved with ``if and only if'' conditions, it gives the most general widening operation that does so while keeping the hidden neurons of the wide network as copies of the narrow network's at each layer.

This can be interpreted in terms of the flow of information on the computation graph of the network.
In the narrow network, neuron $i$ of layer $\ell$ draws the amount $W^{(\ell)}_{ij}$ along its edge from neuron $j$ of layer $\ell-1$.
In the wide network, each of the $k_\ell$ copies of neuron $i$ must draw that same total $W^{(\ell)}_{ij}$ from the $k_{\ell-1}$ copies of neuron $j$ \emph{taken together}, but it is free to split the total among those copies however it likes, and different copies of $i$ may split it differently.
Only the total matters because the copies of neuron $j$ all carry the same activation, so the incoming contribution depends on the weights only through their sum.
At one extreme a copy of $i$ may route the whole of $W^{(\ell)}_{ij}$ through a single copy of $j$ and ignore the rest; at the other it may split it evenly.
These extremes single out the parallel widening and the duplication widening that we discuss below.

\paragraph{Two more rigid subfamilies.}
The block-wise widening~\eqref{eq:blockwise} is flexible but not very structured.
It is worth naming the successively more rigid choices inside the block-wise widening~\eqref{eq:blockwise}, since this is how the duplication widening \eqref{eq:dup} of the main text arises.

\begin{proposition}
\label{prop:kronecker}
Let $P^{(\ell)}\in\RR^{k_\ell\times k_{\ell-1}}$ and set $\Wlup\colonequals W^{(\ell)}\otimes P^{(\ell)}$.
Then for each $\ell$, $\Wlup\varphi_{\ell-1}=\varphi_\ell W^{(\ell)}$ holds for every $W^{(\ell)}\in\RR^{n_\ell\times n_{\ell-1}}$ if and only if $P^{(\ell)}\one_{k_{\ell-1}}=\one_{k_\ell}$, that is, if and only if every row of $P^{(\ell)}$ sums to one.
\end{proposition}

\begin{proof}
By the mixed-product property of the Kronecker product, for every $x\in\RR^{n_{\ell-1}}$,
\[
\left(W^{(\ell)}\otimes P^{(\ell)}\right)\left(x\otimes\one_{k_{\ell-1}}\right)
=\left(W^{(\ell)}x\right)\otimes\left(P^{(\ell)}\one_{k_{\ell-1}}\right),
\]
whereas $\varphi_\ell W^{(\ell)}x=\left(W^{(\ell)}x\right)\otimes\one_{k_\ell}$.
If $P^{(\ell)}\one_{k_{\ell-1}}=\one_{k_\ell}$ the two agree, which gives $\Leftarrow$.
Conversely, choose $W^{(\ell)}$ and $x$ with $u\colonequals W^{(\ell)}x\neq0$; then $u\otimes\left(P^{(\ell)}\one_{k_{\ell-1}}-\one_{k_\ell}\right)=0$ with $u\neq0$ forces $P^{(\ell)}\one_{k_{\ell-1}}=\one_{k_\ell}$.
\end{proof}

We call this \emph{Kronecker widening}.
The condition $P^{(\ell)}\one=\one$ says that the rows of $P^{(\ell)}$ sum to one; if $P^{(\ell)}$ is in addition entrywise nonnegative it is row-stochastic, and the split of each total $W^{(\ell)}_{ij}$ among the copies of $j$ is a genuine convex combination.

Two instances are worth having in mind.
\begin{enumerate}[leftmargin=*,topsep=0pt,itemsep=0pt]
    \item When adjacent hidden-layer multipliers agree, taking $P^{(\ell)}=I_k$ sends each copy of a neuron only to the corresponding copy in the next layer; with the analogous boundary choices at the first and last layers, this gives $k$ parallel, non-interacting copies of the narrow network whose outputs are recombined by the last layer.
    We call this the \emph{parallel widening}.
    \item Taking $P^{(\ell)}=k_{\ell-1}^{-1}\one_{k_\ell}\one_{k_{\ell-1}}^\top$ gives exactly the duplication widening \eqref{eq:dup} of \Cref{sec:transferable_gmn}.
    The factor $k_{\ell-1}^{-1}$ is the normalization that makes the row sums equal to one.
\end{enumerate}
We thus have nested families of widenings, all function-preserving: the block-wise widenings \eqref{eq:blockwise}, the Kronecker widenings inside them, and the parallel widening and the duplication widening \eqref{eq:dup} inside those.

\subsection{Widening by zero-padding neurons}
\label{sec:zero-padding-equivalence}
The other natural way to view a narrow network as a wide one is to add neurons that are switched off rather than copies of existing ones.
We record the equivalence this produces for future reference, followed by an explanation of why we do not use it as a building principle for our transferable GMNs.

Take $N_\ell\ge n_\ell$ arbitrary, put $m_\ell\colonequals N_\ell-n_\ell$ (so $m_0=m_{L+1}=0$), decompose $\RR^{N_\ell}=\RR^{n_\ell}\oplus\RR^{m_\ell}$, and let
\begin{equation}
\label{eq:phi-pad}
\varphi_\ell
\colon \RR^{n_\ell}\to \RR^{N_\ell},
\qquad
x\mapsto (x,0).
\end{equation}
The bias condition of \eqref{eq:phi-squares} reads $\blup=(b^{(\ell)},0)$.
The activation condition now has real content: $\sigma\left(\varphi_\ell x\right)=\left(\sigma(x),\sigma(0)\one_{m_\ell}\right)$, which equals $\varphi_\ell\sigma(x)$ for every $x$ if and only if $\sigma(0)=0$.
Zero-padding therefore already restricts the architectures it applies to, holding for ReLU and $\tanh$ but failing for the sigmoid or any activation with a nonzero offset, whereas duplication holds for every elementwise activation.

\begin{proposition}[The widenings compatible with zero-padding]
Let $\varphi_\ell$ be as in \eqref{eq:phi-pad} and write
\[
\Wlup
=
\begin{pmatrix}
\Wlup_{11} & \Wlup_{12}\\
\Wlup_{21} & \Wlup_{22}
\end{pmatrix}
\]
with the decompositions $\RR^{N_{\ell-1}}=\RR^{n_{\ell-1}}\oplus \RR^{m_{\ell-1}}$ and $\RR^{N_\ell}=\RR^{n_\ell}\oplus \RR^{m_\ell}$.
Then for each $\ell$, $\Wlup\varphi_{\ell-1}=\varphi_\ell W^{(\ell)}$ if and only if $\Wlup_{11}=W^{(\ell)}$ and $\Wlup_{21}=0$; the blocks $\Wlup_{12}$ and $\Wlup_{22}$ are unconstrained.
\end{proposition}

\begin{proof}
For the $\Leftarrow$ direction, block matrix multiplication gives $\Wlup(x,0)=\left(\Wlup_{11}x,\Wlup_{21}x\right)=\left(W^{(\ell)}x,0\right)=\varphi_\ell W^{(\ell)}x$ for every $x\in\RR^{n_{\ell-1}}$.

For the $\Rightarrow$ direction, evaluate the identity at $x=e_j\in\RR^{n_{\ell-1}}$ and read off the first $n_\ell$ and the last $m_\ell$ output coordinates separately, which gives $\Wlup_{11}e_j=W^{(\ell)}e_j$ and $\Wlup_{21}e_j=0$.
As $j$ ranges over a basis of $\RR^{n_{\ell-1}}$, this yields $\Wlup_{11}=W^{(\ell)}$ and $\Wlup_{21}=0$.
\end{proof}

The wide weights compatible with zero-padding are therefore exactly
\[
\Wlup
=
\begin{pmatrix}
W^{(\ell)} & *\\
0 & *
\end{pmatrix},
\qquad
\blup
=
\begin{pmatrix}
b^{(\ell)}\\
0
\end{pmatrix},
\]
with the starred blocks arbitrary, provided $\sigma(0)=0$.
The pattern of constraints is exactly the statement that the padded neurons are dead.
The zero blocks in the first column of $\Wlup$ and the second block of $\blup$ ensure that a padded neuron receives nothing, so its preactivation is $0$ and, since $\sigma(0)=0$, its activation is $0$ as well; the starred blocks, which are the weights leaving the padded neurons, are then unconstrained because they are always applied to zeros.

\paragraph{Why we do not build on this structure.}
Zero-padding is arguably the more familiar of the two structures: it formalizes the intuition that a narrow network is a wide network with unused neurons.
However, we argue that it is not the right structure to focus on for metanetworks, since we ultimately care about weights of different widths trained independently from random initialization (\Cref{sec:experiments,sec:sizegen}).
In a trained network of width $N_\ell$, every hidden neuron is doing work: the initialization is exchangeable across the neurons of a layer, and training under a permutation-equivariant loss preserves that exchangeability, so no neuron is singled out as one that ought to be zero.
In other words, there is a distributional permutation symmetry on weights trained from random initialization, so the wider network actually seen at test time, which we take to be a wide network trained from random initialization, is not expected to be close to a zero-padded version of a narrow network.
Hence we do not expect invariance to zero-padding widening to be as useful for size generalization in practice as invariance to duplication widening.
One might still ask whether imposing both invariances simultaneously could further improve size generalization.
We discuss the obstruction to imposing both in \Cref{sec:zero-padding-compatibility}.
Hence we must choose one invariance over the other.

\subsection{The weight space as a consistent sequence}
\label{sec:consistent-sequence-formulation}
We now focus on the structure of the weight space arising from duplication of neurons (\Cref{sec:duplication-equivalence}).
We refer to Appendix B of \citet{levin2025transferring} for the theoretical background that we follow here.
Out of the nested families of widening operations mentioned there, only the duplication widening \eqref{eq:dup} fits the consistent-sequence framework of \citet{levin2025transferring} (reviewed briefly in \Cref{sec:related-anydim}).
This is because a consistent sequence requires one designated linear embedding $V_n\hookrightarrow V_N$ per comparable pair of widths, whereas the block-wise family \eqref{eq:blockwise} singles out no map, and the Kronecker widening requires one arbitrary choice of $P^{(\ell)}$ per comparable pair of widths.
This framework buys the two things the main text uses: a single space $\Vinf$ holding weights of all widths at once, in which ``the same function at a different width'' becomes ``the same point'', and a distance on that space in which weights of different widths can be compared and continuity across widths can be stated.

\paragraph{Spaces, index set, and group action.}
For $n=(n_1,\dots,n_L)$, with $n_0=\din$ and $n_{L+1}=\dout$ held fixed, let $V_n$ be the weight space of \Cref{sec:transferable_gmn},
\[
V_n \colonequals
\bigoplus_{\ell=1}^{L+1}
\left(\RR^{n_\ell \times n_{\ell-1}} \oplus \RR^{n_\ell}\right).
\]
Order widths by divisibility, writing $n\preceq N$ when $n_i\mid N_i$ for every $i$.
This makes the index set directed, since $n$ and $m$ are both refined by $N_i=\operatorname{lcm}(n_i,m_i)$, which is what the direct limit below needs.

Each $V_n$ carries the hidden-neuron permutation symmetry.
Let $\msf G_n \colonequals S_{n_1}\times \cdots \times S_{n_L}$, identify each permutation with its permutation matrix, and for $g=(\sigma_1,\dots,\sigma_L)\in \msf G_n$ set $\sigma_0=I_{n_0}$ and $\sigma_{L+1}=I_{n_{L+1}}$, since input and output coordinates are fixed.
The action of $\msf G_n$ on $\theta=\left((\Wl,b^{(\ell)})\right)_{\ell=1}^{L+1}\in V_n$ is the familiar one,
\[
(g\cdot \theta)^{(\ell)}
\colonequals
\left(\sigma_\ell W^{(\ell)} \sigma_{\ell-1}^{-1},\
\sigma_\ell b^{(\ell)}\right),
\qquad \ell=1,\dots,L+1,
\]
which relabels the hidden activations by $a^{(\ell)}\mapsto\sigma_\ell a^{(\ell)}$ and leaves the represented function $\mc{R}(\theta)$ unchanged.

\paragraph{Embeddings.}
For $n\preceq N$ put $k_i=N_i/n_i$, with $k_0=k_{L+1}=1$, and let
\[
\varphi_{N,n}\colon V_n\hookrightarrow V_N,
\qquad
\theta\mapsto\theta^\uparrow
\]
be the duplication widening \eqref{eq:dup}.
In the rescaled coordinates the embedding is entrywise duplication,
\begin{equation}
\label{eq:embedding-rescaled}
\Wlupbar
=
\Wlbar\otimes \one_{k_\ell}\one_{k_{\ell-1}}^\top,
\qquad
\blup
=
b^{(\ell)}\otimes \one_{k_\ell},
\end{equation}
because $\Wlupbar=N_{\ell-1}\Wlup=k_{\ell-1}n_{\ell-1}\cdot k_{\ell-1}^{-1}\left(\Wl\otimes\one_{k_\ell}\one_{k_{\ell-1}}^\top\right)$.
This is the form we use below, and it is also why the first modification of the duplication-compatible GMN in~\Cref{sec:transferable_gmn} rescales the edge features by the fan-in: in these coordinates a widening does nothing but repeat features.

Furthermore, set the embedding of groups
\[
\msf G_n\hookrightarrow \msf G_N,
\qquad
g\mapsto g^\uparrow \colonequals (\sigma_1\otimes I_{k_1},\dots,\sigma_L\otimes I_{k_L}).
\]

The transition maps compose under successive duplication widenings, and the embeddings $\varphi_{N,n}$ are equivariant with respect to the group action, i.e.\ $\varphi_{N,n}(g\cdot\theta)=g^\uparrow\cdot\varphi_{N,n}(\theta)$ for every $g\in\msf G_n$.
Thus $(V_n,\msf G_n,\varphi_{N,n})$ satisfies Definition B.1 of \citet{levin2025transferring}, so it is a consistent sequence.

In the rescaled coordinates \eqref{eq:embedding-rescaled}, this consistent sequence can be seen as a direct sum of the duplication consistent sequence on sets in Appendix E.1 of~\cite{levin2025transferring}, except that the group actions are shared across layers.

We then produce the limit space (following Definitions B.2 and B.3 of \citet{levin2025transferring})
$\Vinf \colonequals \varinjlim_n V_n$, which is a $\Ginf \colonequals \varinjlim_n \msf G_n$-representation.
Concretely, one may picture $\Vinf$ as $\bigcup_n V_n$ with the identifications $\theta\sim\theta^\uparrow$, identifying weights of different widths that are related via the duplication widening.
The group $\Ginf$ acts on $\Vinf$ by permuting the hidden neurons of each width.

\paragraph{Compatible norms.}
We now equip $\Vinf$ with a norm that compares weights of different widths, following Appendix~B.3 of \cite{levin2025transferring}.
We use three norms on $V_n$: the \emph{normalized spectral norm}
\begin{equation}\label{eq:norm-op}
\norm[V_n,\op]{\theta}
\colonequals
\sum_{\ell=1}^{L+1}
\left(
\frac{\norm{\Wlbar}}{\sqrt{n_\ell n_{\ell-1}}}
+ \frac{\norm[2]{b^{(\ell)}}}{\sqrt{n_\ell}}
\right)
=
\sum_{\ell=1}^{L+1}
\left(
\sqrt{\frac{n_{\ell-1}}{n_\ell}}\norm{\Wl}
+ \frac{\norm[2]{b^{(\ell)}}}{\sqrt{n_\ell}}
\right),
\end{equation}
the \emph{normalized $\ell_1$ norm}
\begin{equation}
\label{eq:norm-lp}
\norm[V_n,1]{\theta}
\colonequals
\sum_{\ell=1}^{L+1}
\left(
\frac{\norm[1]{\Wlbar}}{n_\ell n_{\ell-1}}
+ \frac{\norm[1]{b^{(\ell)}}}{n_\ell}
\right)
=
\sum_{\ell=1}^{L+1}
\left(
\frac{\norm[1]{\Wl}}{n_\ell}
+ \frac{\norm[1]{b^{(\ell)}}}{n_\ell}
\right),
\end{equation}
and, as an intermediate analytical tool used only in \Cref{sec:one-layer-mup}, the \emph{normalized Frobenius norm}
\begin{equation}
\label{eq:norm-fro}
\norm[V_n,\fro]{\theta}^2
\colonequals
\sum_{\ell=1}^{L+1}
\left(
\frac{\norm[\fro]{\Wlbar}^2}{n_\ell n_{\ell-1}}
+ \frac{\norm[2]{b^{(\ell)}}^2}{n_\ell}
\right)
=
\sum_{\ell=1}^{L+1}
\left(
\frac{n_{\ell-1}}{n_\ell}\norm[\fro]{\Wl}^2
+ \frac{\norm[2]{b^{(\ell)}}^2}{n_\ell}
\right),
\end{equation}
where $\norm{\cdot}$ is the spectral norm, $\norm[1]{\cdot}$ the entrywise $\ell_1$ norm, and $\norm[\fro]{\cdot}$ the entrywise Frobenius norm.
The two norms of \eqref{eq:norms} in the main text are exactly \eqref{eq:norm-op} and \eqref{eq:norm-lp}.
Note that the normalized spectral and $\ell_1$ norms defined here are weight-space analogues of the norms attached to duplication sequences on graphs in Appendix F.1 of \citet{levin2025transferring}.

All three norms are $\msf G_n$-invariant, and are preserved by the embeddings, which is precisely the reason for our choices of normalization factors in \eqref{eq:norm-op}, \eqref{eq:norm-lp}, and \eqref{eq:norm-fro}.
Concretely, for the spectral norm, $\norm{A\otimes B}=\norm{A}\norm{B}$, and $\one_{k_\ell}\one_{k_{\ell-1}}^\top$ has rank one with $\norm{\one_{k_\ell}\one_{k_{\ell-1}}^\top}=\norm[2]{\one_{k_\ell}}\norm[2]{\one_{k_{\ell-1}}}=\sqrt{k_\ell k_{\ell-1}}$, so by \eqref{eq:embedding-rescaled}
\[
\frac{\norm{\Wlupbar}}{\sqrt{N_\ell N_{\ell-1}}}
=
\frac{\norm{\Wlbar}\sqrt{k_\ell k_{\ell-1}}}
{\sqrt{k_\ell n_\ell k_{\ell-1}n_{\ell-1}}}
=
\frac{\norm{\Wlbar}}{\sqrt{n_\ell n_{\ell-1}}},
\qquad
\frac{\norm[2]{\blup}}{\sqrt{N_\ell}}
=
\frac{\norm[2]{b^{(\ell)}}\sqrt{k_\ell}}{\sqrt{k_\ell n_\ell}}
=
\frac{\norm[2]{b^{(\ell)}}}{\sqrt{n_\ell}}.
\]
For the $\ell_1$ norm, $\norm[1]{A\otimes B}=\norm[1]{A}\norm[1]{B}$ and $\norm[1]{\one_{k_\ell}\one_{k_{\ell-1}}^\top}=k_\ell k_{\ell-1}$, so
\[
\frac{\norm[1]{\Wlupbar}}{N_\ell N_{\ell-1}}
=
\frac{\norm[1]{\Wlbar}k_\ell k_{\ell-1}}
{k_\ell n_\ell k_{\ell-1}n_{\ell-1}}
=
\frac{\norm[1]{\Wlbar}}{n_\ell n_{\ell-1}},
\quad
\frac{\norm[1]{\blup}}{N_\ell}
=
\frac{\norm[1]{b^{(\ell)}}k_\ell}{k_\ell n_\ell}
=
\frac{\norm[1]{b^{(\ell)}}}{n_\ell}.
\]
For the Frobenius norm, the same identity holds, $\norm[\fro]{A\otimes B}=\norm[\fro]{A}\norm[\fro]{B}$, and $\one_{k_\ell}\one_{k_{\ell-1}}^\top$ again has $\norm[\fro]{\one_{k_\ell}\one_{k_{\ell-1}}^\top}=\sqrt{k_\ell k_{\ell-1}}$, being rank one, so exactly as in the spectral-norm computation above,
\[
\frac{\norm[\fro]{\Wlupbar}^2}{N_\ell N_{\ell-1}}
=
\frac{\norm[\fro]{\Wlbar}^2}{n_\ell n_{\ell-1}},
\qquad
\frac{\norm[2]{\blup}^2}{N_\ell}
=
\frac{\norm[2]{b^{(\ell)}}^2}{n_\ell}.
\]
Summing over $\ell$ gives $\norm[V_N]{\varphi_{N,n}(\theta)}=\norm[V_n]{\theta}$ for all three norms, so the embeddings are isometries, as claimed in \Cref{sec:transferable_gmn}.
Therefore, all three norms are compatible with respect to the consistent sequence $(V_n,\msf G_n,\varphi_{N,n})$, in the sense of Definition 2.5 of \citet{levin2025transferring}, and hence, they induce norms $\norm[\Vinf,\op]{\cdot}$, $\norm[\Vinf,1]{\cdot}$, or $\norm[\Vinf,\fro]{\cdot}$ on $\Vinf$, whose value at a point may be computed at any representative width (see Proposition~B.8 of \citet{levin2025transferring}).

\paragraph{The symmetrized distances across widths.}
We further follow Definition~B.9 of \citet{levin2025transferring} to define a symmetrized distance on the limit space that takes into account the hidden-neuron permutation symmetry.
This construction is carried out separately for each of the three norms $\norm[\Vinf,\op]{\cdot}$, $\norm[\Vinf,1]{\cdot}$, $\norm[\Vinf,\fro]{\cdot}$ of \eqref{eq:norm-op}--\eqref{eq:norm-fro}, producing three symmetrized distances $d_\infty^{\op}$, $d_\infty^{1}$, $d_\infty^{\fro}$.
Fix $\star\in\{\op,1,\fro\}$, and let $\overline{\Vinf}^\star$ denote the completion of $\Vinf$ with respect to $\norm[\Vinf,\star]{\cdot}$; the $\Ginf$-action extends from $\Vinf$ to $\overline{\Vinf}^\star$ by isometries.
The norm already compares weights of different widths, but it insists on a particular labeling of the hidden neurons, which a metanetwork should not; we therefore symmetrize over $\Ginf$,
\begin{equation*}
d_\infty^\star(x,y)
\colonequals
\inf_{g\in \Ginf} \norm[\overline{\Vinf}^\star]{g\cdot x - y},
\qquad x,y\in \overline{\Vinf}^\star.
\end{equation*}
By Proposition B.10 of \citet{levin2025transferring}, $d_\infty^\star$ is a pseudometric on $\overline{\Vinf}^\star$ and a genuine metric on the set of orbit closures under the $\Ginf$-action.
Repeating this for each $\star\in\{\op,1,\fro\}$ gives the three symmetrized distances used throughout the rest of the paper.

\paragraph{Different interpretations behind different choices of norm.}
We consider here two different norms on the weight space for the metanetwork itself, the normalized spectral norm~\eqref{eq:norm-op} and the normalized $\ell_1$ norm~\eqref{eq:norm-lp}.
The choice of operator norm is natural from the perspective of the represented function, since in an MLP the weight matrices act on the activations of the previous layer via matrix multiplication, and the operator norm is the natural way to measure how much a linear map can stretch a vector.
The $\ell_1$ norm is defined via entrywise inspection of the weight matrices, and hence it is natural from the perspective of a duplication-compatible GMN, which reads each weight entry through its own learnable map.

We will not discuss the interpretation of the normalized Frobenius norm~\eqref{eq:norm-fro} as it serves only as an analytical intermediary in \Cref{sec:one-layer-mup}.

\paragraph{Comparing the three normalized norms.}
We now establish the relationship between the three norms in the limit space $\Vinf$.
\begin{proposition}
\label{prop:op-fro-comparison}
Both the normalized spectral norm and the normalized $\ell_1$ norm are dominated by the normalized Frobenius norm: for every $\theta\in\Vinf$,
\[
\max\{\norm[\Vinf,\op]{\theta},\norm[\Vinf,1]{\theta}\}
\le
\sqrt{2(L+1)}\norm[\Vinf,\fro]{\theta}.
\]
And consequently, the symmetrized distances also satisfy the same domination inequalities:
for every $x,y\in\overline\Vinf$,
\[
\max\{d_\infty^{\op}(x,y),d_\infty^{1}(x,y)\}
\le
\sqrt{2(L+1)}d_\infty^{\fro}(x,y).
\]
\end{proposition}
\begin{proof}
For each width $n$ and every $\theta\in V_n$,
\begin{align*}
\norm[V_n,\op]{\theta}
&=
\sum_{\ell=1}^{L+1}\left(\frac{\norm{\Wlbar}}{\sqrt{n_\ell n_{\ell-1}}}+\frac{\norm[2]{b^{(\ell)}}}{\sqrt{n_\ell}}\right)\\
&\le
\sum_{\ell=1}^{L+1}\left(\frac{\norm[\fro]{\Wlbar}}{\sqrt{n_\ell n_{\ell-1}}}+\frac{\norm[2]{b^{(\ell)}}}{\sqrt{n_\ell}}\right) \tag{$\norm{A}\le\norm[\fro]{A}$}\\
&\le \sqrt{2(L+1)} \norm[V_n,\fro]{\theta} \tag{Cauchy--Schwarz on $2(L+1)$ terms}
\end{align*}
Similarly,
\begin{align*}
\norm[V_n,1]{\theta}
&=
\sum_{\ell=1}^{L+1}\left(\frac{\norm[1]{\Wlbar}}{n_\ell n_{\ell-1}}+\frac{\norm[1]{b^{(\ell)}}}{n_\ell}\right)\\
&\le
\sum_{\ell=1}^{L+1}\left(\frac{\norm[\fro]{\Wlbar}}{\sqrt{n_\ell n_{\ell-1}}}+\frac{\norm[2]{b^{(\ell)}}}{\sqrt n_\ell}\right) \tag{Cauchy--Schwarz on matrix and vector entries}\\
&\le \sqrt{2(L+1)} \norm[V_n,\fro]{\theta} \tag{Cauchy--Schwarz on $2(L+1)$ terms}
\end{align*}
Since the embeddings $\varphi_{N,n}$ are isometries for all three norms, the same inequalities hold on $\Vinf$.

For the symmetrized distances, for any $g\in \Ginf$,
\[\norm[\Vinf, \op]{g\cdot x - y} \leq \sqrt{2(L+1)} \norm[\Vinf, \fro]{g\cdot x - y}.\]
Taking the infimum over $g\in \Ginf$ gives $d_\infty^{\op}(x,y) \leq \sqrt{2(L+1)} d_\infty^{\fro}(x,y)$, and similarly for $d_\infty^1$.
\end{proof}

\begin{proposition}
\label{prop:one-layer-op-domination}
In the case of 1-layer MLP, $L=1$, the normalized Frobenius norm and the normalized $\ell_1$ norm are dominated by the normalized spectral norm:
\[
\max\{\norm[\Vinf,\fro]{\theta},\norm[\Vinf,1]{\theta}\}
\le
\max\{1,\sqrt{\din},\sqrt{\dout}\}\norm[\Vinf,\op]{\theta}.
\]
\end{proposition}
\begin{proof}
For each width $n$ and every $\theta=( W^{(1)}, b^{(1)},  W^{(2)}, b^{(2)})\in V_n$, the two matrix blocks $\overline W^{(1)}$ and $\overline W^{(2)}$ have rank at most $\din$ and at most $\dout$, respectively, so the inequality $\norm[\fro]{A}\le\sqrt{\operatorname{rank}(A)}\norm{A}$ gives
\[
\norm[\fro]{\overline W^{(1)}}
\le
\sqrt{\din}\norm{\overline W^{(1)}},
\qquad
\norm[\fro]{\overline W^{(2)}}
\le
\sqrt{\dout}\norm{\overline W^{(2)}}.
\]
Then,
\begin{align*}
\norm[V_n,\fro]{\theta}
&\le
\frac{\norm[\fro]{\overline W^{(1)}}}{\sqrt{n\din}}
+\frac{\norm[2]{b^{(1)}}}{\sqrt n}
+\frac{\norm[\fro]{\overline W^{(2)}}}{\sqrt{n\dout}}
+\frac{\norm[2]{b^{(2)}}}{\sqrt{\dout}}
\tag{$\sum x_i^2\le (\sum x_i)^2$ for $x_i\geq 0$}\\
&\le
\frac{\sqrt{\din}\norm{\overline W^{(1)}}}{\sqrt{n\din}}
+\frac{\norm[2]{b^{(1)}}}{\sqrt n}
+\frac{\sqrt{\dout}\norm{\overline W^{(2)}}}{\sqrt{n\dout}}
+\frac{\norm[2]{b^{(2)}}}{\sqrt{\dout}}\\
&\le
\max\{1,\sqrt{\din},\sqrt{\dout}\}\norm[V_n,\op]{\theta}.
\end{align*}
Similarly,
\begin{align*}
\norm[V_n,1]{\theta}
&=
\frac{\norm[1]{\overline W^{(1)}}}{n\din}
+\frac{\norm[1]{b^{(1)}}}{n}
+\frac{\norm[1]{\overline W^{(2)}}}{n\dout}
+\frac{\norm[1]{b^{(2)}}}{\dout}\\
&\le
\frac{\norm[\fro]{\overline W^{(1)}}}{\sqrt{n\din}}
+\frac{\norm[2]{b^{(1)}}}{\sqrt n}
+\frac{\norm[\fro]{\overline W^{(2)}}}{\sqrt{n\dout}}
+\frac{\norm[2]{b^{(2)}}}{\sqrt{\dout}}
\tag{Cauchy--Schwarz on matrix and vector entries}\\
&\le
\max\{1,\sqrt{\din},\sqrt{\dout}\}\norm[V_n,\op]{\theta}.
\end{align*}
Since the embeddings $\varphi_{N,n}$ are isometries for all three norms, the same inequalities hold on $\Vinf$.
\end{proof}

In all depths, neither the normalized spectral norm nor the normalized Frobenius norm is dominated by the normalized $\ell_1$ norm with a width-independent constant.
When $L\ge2$, neither the normalized Frobenius norm nor the normalized $\ell_1$ norm is dominated by the normalized spectral norm with a width-independent constant.

\section{Theoretical properties of transferable GMNs}\label{sec:transferable_gmn_proofs}
In Section~\ref{sec:transferable_gmn}, we propose three modifications to the GMN to make it duplication-compatible: fan-in rescaling of the input weights, mean aggregation, and a layer-wise mean readout.
We then discuss its two shortcomings, and propose the matrix-product GMN to address them.
We provide all the proofs of the theoretical properties of the duplication-compatible GMN and the matrix-product GMN in this section of Appendix.

\Cref{sec:dup-comptible-gmn-properties} proves the duplication-compatible GMN's duplication compatibility.
\Cref{sec:blockwise-compat} shows that this comes at the cost of incompatibility with the larger family of block-wise widenings, and that the matrix-product GMN repairs this.
\Cref{sec:misalignment} discusses a further shortcoming of the duplication-compatible GMN, task-model misalignment.
\Cref{sec:matprod-spectral} proves the matrix-product GMN's Lipschitz continuity with respect to the normalized spectral norm.
Finally, \Cref{sec:zero-padding-compatibility} shows that our modified GMNs cannot be further made compatible with zero-padding widening discussed in \Cref{sec:zero-padding-equivalence}.

\subsection{Duplication-compatibility}\label{sec:dup-comptible-gmn-properties}

We prove duplication compatibility for the simple GMN of \Cref{sec:related-weightsym} after the three modifications of \Cref{sec:transferable_gmn}.
The point of the proof is to show that, at each round of message passing, the node and edge features on the wide network's computation graph are literally copies of the ones on the narrow network's computation graph.

\begin{proof}[Proof of \Cref{prop:dup-invariance}]
Let $\theta^\uparrow$ be the duplication widening \eqref{eq:dup} of $\theta$ with width multipliers $k_1,\dots,k_L$ and $k_0=k_{L+1}=1$, and write $v^{(\ell)}_{i,a}$, $a\in[k_\ell]$, for the copies of the narrow node $v^{(\ell)}_i$.
We prove by induction on the message-passing round $t=0,1,\dots,T$ that
\[
h^{(\ell),[t]}_{i,a}=h^{(\ell),[t]}_i,
\qquad
e^{(\ell),[t]}_{(j,b)(i,a)}=e^{(\ell),[t]}_{ji}
\]
for every layer $\ell$, every node index $i$ and predecessor index $j$, and every pair of copy indices $a,b$.

\emph{Base case ($t=0$).}
For the duplicated edge from $v^{(\ell-1)}_{j,b}$ to $v^{(\ell)}_{i,a}$, fan-in rescaling gives
\[
\Wlupbar_{(i,a)(j,b)}
=
n_{\ell-1}k_{\ell-1}
\cdot
k_{\ell-1}^{-1}W^{(\ell)}_{ij}
=
n_{\ell-1}W^{(\ell)}_{ij}
=
\Wlbar_{ij},
\]
so every edge copy carries the same initial raw edge feature as the narrow network's edge, and the bias part of \eqref{eq:dup} gives $b^{(\ell)\uparrow}_{i,a}=b^{(\ell)}_i$, so every node copy carries the same initial raw node feature.
The initial node and edge maps $\psi_v,\psi_e$ of the simple GMN are shared across nodes and edges of the same type, so applying them to these identical raw features gives
\[
h^{(\ell),[0]}_{i,a}=h^{(\ell),[0]}_i,
\qquad
e^{(\ell),[0]}_{(j,b)(i,a)}=e^{(\ell),[0]}_{ji},
\]
which is the claim at $t=0$.

\emph{Inductive step.}
Assume the claim holds at round $t$.
Fix a node $i$ and a predecessor $j$.
For arbitrary copy indices $a,b$, the shared message map gives
\[
m^{(\ell),[t]}_{(j,b)\to(i,a)}
=
\psi_m\left(e^{(\ell),[t]}_{(j,b)(i,a)},h^{(\ell-1),[t]}_{j,b}\right)
=
\psi_m\left(e^{(\ell),[t]}_{ji},h^{(\ell-1),[t]}_j\right)
=
m^{(\ell),[t]}_{j\to i},
\]
so the forward mean aggregate at the copied node is
\[
\frac{1}{n_{\ell-1}k_{\ell-1}}
\sum_{j\in[n_{\ell-1}]}
\sum_{b\in[k_{\ell-1}]}
m^{(\ell),[t]}_{(j,b)\to(i,a)}
=
\frac{1}{n_{\ell-1}}
\sum_{j\in[n_{\ell-1}]}
m^{(\ell),[t]}_{j\to i}.
\]
The same calculation gives the backward aggregate in the bidirectional GMN, with $n_{\ell+1}k_{\ell+1}$ in place of $n_{\ell-1}k_{\ell-1}$.
The copied and narrow node features are then updated by the same shared map $\psi_h$ applied to the same arguments, hence remain equal at round $t+1$; the edge update is likewise shared and depends only on the old edge feature and the two endpoint node features, which are equal by the induction hypothesis, so copied edge features also remain equal at round $t+1$.
This completes the induction.

\emph{Graph-level readout.}
Since $\psi_r$ is applied pointwise and the readout pools by a per-layer mean, the hidden-layer summaries agree:
\[
\frac{1}{n_\ell k_\ell}
\sum_{i\in[n_\ell]}
\sum_{a\in[k_\ell]}
\psi_r\left(h^{(\ell),[T]}_{i,a}\right)
=
\frac{1}{n_\ell}
\sum_{i\in[n_\ell]}
\psi_r\left(h^{(\ell),[T]}_i\right).
\]
Since $k_0=k_{L+1}=1$, the input and output widths are fixed, so their coordinatewise states are listed in the same order for $\theta$ and $\theta^\uparrow$; the final readout map $\rho$ therefore receives the same input for both, and $f^\Theta(\theta^\uparrow)=f^\Theta(\theta)$.
\end{proof}

\subsection{Compatibility with block-wise widening}\label{sec:blockwise-compat}

\paragraph{Incompatibility of the duplication-compatible GMN.}
We show the incompatibility of the duplication-compatible GMN with the full block-wise widening family \eqref{eq:blockwise} by constructing a specific counterexample.

\emph{The narrow MLP.}
Let $\theta$ be a scalar-input, scalar-output MLP with one hidden layer and no biases, with weights
\[
W^{(1)}=w,
\qquad
W^{(2)}=1,
\]
so it represents the function $x\mapsto \sigma(wx)$ for activation function $\sigma$.

\emph{The widened MLP.}
Let $\theta^\uparrow$ widen only the hidden layer, to width $2$, with weights
\[
{W^\uparrow}^{(1)}=\begin{bmatrix}w\\w\end{bmatrix},
\qquad
{W^\uparrow}^{(2)}=\begin{bmatrix}1&0\end{bmatrix}.
\]
This satisfies the block-wise widening condition \eqref{eq:blockwise} but not the duplication widening \eqref{eq:dup}, which would instead split the hidden-to-output weight evenly as $(\tfrac12,\tfrac12)$.
It represents the same function as $\theta$: both hidden copies compute $\sigma(wx)$, and the output recombines them as $1\cdot\sigma(wx)+0\cdot\sigma(wx)=\sigma(wx)$.

\emph{The GMN.}
Take one round of message passing, $T=1$.
Let the initial edge map be the identity, $\psi_e=\mathrm{id}$ and the initial node map $\psi_v$ be arbitrary.
On the narrower network $\theta$, the output edge carries initial edge feature $1$.
On the widened network $\theta^\uparrow$, the two output edges carry features $2$ and $0$.
Choose the message MLP $\psi_m(e,h)=\operatorname{ReLU}(e-\tfrac12)$, which discards the source-node feature $h$; choose the node-update MLP to copy the mean incoming message, $\psi_h(h,m^\rightarrow,m^\leftarrow)=m^\rightarrow$; and choose the final readout $\rho$ to select the resulting output-node feature.

\emph{The two outputs.}
On $\theta$, the output-node feature after one round is $\operatorname{ReLU}(1-\tfrac12)=\tfrac12$.
On $\theta^\uparrow$, mean aggregation over the $k_1=2$ incoming edges gives
\[
\frac12\operatorname{ReLU}\left(2-\tfrac12\right)
+
\frac12\operatorname{ReLU}\left(0-\tfrac12\right)
=
\frac34.
\]
Since $\tfrac12\ne\tfrac34$, this GMN's two graph-level outputs differ although $\theta$ and $\theta^\uparrow$ represent the same function.

\paragraph{Compatibility of the matrix-product GMN.}

\begin{proof}[Proof of \Cref{prop:matprod}(ii)]
\textbf{We first show that the forward matrix-product GMN is fully compatible with block-wise widenings \eqref{eq:blockwise}.}

Let $\theta^\uparrow$ be a block-wise widening \eqref{eq:blockwise} of $\theta$ with multipliers $k_1,\dots,k_L$ and $k_0=k_{L+1}=1$, so the widened widths are $N_\ell=n_\ell k_\ell$, and write $v^{(\ell)}_{i,a}$, $a\in[k_\ell]$, for the copies of the narrow node $v^{(\ell)}_i$.
We prove by induction on $t=0,\dots,T$ that
\begin{equation}
\label{eq:matprod-copies}
h^{(\ell),[t]\uparrow}_{i,a}=h^{(\ell),[t]}_i
\qquad\text{for every }\ell,\ i\in[n_\ell],\ a\in[k_\ell].
\end{equation}

\emph{Base case ($t=0$).}
The bias part of \eqref{eq:blockwise} gives $b^{(\ell)\uparrow}_{i,a}=b^{(\ell)}_i$, so the shared initial node map $\psi_v$ returns the same feature for every copy, and the input nodes carry the same constant at both widths.

\emph{Inductive step.}
Assume \eqref{eq:matprod-copies} at round $t$.
Then,
\[
\begin{split}
z^{\mathrm{in},[t]\uparrow}_{v^{(\ell)}_{i,a}}
&=
\frac{1}{N_{\ell-1}}
\sum_{j\in[n_{\ell-1}]}\sum_{b\in[k_{\ell-1}]}
\Wlupbar_{(i,a)(j,b)}A^{\mathrm{in}}_\ell h^{(\ell-1),[t]\uparrow}_{j,b}
\\
&=
\sum_{j\in[n_{\ell-1}]}
\left(\sum_{b\in[k_{\ell-1}]}\left(\Wlup_{ij}\right)_{ab}\right)
A^{\mathrm{in}}_\ell h^{(\ell-1),[t]}_j
\\
&=
\sum_{j\in[n_{\ell-1}]}
\left(\Wlup_{ij}\one_{k_{\ell-1}}\right)_a
A^{\mathrm{in}}_\ell h^{(\ell-1),[t]}_j
=
\sum_{j\in[n_{\ell-1}]}
W^{(\ell)}_{ij}A^{\mathrm{in}}_\ell h^{(\ell-1),[t]}_j
=
z^{\mathrm{in},[t]}_{v^{(\ell)}_i},
\end{split}
\]
In the forward variant the node update therefore receives identical arguments at every copy, and since $\psi_h$ is shared, \eqref{eq:matprod-copies} holds at round $t+1$, completing the induction.

\emph{Graph-level readout.}
Since $\psi_r$ is applied pointwise and all $k_\ell$ copies in hidden layer $\ell$ carry the same feature by \eqref{eq:matprod-copies}, the layer-wise mean over the widened hidden layer $\ell$ collapses back to the narrow summary:
\[
\frac{1}{N_\ell}
\sum_{i\in[n_\ell]}
\sum_{a\in[k_\ell]}
\psi_r\left(h^{(\ell),[T]\uparrow}_{i,a}\right)
=
\frac{1}{n_\ell}
\sum_{i\in[n_\ell]}
\psi_r\left(h^{(\ell),[T]}_i\right)
=:r_\ell.
\]
The input and output node features are passed to $\rho$ coordinatewise, in the same order at both widths because $k_0=k_{L+1}=1$.
Hence $f^\Theta(\theta^\uparrow)=f^\Theta(\theta)$.

\textbf{Next, we show that the bidirectional matrix-product GMN is compatible with the subclass of widenings that further satisfies the column-balance condition of \Cref{prop:matprod}(ii).}

\emph{Inductive step, backward aggregate.}
For the bidirectional variant the same computation on the reversed edges gives
\[
z^{\mathrm{out},[t]\uparrow}_{v^{(\ell)}_{i,a}}
=
\frac{N_\ell}{N_{\ell+1}}
\sum_{j\in[n_{\ell+1}]}
\left(\one_{k_{\ell+1}}^\top W^{(\ell+1)\uparrow}_{ji}\right)_a
A^{\mathrm{out}}_\ell h^{(\ell+1),[t]}_j,
\]
whereas
\[
z^{\mathrm{out},[t]}_{v^{(\ell)}_i}
=
\frac{n_\ell}{n_{\ell+1}}
\sum_{j\in[n_{\ell+1}]}
W^{(\ell+1)}_{ji}A^{\mathrm{out}}_\ell h^{(\ell+1),[t]}_j,
\]
because a copy of neuron $i$ now collects the weights in the $a$th \emph{column} of the block $W^{(\ell+1)\uparrow}_{ji}$.
As $N_\ell/N_{\ell+1}=(k_\ell/k_{\ell+1})(n_\ell/n_{\ell+1})$, the column-balance condition
\[
\one_{k_{\ell+1}}^\top W^{(\ell+1)\uparrow}_{ji}
=
\frac{k_{\ell+1}}{k_\ell}W^{(\ell+1)}_{ji}\one_{k_\ell}^\top
\qquad\text{for all }i,j,
\]
which is the condition of \Cref{prop:matprod}(ii) at layer $\ell+1$, makes the backward aggregate agree with the narrow one whenever \eqref{eq:matprod-copies} holds at round $t$.
Together with the forward aggregate calculation above, the node update again receives the same arguments at every copy, so the induction and the readout argument go through as before.
\end{proof}


\subsection{Task-model misalignment}\label{sec:misalignment}
Here we prove \Cref{prop:misalignment} regarding the misalignment between the target function and the duplication-compatible GMN model: they are continuous with respect to different norms on the weight space.

\paragraph{Continuity with respect to normalized spectral norm.}
\begin{proof}[Proof of \Cref{prop:misalignment}(i)]
\textbf{We first show the Lipschitz continuity of the realization map $\mc{R}$ with respect to the normalized spectral norm.}
Since we have set up in \Cref{sec:consistent-sequence-formulation} a consistent sequence of weight spaces $(V_n,\msf G_n,\varphi_{N,n})$ with compatible norms $\norm[V_n,\op]{\cdot}$, it is enough, by Proposition~C.2 of \citet{levin2025transferring}, to show that in each finite space $V_n$, the realization map $\mc{R}$ is $L_R$ Lipschitz on $\{\theta\in V_n: \norm[V_n,\op]{\theta}\le R\}$, where the constant $L_R$ is independent of the hidden widths.

Let $\theta,\widetilde\theta\in V_n$ satisfy
\[
\norm[V_n,\op]{\theta}\le R,
\qquad
\norm[V_n,\op]{\widetilde\theta}\le R.
\]
Since the input space $X$ is compact, choose $R_X$ such that $\norm[2]{u}\le R_X$ for every $u\in X$.

For $\eta\in V_n$ and $u\in X$, write $x_\eta^{(\ell)}(u)$ and $a_\eta^{(\ell)}(u)$ for the pre-activations and activations of \eqref{eq:mlp} with weights $\eta$ and input $a_\eta^{(0)}(u)=u$, reading the nonlinearity of the output layer as the identity, so that $a_\eta^{(L+1)}(u)=x_\eta^{(L+1)}(u)=\mc{R}(\eta)(u)$.
Set $\mc{L}\colonequals\max\{\mc{L}_\sigma,1\}$, where $\mc{L}_\sigma$ is the Lipschitz constant of $\sigma$; the nonlinearity of every layer $1\le\ell\le L+1$ is then $\mc{L}$-Lipschitz and maps $0\in\RR^{n_\ell}$ to a vector of norm at most $\sqrt{n_\ell}\abs{\sigma(0)}$.

Define, for $0\le\ell\le L+1$, the normalized activation size and discrepancy
\[
\rho_\ell
\colonequals
\sup_{u\in X}
\frac{\max\left\{\norm[2]{a_\theta^{(\ell)}(u)},\norm[2]{a_{\widetilde\theta}^{(\ell)}(u)}\right\}}{\sqrt{n_\ell}},
\qquad
d_\ell
\colonequals
\sup_{u\in X}
\frac{\norm[2]{a_\theta^{(\ell)}(u)-a_{\widetilde\theta}^{(\ell)}(u)}}{\sqrt{n_\ell}},
\]
and, for $1\le\ell\le L+1$, the per-layer parameter discrepancy
\[
\delta_\ell
\colonequals
\sqrt{\frac{n_{\ell-1}}{n_\ell}}
\norm{W_\theta^{(\ell)}-W_{\widetilde\theta}^{(\ell)}}
+
\frac{\norm[2]{b_\theta^{(\ell)}-b_{\widetilde\theta}^{(\ell)}}}{\sqrt{n_\ell}},
\qquad\text{so that}\qquad
\norm[V_n,\op]{\theta-\widetilde\theta}
=
\sum_{\ell=1}^{L+1}\delta_\ell.
\]
We prove by induction on the layer index $\ell$ that
\[
\rho_\ell\le C_\ell,
\qquad
d_\ell\le C'_\ell\norm[V_n,\op]{\theta-\widetilde\theta}
\]
for constants $C_\ell,C'_\ell$ depending only on $R,R_X,\mc{L}_\sigma,\sigma(0),L,\din$ and not on the hidden widths.
Throughout we use that each summand of $\norm[V_n,\op]{\eta}$ is bounded by the whole sum, so $\norm[V_n,\op]{\eta}\le R$ gives
\begin{equation}
\label{eq:misalign-op-summand}
\sqrt{\frac{n_{\ell-1}}{n_\ell}}\norm{W_\eta^{(\ell)}}\le R,
\qquad
\frac{\norm[2]{b_\eta^{(\ell)}}}{\sqrt{n_\ell}}\le R,
\qquad
1\le\ell\le L+1,
\end{equation}
and likewise $\delta_\ell\le\norm[V_n,\op]{\theta-\widetilde\theta}$.

\emph{Base case ($\ell=0$).}
Both networks read the same input, $a_\theta^{(0)}(u)=a_{\widetilde\theta}^{(0)}(u)=u$, so the two claims hold with
\[
C_0\colonequals\frac{R_X}{\sqrt{\din}},
\qquad
C'_0\colonequals0.
\]

\emph{Induction step.}
Let $1\le\ell\le L+1$ and assume the two bounds at $\ell-1$.
For the size bound, \eqref{eq:misalign-op-summand} give, for $\eta\in\{\theta,\widetilde\theta\}$ and every $u\in X$,
\[
\frac{\norm[2]{x_\eta^{(\ell)}(u)}}{\sqrt{n_\ell}}
\le
\sqrt{\frac{n_{\ell-1}}{n_\ell}}
\norm{W_\eta^{(\ell)}}
\frac{\norm[2]{a_\eta^{(\ell-1)}(u)}}{\sqrt{n_{\ell-1}}}
+
\frac{\norm[2]{b_\eta^{(\ell)}}}{\sqrt{n_\ell}}
\le
R\left(\rho_{\ell-1}+1\right).
\]
Applying the nonlinearity, and then the induction hypothesis $\rho_{\ell-1}\le C_{\ell-1}$,
\[
\rho_\ell
\le
\mc{L}R\left(\rho_{\ell-1}+1\right)+\abs{\sigma(0)}
\le
C_\ell
\colonequals
\mc{L}R\left(C_{\ell-1}+1\right)+\abs{\sigma(0)}.
\]
For the discrepancy bound, we have
\[
x_\theta^{(\ell)}(u)-x_{\widetilde\theta}^{(\ell)}(u)
=
\left(W_\theta^{(\ell)}-W_{\widetilde\theta}^{(\ell)}\right)a_\theta^{(\ell-1)}(u)
+
W_{\widetilde\theta}^{(\ell)}\left(a_\theta^{(\ell-1)}(u)-a_{\widetilde\theta}^{(\ell-1)}(u)\right)
+
\left(b_\theta^{(\ell)}-b_{\widetilde\theta}^{(\ell)}\right).
\]
Normalizing by $\sqrt{n_\ell}$ and apply the nonlinearity, we get
\[
\begin{split}
d_\ell
&\le
\mc{L}
\left(
\rho_{\ell-1}
\sqrt{\frac{n_{\ell-1}}{n_\ell}}
\norm{W_\theta^{(\ell)}-W_{\widetilde\theta}^{(\ell)}}
+
Rd_{\ell-1}
+
\frac{\norm[2]{b_\theta^{(\ell)}-b_{\widetilde\theta}^{(\ell)}}}{\sqrt{n_\ell}}
\right)
\\
&\le
\mc{L}\left(
(C_{\ell-1}+1)\delta_\ell
+
Rd_{\ell-1}
\right).
\end{split}
\]
By the induction hypothesis $d_{\ell-1}\le C'_{\ell-1}\norm[V_n,\op]{\theta-\widetilde\theta}$ and by $\delta_\ell\le\norm[V_n,\op]{\theta-\widetilde\theta}$,
\[
d_\ell
\le
C'_\ell\norm[V_n,\op]{\theta-\widetilde\theta},
\qquad
C'_\ell
\colonequals
\mc{L}\left(C_{\ell-1}+1+RC'_{\ell-1}\right).
\]
Both recursions involve only $R,\mc{L}_\sigma,\sigma(0)$ and the constants at $\ell-1$, so $C_\ell$ and $C'_\ell$ are again independent of the hidden widths, completing the induction.

\emph{Conclusion.}
The case $\ell=L+1$ of the induction is the desired bound: since $a_\theta^{(L+1)}(u)=\mc{R}(\theta)(u)$ and $n_{L+1}=\dout$ is fixed,
\[
\norm[\infty]{\mc{R}(\theta)-\mc{R}(\widetilde\theta)}
=
\sup_{u\in X}\norm[2]{\mc{R}(\theta)(u)-\mc{R}(\widetilde\theta)(u)}
=
\sqrt{\dout}d_{L+1}
\le
L_R\norm[V_n,\op]{\theta-\widetilde\theta},
\]
where $L_R\colonequals\sqrt{\dout}C'_{L+1}$ is independent of the hidden widths.

\textbf{We next show that, if $\sigma$ is unbounded, the realization map need not be continuous with respect to the normalized $\ell_1$ norm.}
Choose a sequence $a_m\in\RR$ such that $\abs{\sigma(a_m)}\to\infty$, and choose hidden widths $n_m$ so that $n_m\ge m\abs{a_m}$.
Passing to a tail of the sequence if necessary, assume also that $\sigma(a_m)\neq0$ for every $m$.
For each $m$, take all hidden widths equal to $n_m$.
Let $\theta_m=((W^{(\ell)},b^{(\ell)}))_{\ell=1}^{L+1}$ be the weights of a scalar-output, $L$-hidden-layer network, where all weights and biases are zero except
\[
b^{(L)}=a_m e_1\in\RR^{n_m},
\qquad
W^{(L+1)}=\sigma(a_m)^{-1}e_1^\top\in\RR^{1\times n_m},
\]
where $e_1$ is the first standard basis vector.
Then
\[
\norm[V_{(n_m,\dotsc,n_m)},1]{\theta_m}
=
\frac{\abs{a_m}}{n_m}
+
\frac{1}{\abs{\sigma(a_m)}}
\le
\frac{1}{m}
+
\frac{1}{\abs{\sigma(a_m)}}
\to0.
\]
However, for every $u\in X$, the first coordinate of $a_{\theta_m}^{(L)}(u)$ is $\sigma(a_m)$, so
\[
\mc{R}(\theta_m)(u)
=
\sigma(a_m)^{-1}\sigma(a_m)
=
1,
\qquad
\mc{R}(0)(u)=0.
\]
Hence $\theta_m\to0$ in $\norm[\Vinf,1]{\cdot}$ but $\mc{R}(\theta_m)$ does not converge to $\mc{R}(0)$ in the uniform norm.
\end{proof}
\begin{rmk}[Consequence for the prediction target]
Throughout the paper, we assumed that the prediction target depends only on the function the input MLP represents, i.e.\ $f^*=F^*\circ\mc{R}$ for a functional $F^*\colon\mc{C}(X,\RR^{\dout})\to\RR^k$, which it is reasonable to assume Lipschitz with respect to the uniform norm.
The target $f^*$ is then Lipschitz on every bounded ball of $\norm[\Vinf,\op]{\cdot}$ with a width-independent constant, but it need not be continuous with respect to $\norm[\Vinf,1]{\cdot}$.
For the latter, take $\dout=1$ and $F^*(g)=g(u_0)$ for a fixed $u_0\in X$, which is $1$-Lipschitz in the uniform norm. 
However $f^*$ is not continuous with respect to $\norm[\Vinf,1]{\cdot}$: the sequence $\theta_m$ above has $f^*(\theta_m)=1$ for every $m$ while $f^*(0)=0$, even though $\theta_m\to0$ in $\norm[\Vinf,1]{\cdot}$.
Together with \Cref{prop:misalignment}(ii), the target and the duplication-compatible GMN are thus Lipschitz with respect to different norms on $\Vinf$. 
This is exactly the misalignment that \citet{levin2025transferring} identify as an obstruction to size generalization.
\end{rmk}

\paragraph{Continuity with respect to normalized $\ell_1$ norm.}
\begin{proof}[Proof of \Cref{prop:misalignment}(ii)]
    \textbf{We first show the Lipschitz continuity of the GMN map $f^\Theta$ with respect to the normalized $\ell_1$ norm.}
Fix the learnable GMN parameters $\Theta$.
Since we have set up in \Cref{sec:consistent-sequence-formulation} a consistent sequence of weight spaces $(V_n,\msf G_n,\varphi_{N,n})$ with compatible norms $\norm[V_n,1]{\cdot}$, it is enough, by Proposition~C.2 of \citet{levin2025transferring}, to show that in each finite space $V_n$, the GMN map $f^\Theta$ is Lipschitz with a constant independent of the hidden widths.

Let $\theta,\widetilde\theta\in V_n$, and for the node and edge features after round $t$ define the average discrepancies
\[
\Delta_t^{(\ell)}
\colonequals
\frac{1}{n_\ell}
\sum_{i=1}^{n_\ell}
\norm[1]{h^{(\ell),[t]}_i-\widetilde h^{(\ell),[t]}_i},
\qquad
\Gamma_t^{(\ell)}
\colonequals
\frac{1}{n_{\ell-1}n_\ell}
\sum_{j=1}^{n_{\ell-1}}
\sum_{i=1}^{n_\ell}
\norm[1]{e^{(\ell),[t]}_{ji}-\widetilde e^{(\ell),[t]}_{ji}},
\]
with totals $\Delta_t\colonequals\sum_{\ell=0}^{L+1}\Delta_t^{(\ell)}$ and $\Gamma_t\colonequals\sum_{\ell=1}^{L+1}\Gamma_t^{(\ell)}$.
Let $A_\Theta\ge1$ be a common Lipschitz constant of the shared maps $\psi_v,\psi_e,\psi_m,\psi_m^\leftarrow,\psi_h,\psi_u$, a constant depending only on $\Theta$.

We prove by induction on the message-passing round $t$ that 
\[\Delta_t + \Gamma_t \le C_t \norm[V_n,1]{\theta-\widetilde\theta}\]
for constant $C$ depending only on $\Theta,T,\din,\dout$ and not on the hidden widths.

\emph{Base case ($t=0$).}
At the start of the message-passing $t=0$, we have $\Delta_0^{(0)}=0$ and
\[
\Delta_0^{(\ell)}
\le
\frac{A_\Theta}{n_\ell}
\norm[1]{b^{(\ell)}-\widetilde b^{(\ell)}},
\qquad
\Gamma_0^{(\ell)}
\le
\frac{A_\Theta}{n_{\ell-1}n_\ell}
\norm[1]{\Wlbar-\widetilde{\overline W}^{(\ell)}},
\qquad
\ell=1,\dots,L+1,
\]
and so $\Delta_0+\Gamma_0\le A_\Theta\norm[V_n,1]{\theta-\widetilde\theta}$.

\emph{Induction step.}
For $t\geq 0$, by the Lipschitz continuity of $\psi_m,\psi_m^\leftarrow,\psi_h,\psi_u$,
\[
\begin{split}
\Delta_{t+1}^{(\ell)}
&\le
A_\Theta^2\left(
\Delta_t^{(\ell)}+\Delta_t^{(\ell-1)}+\Delta_t^{(\ell+1)}
+\Gamma_t^{(\ell)}+\Gamma_t^{(\ell+1)}\right),
\\
\Gamma_{t+1}^{(\ell)}
&\le
A_\Theta\left(\Gamma_t^{(\ell)}+\Delta_t^{(\ell-1)}+\Delta_t^{(\ell)}\right),
\end{split}
\]
with the conventions $\Delta_t^{(-1)}=\Delta_t^{(L+2)}=\Gamma_t^{(0)}=\Gamma_t^{(L+2)}=0$ at the input and output layers.
For the forward GMN, $\psi_m^\leftarrow=0$ and hence the two terms $\Delta_t^{(\ell+1)},\Gamma_t^{(\ell+1)}$ drop out.

Summing over $\ell$, we have $\Delta_{t+1}\le A_\Theta^2(3\Delta_t+2\Gamma_t)$ and $\Gamma_{t+1}\le A_\Theta(2\Delta_t+\Gamma_t)$, so $\Delta_{t+1}+\Gamma_{t+1}\le5A_\Theta^2(\Delta_t+\Gamma_t)$.
Hence, the induction hypothesis holds at round $t+1$ with constant $C_{t+1}=5A_\Theta^2 C_t$.

\emph{Graph-level readout.}
In the layer-wise mean readout, by the Lipschitz continuity of $\psi_r$ and $\rho$,
\[
\norm[1]{f^\Theta(\theta)-f^\Theta(\widetilde\theta)}
\le
C_{\Theta,T}\norm[V_n,1]{\theta-\widetilde\theta},
\]
where $C_{\Theta,T}$ depends only on $\Theta,T,\din,\dout$ and not on the hidden widths.

\textbf{We now show that the duplication-compatible GMN class contains functions that are discontinuous in the normalized spectral norm.}
We construct a specific counterexample.

\emph{The weight sequence.}
For any depth $L\ge2$, take scalar input and output ($n_0=n_{L+1}=1$), and hidden widths $n_1=\cdots=n_L=n$.
Let $n$ range over powers of two, and let $H_n\in\{\pm1\}^{n\times n}$ be the Sylvester Hadamard matrix, so $H_n^\top H_n=nI_n$ and $\norm{H_n}=\sqrt n$.
Define weight $\theta_n\in V_n$ in the rescaled coordinates by $\overline W^{(2)}\colonequals H_n$ and all other rescaled weights and biases zero.
Then
\[
\norm[V_n,\op]{\theta_n}
=
\frac{\norm{H_n}}{\sqrt{n\cdot n}}
=
\frac{1}{\sqrt n}
\xrightarrow[n\to\infty]{}0,
\]
hence $\theta_n\to0$ in $\Vinf$.

\emph{The GMN.}
Take the duplication-compatible GMN of \Cref{sec:transferable_gmn} with $d_h=1$, one round of message passing $T=1$, forward messages only ($\psi_m^\leftarrow=0$), and
\[
\psi_v\equiv0,
\qquad
\psi_e=\mathrm{id},
\qquad
\psi_m(e,h)=\operatorname{ReLU}(e)+\operatorname{ReLU}(-e)=\abs{e},
\]
\[
\psi_h(h,m^\rightarrow,m^\leftarrow)=m^\rightarrow,
\qquad
\psi_r=\mathrm{id},
\qquad
\rho\left(r_1,\dots,r_L,h^{(0),[T]}_1,h^{(L+1),[T]}_1\right)=r_2 .
\]
The edge update $\psi_u$ may be taken to be zero, since with $T=1$ its output is never read.
All of these can be computed by ReLU-MLPs, so the GMN is realizable.

The function computed by this GMN is
\[
f^\Theta(\theta)
=
\frac{1}{n_2}
\sum_{i=1}^{n_2}
\frac{1}{n_1}
\sum_{j=1}^{n_1}
\abs{\overline W^{(2)}_{ij}}.
\]
Every entry of $H_n$ has modulus one, so $f^\Theta(\theta_n)=1$ for every $n$ while $f^\Theta(0)=0$.
Thus $\theta_n\to0$ but $f^\Theta(\theta_n)\not\to f^\Theta(0)$.
So the GMN map $f^\Theta$ is not continuous with respect to the normalized spectral norm.
\end{proof}

\subsection{Spectral continuity of matrix-product GMNs}\label{sec:matprod-spectral}
Here we prove \Cref{prop:matprod}(i), the Lipschitz continuity of the matrix-product GMN with respect to the normalized spectral norm.

It is convenient to write the message passing in matrix form.
Let $H^{(\ell),[t]}\in\RR^{n_\ell\times d_h}$ be the matrix whose $i$th row is the node feature $h^{(\ell),[t]}_i$ after $t$ rounds.
Since the fan-in rescaling cancels the $1/n_{\ell-1}$ of the mean, the two mean aggregates of \eqref{eq:matprod} are the rows of
\begin{equation}
\label{eq:matprod-agg}
Z^{(\ell),\mathrm{in},[t]}
=
\Wl H^{(\ell-1),[t]}A_\ell^{\mathrm{in}},
\qquad
Z^{(\ell),\mathrm{out},[t]}
=
\frac{n_\ell}{n_{\ell+1}}
\left(W^{(\ell+1)}\right)^\top H^{(\ell+1),[t]}A_\ell^{\mathrm{out}},
\end{equation}
with the boundary conventions $Z^{(0),\mathrm{in},[t]}\colonequals0$ and $Z^{(L+1),\mathrm{out},[t]}\colonequals0$, and the node update applies the map $\psi_h$ to the rows of $H^{(\ell),[t]},Z^{(\ell),\mathrm{in},[t]},Z^{(\ell),\mathrm{out},[t]}$.
The forward variant is the case $A^{\mathrm{out}}_\ell=0$.
Note that, unlike the duplication-compatible GMN, this model carries no learnable edge features: the weights enter only through \eqref{eq:matprod-agg}.

\begin{proof}[Proof of \Cref{prop:matprod}(i)]
Since \Cref{sec:consistent-sequence-formulation} sets up the consistent sequence $(V_n,\msf G_n,\varphi_{N,n})$ with the compatible norms $\norm[V_n,\op]{\cdot}$, it is enough, by Proposition~C.2 of \citet{levin2025transferring}, to prove for each finite space $V_n$ a Lipschitz bound on $\{\theta\in V_n:\norm[V_n,\op]{\theta}\le R\}$ with a constant independent of the hidden widths $n_1,\dots,n_L$.

Let $A_\Theta\ge1$ be a common Lipschitz constant of the shared maps $\psi_v,\psi_h,\psi_r,\rho$, and let $a_\Theta\ge 1$ bound the spectral norms $\norm{A^{\mathrm{in}}_\ell},\norm{A^{\mathrm{out}}_\ell}$ for all $\ell$; both depend only on $\Theta$.
Fix $\theta,\widetilde\theta\in V_n$ with $\norm[V_n,\op]{\theta},\norm[V_n,\op]{\widetilde\theta}\le R$, decorate with a tilde every quantity computed from $\widetilde\theta$, and set
\[
M_t^{(\ell)}
\colonequals
\frac{\norm[\fro]{H^{(\ell),[t]}}}{\sqrt{n_\ell}},
\quad
\Delta_t^{(\ell)}
\colonequals
\frac{\norm[\fro]{H^{(\ell),[t]}-\widetilde H^{(\ell),[t]}}}{\sqrt{n_\ell}},
\quad
M_t\colonequals\sum_{\ell=0}^{L+1}M_t^{(\ell)},
\quad
\Delta_t\colonequals\sum_{\ell=0}^{L+1}\Delta_t^{(\ell)}.
\]
We prove by induction on the message-passing round $t$ that
\[M_t, \widetilde M_t \le K_t,\qquad
\Delta_t \le C_t \norm[V_n,\op]{\theta-\widetilde\theta}\]
for constants $C_t,K_t$ depending only on $\Theta,t,R,L,\din,\dout$ and not on the hidden widths.

We use two elementary facts.
First, the aggregates \eqref{eq:matprod-agg} obey
\begin{equation}
\label{eq:matprod-agg-bound}
\frac{\norm[\fro]{Z^{(\ell),\mathrm{in},[t]}}}{\sqrt{n_\ell}}
\le
a_\Theta\sqrt{\frac{n_{\ell-1}}{n_\ell}}\norm{\Wl}M_t^{(\ell-1)}
\le
a_\Theta RM_t^{(\ell-1)},
\qquad
\frac{\norm[\fro]{Z^{(\ell),\mathrm{out},[t]}}}{\sqrt{n_\ell}}
\le
a_\Theta RM_t^{(\ell+1)}.
\end{equation}
Second, if a map $\psi$ of $q$ arguments with Lipschitz constant $A$ is applied to the rows of matrices $X_1,\dots,X_q$ with $m$ rows each, then
\begin{equation}
\label{eq:rowwise-lip}
\begin{gathered}
\norm[\fro]{\psi(X_1,\dots,X_q)-\psi(Y_1,\dots,Y_q)}
\le
\left(A^2 \sum_{j=1}^q \norm[\fro]{X_j-Y_j}^2\right)^{1/2}
\le
A\sum_{j=1}^q\norm[\fro]{X_j-Y_j},
\\
\norm[\fro]{\psi(X_1,\dots,X_q)}
\le
\sqrt m\norm[2]{\psi(0)}+A\sum_{j=1}^q\norm[\fro]{X_j}.
\end{gathered}
\end{equation}

\emph{Base case ($t=0$).}
The input nodes carry a constant, so $H^{(0),[0]}=\widetilde H^{(0),[0]}$ is a fixed matrix with $\din$ rows, and $h^{(\ell),[0]}_i=\psi_v(b^{(\ell)}_i)$ for $\ell\ge1$.
Hence, by \eqref{eq:rowwise-lip},
\[
M_0^{(\ell)}
\le
\norm[2]{\psi_v(0)}+A_\Theta\frac{\norm[2]{b^{(\ell)}}}{\sqrt{n_\ell}}
\le
\norm[2]{\psi_v(0)}+A_\Theta R,
\qquad \ell\ge1,
\]
and $M_0^{(0)}$ is a constant depending only on $\Theta$ and $\din$, so $M_0,\widetilde M_0\le K_0$ for a constant $K_0$ depending only on $\Theta,R,L,\din,\dout$.
Likewise $\Delta_0^{(0)}=0$ and $\Delta_0^{(\ell)}\le A_\Theta\norm[2]{b^{(\ell)}-\widetilde b^{(\ell)}}/\sqrt{n_\ell}$ for $\ell\ge1$, whose sum over $\ell$ gives
$
\Delta_0\le C_0\norm[V_n,\op]{\theta-\widetilde\theta}$ where $C_0=A_\Theta$.

\emph{Induction step.}
Combining \eqref{eq:rowwise-lip} with \eqref{eq:matprod-agg-bound},
\[
M_{t+1}^{(\ell)}
\le
\norm[2]{\psi_h(0)}
+A_\Theta\left(M_t^{(\ell)}+a_\Theta RM_t^{(\ell-1)}+a_\Theta RM_t^{(\ell+1)}\right),
\]
with the conventions $M_t^{(-1)}=M_t^{(L+2)}=0$.
Summing over $\ell$ gives the scalar recurrence $M_{t+1}\le(L+2)\norm[2]{\psi_h(0)}+A_\Theta(1+2a_\Theta R)M_t$.
By induction, there exists a constant $K_{t+1}$ depending only on $\Theta,t+1,R,L,\din,\dout$ such that
\begin{equation*}
M_{t+1},\widetilde M_{t+1}\le K_{t+1}.
\end{equation*}

Splitting the difference of two forward aggregates into a weight part and a feature part,
\[
Z^{(\ell),\mathrm{in},[t]}-\widetilde Z^{(\ell),\mathrm{in},[t]}
=
\left(\Wl-\widetilde W^{(\ell)}\right)H^{(\ell-1),[t]}A^{\mathrm{in}}_\ell
+
\widetilde W^{(\ell)}\left(H^{(\ell-1),[t]}-\widetilde H^{(\ell-1),[t]}\right)A^{\mathrm{in}}_\ell,
\]
and estimating each term as in \eqref{eq:matprod-agg-bound}, using the induction bound on $M_t$ for the first and $\norm[V_n,\op]{\widetilde\theta}\le R$ on the second, gives
\[
\begin{split}
\frac{\norm[\fro]{Z^{(\ell),\mathrm{in},[t]}-\widetilde Z^{(\ell),\mathrm{in},[t]}}}{\sqrt{n_\ell}}
&\le
a_\Theta\left(
 K_t\sqrt{\tfrac{n_{\ell-1}}{n_\ell}}\norm{\Wl-\widetilde W^{(\ell)}}
+R\Delta_t^{(\ell-1)}
\right)
\\
&\le
a_\Theta\left(K_t\norm[V_n,\op]{\theta-\widetilde\theta}+R\Delta_t^{(\ell-1)}\right),
\end{split}
\]
The same computation on $(W^{(\ell+1)})^\top$ bounds the backward discrepancy by the same expression with $\Delta_t^{(\ell+1)}$ in place of $\Delta_t^{(\ell-1)}$.
Feeding these into \eqref{eq:rowwise-lip} for $\psi_h$ and summing over $\ell$,
\[
\Delta_{t+1}
\le
A_\Theta(1+2a_\Theta R)\Delta_t
+2(L+2)A_\Theta a_\Theta K_t\norm[V_n,\op]{\theta-\widetilde\theta}.
\]
By induction, $\Delta_{t+1}\le C_{t+1}\norm[V_n,\op]{\theta-\widetilde\theta}$ for a constant $C_{t+1}$ depending only on $\Theta,R,L,\din,\dout$ and not on the hidden widths.

\emph{Graph-level readout.}
For a hidden layer, Cauchy--Schwarz applied to the layer-wise mean $r_\ell=\frac{1}{n_\ell}\sum_{i\in[n_\ell]}\psi_r(h^{(\ell),[T]}_i)$ gives
\[
\norm[2]{r_\ell-\widetilde r_\ell}
\le
\frac{A_\Theta}{n_\ell}\sum_{i\in[n_\ell]}\norm[2]{h^{(\ell),[T]}_i-\widetilde h^{(\ell),[T]}_i}
\le
\frac{A_\Theta}{\sqrt{n_\ell}}\norm[\fro]{H^{(\ell),[T]}-\widetilde H^{(\ell),[T]}}
=
A_\Theta\Delta_T^{(\ell)}.
\]
The input and output node features enter $\rho$ coordinatewise, and the Euclidean norm of their concatenated discrepancy is $\sqrt{\din}\Delta_T^{(0)}+\sqrt{\dout}\Delta_T^{(L+1)}$ at most, with $\din,\dout$ fixed.
Since $\rho$ is $A_\Theta$-Lipschitz,
\[
\norm[2]{f^\Theta(\theta)-f^\Theta(\widetilde\theta)}
\le
C'_{\Theta,T,R}\Delta_T
\le
C''_{\Theta,T,R}\norm[V_n,\op]{\theta-\widetilde\theta},
\]
which is the required width-uniform Lipschitz bound on $V_n$.
\end{proof}

\subsection{Zero-padding compatibility}\label{sec:zero-padding-compatibility}
We discuss whether our proposed transferable GMNs are compatible with the zero-padding widening of \Cref{sec:zero-padding-equivalence}, and if not, whether they can be modified to be so.

\paragraph{Duplication-compatible GMNs.}
The duplication-compatible GMN cannot be made zero-padding compatible using a similar recipe.
Indeed, suppose the widened weight is
\[
\Wlup
=
\begin{pmatrix}
\Wl & B^{(\ell)}\\
0 & D^{(\ell)}
\end{pmatrix}
\]
with respect to $\RR^{N_{\ell-1}}=\RR^{n_{\ell-1}}\oplus\RR^{m_{\ell-1}}$ and $\RR^{N_\ell}=\RR^{n_\ell}\oplus\RR^{m_\ell}$, with $B^{(\ell)}$ and $D^{(\ell)}$ arbitrary (\Cref{sec:zero-padding-equivalence}).
Suppose also that the source-node features are zero-padded: the original nodes $v^{(\ell-1)}_j$, $j\in[n_{\ell-1}]$, keep the narrow feature $h^{(\ell-1),[t]}_j$, while the new nodes $v^{(\ell-1)}_a$, $a\in[m_{\ell-1}]$, carry $h^{(\ell-1),[t]}_a=0$.
For a forward edgewise message rule and a target node $v^{(\ell)}_i$ among the original ones, $i\in[n_\ell]$, the wide mean aggregate splits into a sum over the $n_{\ell-1}$ old edges and a sum over the $m_{\ell-1}$ new edges:
\[
\frac{1}{N_{\ell-1}}
\sum_{j=1}^{n_{\ell-1}}
\psi_m\left(N_{\ell-1}W^{(\ell)}_{ij},h^{(\ell-1),[t]}_j\right)
+
\frac{1}{N_{\ell-1}}
\sum_{a=1}^{m_{\ell-1}}
\psi_m\left(N_{\ell-1}B^{(\ell)}_{ia},0\right),
\]
whereas the narrow aggregate at the same node is
\[
\frac{1}{n_{\ell-1}}
\sum_{j=1}^{n_{\ell-1}}
\psi_m\left(n_{\ell-1}\Wl_{ij},h^{(\ell-1),[t]}_j\right).
\]
These two expressions need not agree in general.

\paragraph{Forward matrix-product GMNs, before readout.}
The forward matrix-product aggregation can be easily made compatible with the zero-padding widening of \Cref{sec:zero-padding-equivalence} at the level of message passing (before final readout).

If the source-node features are zero-padded, 
\[
H^{(\ell-1),[t]\uparrow}
=
\begin{pmatrix}
H^{(\ell-1),[t]}\\
0
\end{pmatrix},
\]
then \eqref{eq:matprod-agg} gives
\[
Z^{(\ell),\mathrm{in},[t]\uparrow}
=
\Wlup H^{(\ell-1),[t]\uparrow}A^{\mathrm{in}}_\ell
=
\begin{pmatrix}
\Wl H^{(\ell-1),[t]}A^{\mathrm{in}}_\ell\\
0
\end{pmatrix},
\]
so the original nodes receive the narrow aggregate and the padded ones receive zero.
Propagating the zero-padded relation through the rest of the forward model would additionally require the preprocessing and the node update to preserve zeros, for example $\psi_v(0)=0$ and $\psi_h(0,0)=0$ in the forward-only update.

\paragraph{Bidirectional matrix-product GMN, before readout.}
The bidirectional matrix-product aggregation is not compatible with the full zero-padding family in \Cref{sec:zero-padding-equivalence}.
Indeed, if the successor-node features are zero-padded, then the backward part of \eqref{eq:matprod-agg} becomes
\[
Z^{(\ell),\mathrm{out},[t]\uparrow}
=
\frac{N_\ell}{N_{\ell+1}}
\begin{pmatrix}
\left(W^{(\ell+1)}\right)^\top H^{(\ell+1),[t]}A^{\mathrm{out}}_\ell\\
\left(B^{(\ell+1)}\right)^\top H^{(\ell+1),[t]}A^{\mathrm{out}}_\ell
\end{pmatrix}.
\]
The original coordinates have the same normalization as the narrow backward aggregate only if $N_\ell/N_{\ell+1}=n_\ell/n_{\ell+1}$.
The padded coordinates vanish only under the additional condition that $\left(B^{(\ell+1)}\right)^\top H^{(\ell+1),[t]}=0$ for the relevant hidden states, for instance if $B^{(\ell+1)}=0$.
Thus the bidirectional matrix-product GMN can be made compatible only with a restricted zero-padding subclass, not with the full equivalence of \Cref{sec:zero-padding-equivalence}.

\paragraph{Graph-level readout.}
Even if the message-passing part preserved zero-padded hidden states (e.g. in the forward matrix-product GMN), the readout stage would still conflict with zero-padding, because zero-padding and duplication require different pooling normalization.
After applying the pointwise $\psi_r$, consider a pooling rule $P_n(x_1,\dots,x_n)=c_n\sum_{i\in[n]}x_i$.
Invariance under zero-padding alone forces $c_{n+m}=c_n$ for every $m$, i.e.\ $c_n$ is constant in $n$, which is exactly sum pooling.
Invariance under duplication alone forces $c_{kn}=c_n/k$, i.e.\ $c_n\propto 1/n$, which is exactly mean pooling.
A rule invariant under both must satisfy $c_n=c_n/k$ for every $k$, i.e.\ $c_n\equiv0$, so no nontrivial pooling rule realizes both at once.
We keep the mean because duplication is the structure we target for the reasons given in \Cref{sec:zero-padding-equivalence}, at the cost of readout compatibility with zero-padding.

\subsubsection{Lipschitz sensitivity of GMNs to their parameters.}
We isolate the parameter-sensitivity estimate used in the uniform-convergence step of \Cref{thm:mup-modified-gmn-ipm}, because it is largely a bookkeeping exercise and is merely for technical convenience.
Readers can feel free to skip this subsection on a first reading.

We work with the matrix-product GMN of \eqref{eq:matprod}, that is, the simple GMN \eqref{eq:gmn} of \Cref{sec:related-weightsym} carrying the three modifications of \Cref{sec:transferable_gmn} with the two message maps replaced by \eqref{eq:matprod}.
Its learnable parameter collects the parameters of the four shared MLPs $\psi_v,\psi_h,\psi_r,\rho$ together with the aggregation matrices of \eqref{eq:matprod},
\[
\Theta
=
\left(
\vartheta_v,\vartheta_h,\vartheta_r,\vartheta_\rho,
(A^{\mathrm{in}}_\ell,A^{\mathrm{out}}_\ell)_{\ell=0}^{L+1}
\right)
\in\mathcal A\subset\RR^q,
\]
where we assume $\mathcal{A}$ is compact.

We first prove that MLPs are Lipschitz in their parameters, uniformly over a compact parameter set.

\begin{lemma}
\label{lem:compact-mlp-sensitivity}
Let $\psi^\vartheta\colon\RR^{d}\to\RR^{d'}$ be an MLP with fixed architecture and Lipschitz activations.
For every compact parameter set $\mathcal A_\psi$, there is $A_{\mathcal A_\psi}\ge1$ such that, for all $\vartheta,\widetilde\vartheta\in\mathcal A_\psi$ and $u,\widetilde u\in\RR^{d}$,
\[
\norm[2]{\psi^\vartheta(u)-\psi^\vartheta(\widetilde u)}
\le
A_{\mathcal A_\psi}\norm[2]{u-\widetilde u},
\qquad
\norm[2]{\psi^\vartheta(u)-\psi^{\widetilde\vartheta}(u)}
\le
A_{\mathcal A_\psi}\left(1+\norm[2]{u}\right)\norm[2]{\vartheta-\widetilde\vartheta}.
\]
\end{lemma}

\begin{proof}
Following the notation of~\eqref{eq:mlp}, write $\psi^\vartheta$ as an $L$-layer MLP: with $a_\vartheta^{(0)}(u)=u$,
\[
x_\vartheta^{(\ell)}(u) = W_\vartheta^{(\ell)} a_\vartheta^{(\ell-1)}(u) + b_\vartheta^{(\ell)},
\qquad
a_\vartheta^{(\ell)}(u) = \sigma\left(x_\vartheta^{(\ell)}(u)\right),
\qquad \ell=1,\dots,L+1,
\]
where $\vartheta=\left((W_\vartheta^{(\ell)},b_\vartheta^{(\ell)})\right)_{\ell=1}^{L+1}$, and $\psi^\vartheta(u)=x_\vartheta^{(L+1)}(u)$.

Compactness of $\mathcal A_\psi$ gives a constant $B\ge1$ with $\norm{W_\vartheta^{(\ell)}}\le B$ and $\norm[2]{b_\vartheta^{(\ell)}}\le B$ for every layer $\ell$ and every $\vartheta\in\mathcal A_\psi$.
Let $\lambda\ge1$ be a Lipschitz constant of $\sigma$ and, by the triangle inequality, $\norm[2]{\sigma(z)}\le\lambda\norm[2]{z}+\norm[2]{\sigma(0)}$.

We prove by induction on $\ell$ that there are constants $C_\ell\ge1$, depending only on $\mathcal A_\psi$ and the architecture, such that for every $\ell=0,\dots,L$, all $\vartheta,\widetilde\vartheta\in\mathcal A_\psi$, and all $u,\widetilde u\in\RR^{d}$,
\[
\begin{aligned}
\norm[2]{a_\vartheta^{(\ell)}(u)}
&\le
C_\ell\left(1+\norm[2]{u}\right),
\\
\norm[2]{a_\vartheta^{(\ell)}(u)-a_\vartheta^{(\ell)}(\widetilde u)}
&\le
C_\ell\norm[2]{u-\widetilde u},
\\
\norm[2]{a_\vartheta^{(\ell)}(u)-a_{\widetilde\vartheta}^{(\ell)}(u)}
&\le
C_\ell\left(1+\norm[2]{u}\right)\norm[2]{\vartheta-\widetilde\vartheta}.
\end{aligned}
\]

\emph{Base case $\ell=0$.}
We have $a_\vartheta^{(0)}(u)=a_{\widetilde\vartheta}^{(0)}(u)=u$, so the three left-hand sides are $\norm[2]{u}$, $\norm[2]{u-\widetilde u}$ and $0$, and $C_0=1$ suffices.

\emph{Inductive step.}
Assume the three bounds at $\ell-1$.

The first follows from $\norm[2]{a_\vartheta^{(\ell)}(u)}\le\lambda\norm[2]{x_\vartheta^{(\ell)}(u)}+\norm[2]{\sigma(0)}$ together with $\norm[2]{x_\vartheta^{(\ell)}(u)}\le B\norm[2]{a_\vartheta^{(\ell-1)}(u)}+B$, and then the inductive growth bound on $a_\vartheta^{(\ell-1)}(u)$.

The second follows because $\norm[2]{a_\vartheta^{(\ell)}(u)-a_\vartheta^{(\ell)}(\widetilde u)}\le\lambda B\norm[2]{a_\vartheta^{(\ell-1)}(u)-a_\vartheta^{(\ell-1)}(\widetilde u)}$, to which the inductive Lipschitz bound applies.

For the third, we first write
\[
x_\vartheta^{(\ell)}(u)-x_{\widetilde\vartheta}^{(\ell)}(u)
=
W_\vartheta^{(\ell)}\left(a_\vartheta^{(\ell-1)}(u)-a_{\widetilde\vartheta}^{(\ell-1)}(u)\right)
+
\left(W_\vartheta^{(\ell)}-W_{\widetilde\vartheta}^{(\ell)}\right)a_{\widetilde\vartheta}^{(\ell-1)}(u)
+
\left(b_\vartheta^{(\ell)}-b_{\widetilde\vartheta}^{(\ell)}\right).
\]
The first term satisfies
\[
\norm[2]{W_\vartheta^{(\ell)}\left(a_\vartheta^{(\ell-1)}(u)-a_{\widetilde\vartheta}^{(\ell-1)}(u)\right)}
\le
B\norm[2]{a_\vartheta^{(\ell-1)}(u)-a_{\widetilde\vartheta}^{(\ell-1)}(u)}.
\]
The second term satisfies
\[
\norm[2]{\left(W_\vartheta^{(\ell)}-W_{\widetilde\vartheta}^{(\ell)}\right)a_{\widetilde\vartheta}^{(\ell-1)}(u)}
\le
\norm[2]{\vartheta-\widetilde\vartheta}\norm[2]{a_{\widetilde\vartheta}^{(\ell-1)}(u)}.
\]
Finally, the third term satisfies
\[
\norm[2]{b_\vartheta^{(\ell)}-b_{\widetilde\vartheta}^{(\ell)}}
\le
\norm[2]{\vartheta-\widetilde\vartheta}.
\]
Combining these three estimates and applying the inductive bounds gives the desired parameter-difference estimate for $x_\vartheta^{(\ell)}$.
Applying $\sigma$ costs another factor $\lambda$.

All three bounds therefore hold at $\ell$ with $C_\ell=\lambda\left((B+1)C_{\ell-1}+B+1\right)+\norm[2]{\sigma(0)}$, which completes the induction.

Finally, $\psi^\vartheta(u)=x_\vartheta^{(L+1)}(u)=W_\vartheta^{(L+1)}a_\vartheta^{(L)}(u)+b_\vartheta^{(L+1)}$, so the three estimates for the affine step above, applied at $\ell=L$ and without the activation, give the two claims of the lemma with $A_{\mathcal A_\psi}=(B+1)C_L+1$.
\end{proof}

We can now prove the parameter-sensitivity estimate for the matrix-product GMN, whose MLP components $\psi_v,\psi_h,\psi_r,\rho$ are controlled by \Cref{lem:compact-mlp-sensitivity}.

\begin{proposition}
\label{lem:matprod-param-sensitivity}
Assume that the MLP components $\psi_v,\psi_h,\psi_r,\rho$ of the matrix-product GMN have fixed architectures and Lipschitz activations.
Then, for every width $n$ and every $\theta_n\in V_n$,
\begin{equation}
\label{eq:matprod-param-lip-poly}
\norm[2]{f_n^\Theta(\theta_n)-f_n^{\widetilde\Theta}(\theta_n)}
\le
c_{\mathcal A,T}\left(1+\norm[V_n,\op]{\theta_n}\right)^{T+1}\norm[2]{\Theta-\widetilde\Theta},
\end{equation}
for all $\Theta,\widetilde\Theta\in\mathcal A$, where $c_{\mathcal A,T}$ is independent of the hidden widths.
\end{proposition}

\begin{proof}
We first record the uniform MLP estimates that will be used for every shared map.
By \Cref{lem:compact-mlp-sensitivity}, applied to the projections of the compact set $\mathcal A$ onto the parameter spaces of $\psi_v,\psi_h,\psi_r,\rho$, there is $A_{\mathcal A}\ge1$ such that, for every $\psi\in\{\psi_v,\psi_h,\psi_r,\rho\}$,
\begin{equation}
\label{eq:matprod-joint-lip}
\norm[2]{\psi^{\Theta}(u)-\psi^{\Theta}(\widetilde u)}
\le
A_{\mathcal A}\norm[2]{u-\widetilde u},
\qquad
\norm[2]{\psi^{\Theta}(u)-\psi^{\widetilde\Theta}(u)}
\le
A_{\mathcal A}\left(1+\norm[2]{u}\right)\norm[2]{\Theta-\widetilde\Theta},
\end{equation}
for all $\Theta,\widetilde\Theta\in\mathcal A$ and all inputs $u,\widetilde u$ of the corresponding MLP.
Here the norm of the MLP-parameter difference is bounded by $\norm[2]{\Theta-\widetilde\Theta}$ because each MLP parameter is a coordinate subvector of $\Theta$.

Fix a width $n$ and $\theta_n\in V_n$, and put
\[
R_n\colonequals\norm[V_n,\op]{\theta_n}.
\]
Fix $\Theta,\widetilde\Theta\in\mathcal A$, decorate with a tilde every quantity computed from $\widetilde\Theta$, and set
\[
M_t^{(\ell)}
\colonequals
\frac{\norm[\fro]{H^{(\ell),[t]}}}{\sqrt{n_\ell}},
\quad
\Delta_t^{(\ell)}
\colonequals
\frac{\norm[\fro]{H^{(\ell),[t]}-\widetilde H^{(\ell),[t]}}}{\sqrt{n_\ell}},
\quad
M_t\colonequals\sum_{\ell=0}^{L+1}M_t^{(\ell)},
\quad
\Delta_t\colonequals\sum_{\ell=0}^{L+1}\Delta_t^{(\ell)},
\]
with the conventions $M_t^{(-1)}=M_t^{(L+2)}=0$.
Let
\[
a_{\mathcal A}
\colonequals
1\vee\sup_{\Theta\in\mathcal A}\max_{0\le\ell\le L+1}
\max\left\{\norm{A^{\mathrm{in}}_\ell},\norm{A^{\mathrm{out}}_\ell}\right\},
\qquad
\kappa_{\mathcal A}
\colonequals
\max_{\psi\in\{\psi_v,\psi_h,\psi_r,\rho\}}\ \sup_{\Theta\in\mathcal A}\norm[2]{\psi^\Theta(0)},
\]
both finite because $\mathcal A$ is compact and, by \eqref{eq:matprod-joint-lip}, $\Theta\mapsto\psi^\Theta(0)$ is continuous.
Finally, since the normalized spectral norm \eqref{eq:norms} dominates each of its summands,
\begin{equation}
\label{eq:matprod-param-weight-bound}
\sqrt{\frac{n_{\ell-1}}{n_\ell}}\norm{\Wl}\le R_n,
\qquad
\frac{\norm[2]{b^{(\ell)}}}{\sqrt{n_\ell}}\le R_n
\qquad(1\le\ell\le L+1).
\end{equation}
We prove by induction on the round $t$ that
\begin{equation}
\label{eq:matprod-param-induction}
M_t,\widetilde M_t\le K_t,
\qquad
\Delta_t\le C_t\norm[2]{\Theta-\widetilde\Theta},
\end{equation}
for finite constants $K_t,C_t$ given by the explicit recursions \eqref{eq:matprod-param-K}, \eqref{eq:matprod-param-C} below, which depend only on $\mathcal A,t,L,\din,\dout,d_h$ and on $R_n$.

\emph{Row-wise application of MLP.}
Let $\psi\in\{\psi_v,\psi_h,\psi_r,\rho\}$ take $J$ arguments, and apply it to the rows of matrices $X_1,\dots,X_J$ with $m$ rows each.
Write $u_i\colonequals(X_{1,i:},\dots,X_{J,i:})$ and $v_i\colonequals(Y_{1,i:},\dots,Y_{J,i:})$ for the $i$th stacked arguments.
Since $\norm[2]{u_i}\le\sum_j\norm[2]{X_{j,i:}}$, the two bounds of \eqref{eq:matprod-joint-lip} hold row by row.
Stacking them, we have
\begin{align*}
\norm[\fro]{\psi^\Theta(X_1,\dots,X_J)}
&=\norm[2]{\left(\norm[2]{\psi^\Theta(u_i)}\right)_{i=1}^m}
\le\norm[2]{\left(\kappa_{\mathcal A}+A_{\mathcal A}\sum_{j=1}^J\norm[2]{X_{j,i:}}\right)_{i=1}^m}
\\
&\le\kappa_{\mathcal A}\sqrt m+A_{\mathcal A}\sum_{j=1}^J\norm[2]{\left(\norm[2]{X_{j,i:}}\right)_{i=1}^m}
=\kappa_{\mathcal A}\sqrt m+A_{\mathcal A}\sum_{j=1}^J\norm[\fro]{X_j},
\end{align*}
\begin{align*}
&\norm[\fro]{\psi^\Theta(X_1,\dots,X_J)-\psi^{\widetilde\Theta}(Y_1,\dots,Y_J)}
=\norm[2]{\left(\norm[2]{\psi^\Theta(u_i)-\psi^{\widetilde\Theta}(v_i)}\right)_{i=1}^m}
\\
&\quad\le A_{\mathcal A}\norm[2]{\left(\sum_{j=1}^J\norm[2]{X_{j,i:}-Y_{j,i:}}\right)_{i=1}^m}
+A_{\mathcal A}\norm[2]{\Theta-\widetilde\Theta}\norm[2]{\left(1+\sum_{j=1}^J\norm[2]{X_{j,i:}}\right)_{i=1}^m}
\\
&\quad\le A_{\mathcal A}\sum_{j=1}^J\norm[\fro]{X_j-Y_j}
+A_{\mathcal A}\left(\sqrt m+\sum_{j=1}^J\norm[\fro]{X_j}\right)\norm[2]{\Theta-\widetilde\Theta} .
\end{align*}
Dividing by $\sqrt m$ gives
\begin{equation}
\label{eq:matprod-param-rowwise}
\begin{gathered}
\frac{\norm[\fro]{\psi^\Theta(X_1,\dots,X_J)}}{\sqrt m}
\le
\kappa_{\mathcal A}
+A_{\mathcal A}\sum_{j=1}^J\frac{\norm[\fro]{X_j}}{\sqrt m},
\\
\frac{\norm[\fro]{\psi^\Theta(X_1,\dots,X_J)-\psi^{\widetilde\Theta}(Y_1,\dots,Y_J)}}{\sqrt m}
\\
\le
A_{\mathcal A}\sum_{j=1}^J\frac{\norm[\fro]{X_j-Y_j}}{\sqrt m}
+
A_{\mathcal A}\left(1+\sum_{j=1}^J\frac{\norm[\fro]{X_j}}{\sqrt m}\right)\norm[2]{\Theta-\widetilde\Theta}.
\end{gathered}
\end{equation}

\emph{Base case ($t=0$).}
The input nodes carry a constant that does not depend on $\Theta$, so $\Delta_0^{(0)}=0$ and $M_0^{(0)}=\widetilde M_0^{(0)}$ is a fixed number depending only on $\din$ and $d_h$.
For $\ell\ge1$ the initial features are $\psi_v^\Theta$ applied to the rows of $b^{(\ell)}$, so \eqref{eq:matprod-param-rowwise} and \eqref{eq:matprod-param-weight-bound} give
\[
M_0^{(\ell)}
\le
\kappa_{\mathcal A}+A_{\mathcal A}R_n,
\qquad
\Delta_0^{(\ell)}
\le
A_{\mathcal A}\left(1+R_n\right)\norm[2]{\Theta-\widetilde\Theta}.
\]
Summing over $\ell$ gives \eqref{eq:matprod-param-induction} at $t=0$ with
\[
K_0\colonequals M_0^{(0)}+(L+1)\left(\kappa_{\mathcal A}+A_{\mathcal A}R_n\right),
\qquad
C_0\colonequals (L+1)A_{\mathcal A}\left(1+R_n\right).
\]

\emph{Induction step: feature norms.}
The aggregate bound \eqref{eq:matprod-agg-bound}, with $R$ there replaced by $R_n$, reads
\begin{equation}
\label{eq:matprod-param-agg-bound}
\frac{\norm[\fro]{Z^{(\ell),\mathrm{in},[t]}}}{\sqrt{n_\ell}}
\le
a_{\mathcal A}R_nM_t^{(\ell-1)},
\qquad
\frac{\norm[\fro]{Z^{(\ell),\mathrm{out},[t]}}}{\sqrt{n_\ell}}
\le
a_{\mathcal A}R_nM_t^{(\ell+1)} .
\end{equation}
Feeding this into the first line of \eqref{eq:matprod-param-rowwise} for the node update $\psi_h$,
\[
M_{t+1}^{(\ell)}
\le
\kappa_{\mathcal A}
+A_{\mathcal A}\left(
M_t^{(\ell)}
+a_{\mathcal A}R_nM_t^{(\ell-1)}
+a_{\mathcal A}R_nM_t^{(\ell+1)}
\right),
\]
and summing over $\ell$ gives the scalar recurrence $M_{t+1}\le(L+2)\kappa_{\mathcal A}+A_{\mathcal A}(1+2a_{\mathcal A}R_n)M_t$, and likewise for $\widetilde M_{t+1}$.
So \eqref{eq:matprod-param-induction} holds at round $t+1$ with
\begin{equation}
\label{eq:matprod-param-K}
K_{t+1}
\colonequals
(L+2)\kappa_{\mathcal A}
+A_{\mathcal A}\left(1+2a_{\mathcal A}R_n\right)K_t .
\end{equation}

\emph{Induction step: feature differences.}
Splitting the difference of two forward aggregates into a parameter part and a feature part,
\[
\begin{split}
Z^{(\ell),\mathrm{in},[t]}-\widetilde Z^{(\ell),\mathrm{in},[t]}
={}&
\Wl\left(H^{(\ell-1),[t]}-\widetilde H^{(\ell-1),[t]}\right)A^{\mathrm{in}}_\ell(\Theta)
\\
&+
\Wl\widetilde H^{(\ell-1),[t]}\left(A^{\mathrm{in}}_\ell(\Theta)-A^{\mathrm{in}}_\ell(\widetilde\Theta)\right),
\end{split}
\]
and estimating each term as in \eqref{eq:matprod-param-agg-bound}, using for the second that the entries of $A^{\mathrm{in}}_\ell$ are coordinates of $\Theta$, so that $\norm{A^{\mathrm{in}}_\ell(\Theta)-A^{\mathrm{in}}_\ell(\widetilde\Theta)}\le\norm[2]{\Theta-\widetilde\Theta}$,
\[
\frac{\norm[\fro]{Z^{(\ell),\mathrm{in},[t]}-\widetilde Z^{(\ell),\mathrm{in},[t]}}}{\sqrt{n_\ell}}
\le
a_{\mathcal A}R_n\Delta_t^{(\ell-1)}
+
R_n\widetilde M_t^{(\ell-1)}\norm[2]{\Theta-\widetilde\Theta} .
\]
The same computation on $(W^{(\ell+1)})^\top$ bounds the backward discrepancy by the same expression with $\Delta_t^{(\ell+1)}$ and $\widetilde M_t^{(\ell+1)}$ in place of $\Delta_t^{(\ell-1)}$ and $\widetilde M_t^{(\ell-1)}$.
Feeding these and \eqref{eq:matprod-param-agg-bound} into the second line of \eqref{eq:matprod-param-rowwise} for $\psi_h$,
\[
\begin{split}
\Delta_{t+1}^{(\ell)}
\le{}&
A_{\mathcal A}\left(
\Delta_t^{(\ell)}
+a_{\mathcal A}R_n\left(\Delta_t^{(\ell-1)}+\Delta_t^{(\ell+1)}\right)
\right)
\\
&+A_{\mathcal A}\left(1+M_t^{(\ell)}\right)\norm[2]{\Theta-\widetilde\Theta}
\\
&+A_{\mathcal A}(1+a_{\mathcal A})R_n\left(M_t^{(\ell-1)}+M_t^{(\ell+1)}+\widetilde M_t^{(\ell-1)}+\widetilde M_t^{(\ell+1)}\right)\norm[2]{\Theta-\widetilde\Theta} .
\end{split}
\]
Summing over $\ell$, each of $\sum_\ell M_t^{(\ell\pm1)}$ and $\sum_\ell\widetilde M_t^{(\ell\pm1)}$ is at most $K_t$ by the induction hypothesis, so
\[
\Delta_{t+1}
\le
A_{\mathcal A}\left(1+2a_{\mathcal A}R_n\right)\Delta_t
+
A_{\mathcal A}\left((L+2)+\left(1+4(1+a_{\mathcal A})R_n\right)K_t\right)\norm[2]{\Theta-\widetilde\Theta},
\]
which is \eqref{eq:matprod-param-induction} at round $t+1$ with
\begin{equation}
\label{eq:matprod-param-C}
C_{t+1}
\colonequals
A_{\mathcal A}\left(1+2a_{\mathcal A}R_n\right)C_t
+
A_{\mathcal A}\left((L+2)+\left(1+4(1+a_{\mathcal A})R_n\right)K_t\right).
\end{equation}
This completes the induction, which we run up to the last round $t=T$.

\emph{Degree of the constants.}
Both \eqref{eq:matprod-param-K} and \eqref{eq:matprod-param-C} are affine recursions whose coefficients are bounded by a multiple of $1+R_n$, and $K_0,C_0\le c(1+R_n)$.
Hence, by induction on $t$, $K_t,C_t\le c_t(1+R_n)^{t+1}$ with $c_t$ depending only on $\mathcal A,t,L,\din,\dout,d_h$; in particular
\begin{equation}
\label{eq:matprod-param-KC-degree}
K_T,C_T\le c_T\left(1+R_n\right)^{T+1} .
\end{equation}

\emph{Readout.}
By the layer-wise mean readout of \Cref{sec:transferable_gmn}, the argument of $\rho^\Theta$ is
\[u_\Theta\colonequals\left(r_1,\dots,r_L,(h^{(0),[T]}_i)_{i\in[\din]},(h^{(L+1),[T]}_i)_{i\in[\dout]}\right),\quad
r_\ell=\frac{1}{n_\ell}\sum_{i\in[n_\ell]}\psi_r^\Theta\left(h^{(\ell),[T]}_i\right).\]
Applying \eqref{eq:matprod-joint-lip} to $\psi_r$ row by row and then Cauchy--Schwarz, $\sum_{i\in[n_\ell]}c_i\le\sqrt{n_\ell}\left(\sum_ic_i^2\right)^{1/2}$, gives for $1\le\ell\le L$
\[
\norm[2]{r_\ell}
\le
\kappa_{\mathcal A}+A_{\mathcal A}M_T^{(\ell)},
\qquad
\norm[2]{r_\ell-\widetilde r_\ell}
\le
A_{\mathcal A}\Delta_T^{(\ell)}
+A_{\mathcal A}\left(1+M_T^{(\ell)}\right)\norm[2]{\Theta-\widetilde\Theta} .
\]
The remaining arguments are listed coordinatewise, and $\norm[2]{(h^{(0),[T]}_i)_i}=\sqrt{\din}M_T^{(0)}$ and $\norm[2]{(h^{(0),[T]}_i)_i-(\widetilde h^{(0),[T]}_i)_i}=\sqrt{\din}\Delta_T^{(0)}$, and similarly at $\ell=L+1$ with $\dout$.
Summing the $L+2$ blocks and using \eqref{eq:matprod-param-induction},
\[
\norm[2]{u_\Theta}
\le
c\left(1+K_T\right),
\qquad
\norm[2]{u_\Theta-u_{\widetilde\Theta}}
\le
c\left(C_T+1+K_T\right)\norm[2]{\Theta-\widetilde\Theta},
\]
with $c$ depending only on $\mathcal A,L,\din,\dout$.
Applying \eqref{eq:matprod-joint-lip} to $\rho$ once more,
\[
\norm[2]{f_n^\Theta(\theta_n)-f_n^{\widetilde\Theta}(\theta_n)}
\le
A_{\mathcal A}\norm[2]{u_\Theta-u_{\widetilde\Theta}}
+A_{\mathcal A}\left(1+\norm[2]{u_\Theta}\right)\norm[2]{\Theta-\widetilde\Theta}
\le
B_n\norm[2]{\Theta-\widetilde\Theta},
\]
where $B_n\colonequals c'\left(1+K_T+C_T\right)$ is finite and, by \eqref{eq:matprod-param-KC-degree}, at most $c''(1+R_n)^{T+1}$.

\emph{Conclusion.}
Taking $c_{\mathcal A,T}=c''$ gives \eqref{eq:matprod-param-lip-poly}.
\end{proof}

\section{Implementations of transferable GMNs}
\label{app:gmns-variants}
The two recipes of \Cref{sec:transferable_gmn} were stated for the simple GMN defined in~\eqref{eq:gmn}.
This section applies them to three published GMN variants, whose design choices are more elaborate: the GMN of \citet{lim2023graph} (\Cref{app:gmn-lim}), the NG-GNN of \citet{kofinas2024graph} (\Cref{app:gmn-nggnn}), and the ScaleGMN of \citet{kalogeropoulos2024scale} (\Cref{app:gmn-scalegmn}).
We do not repeat the proofs of \Cref{prop:dup-invariance,prop:matprod} for each variant, because they follow the same ideas, as we explain below.

\paragraph{General recipe for duplication-compatible GMNs.}
The proof of \Cref{prop:dup-invariance} is an induction over the message-passing rounds, showing at each step that the node features, and respectively the edge features, on the widened network's graph are copies of the corresponding features on the narrow network's graph.

\emph{Initial node and edge features.}
The modification needed is to initialize the edge features with the fan-in rescaled weights $\Wlbar=n_{\ell-1}\Wl$ instead of the raw weights.
Further preprocessing of these scalars needs no modification as long as it is applied entrywise and shared across parameters of the same type.
Additional features not read off a single weight or bias entry have to be checked one at a time: we do so in the later sections for the positional encodings and for the probe features of the NG-GNN.

\emph{Message and update maps.}
These are shared across all nodes, or all edges, of a given type, and therefore generally need no modification: equal arguments fed to a shared map return equal outputs at every copy.
All that is required is that the only graph-dependent quantities they read are node, edge, and global features that already satisfy the inductive hypothesis.

\emph{Aggregation.}
This is the step that needs modification, being the only GMN component whose inputs may grow with the width.
An aggregation over a set of fixed cardinality is harmless; this covers the input and output layers, whose widths $n_0=\din$ and $n_{L+1}=\dout$ are fixed, and any combination over the $L+1$ layers.
An aggregation whose number of terms grows with the width has to be made (i) layer-wise and direction-wise, and (ii) invariant under repeating every element of the aggregated multiset $k$ times.
Requirement (ii) rules out the sum and admits the mean, the minimum, the maximum, and the population standard deviation.
Requirement (i) is needed because distinct layers may be widened by distinct multipliers $k_\ell$.

\emph{Graph-level readout.}
Similarly, the graph-level readout needs to be modified following the same requirement as the local aggregation stated above.

\paragraph{General recipe for matrix-product GMNs.}
On top of a duplication-compatible GMN, the matrix-product recipe requires an audit of every path through which the input weights or a width-growing collection can influence the computation.

Usually, the weight matrices $(\Wl)$ enter the computation of a GMN through initial edge features.
In this case, three modifications are needed.
First, drop all learnable hidden-edge features together with the hidden-edge update, so that a hidden weight edge carries only the fan-in rescaled scalar $\Wlbar_{ij}$, unchanged across rounds.
Second, for messages along a weight edge, restrict the message to multiplication of the weight scalar by a shared linear map of the source state alone, dropping any dependence on the destination state, on global features, and on learnable edge features.
Third, use layer-wise means for local aggregation and for any global update or readout involving a width-growing layer.
Together these restrictions put every appearance of the weight matrices in the form $\Wl H^{(\ell-1)}$, or, up to the backward normalization, $\left(W^{(\ell)}\right)^\top H^{(\ell)}$ in the backward direction, where $H^{(\ell)}$ has rows equal to the layer-$\ell$ node features.

More generally, any other GMN component that directly reads weight-matrix entries must also be modified so that the weight matrices enter only through these matrix products.
This single property is what both parts of \Cref{prop:matprod} use.
For compatibility, the induction of \Cref{prop:dup-invariance} now has to respect a block-wise widening \eqref{eq:blockwise}, whose blocks $\Wlup_{ij}$ are no longer constant.
If the backward aggregate is used, it respects only the block-wise widenings satisfying the column-balance condition of \Cref{prop:matprod}(ii).
For continuity, the induction propagates bounds on the width-normalized node features $\norm[\fro]{H^{(\ell),[t]}}/\sqrt{n_\ell}$ and on their discrepancies between two weights, and a matrix product is exactly what the normalized spectral norm controls in these terms:
\[
\frac{\norm[\fro]{\Wl X}}{\sqrt{n_\ell}}
\le
\sqrt{\frac{n_{\ell-1}}{n_\ell}}\norm{\Wl}\cdot\frac{\norm[\fro]{X}}{\sqrt{n_{\ell-1}}},
\]
the first factor on the right being the layer-$\ell$ weight term of $\norm[V_n,\op]{\cdot}$ in \eqref{eq:norms}, while shared row-wise Lipschitz maps and layer-wise means are Lipschitz in the same quantities.
For the continuity property, all shared maps must be Lipschitz with constants independent of the hidden widths, either globally or uniformly on the on a fixed-radius ball.
This holds for most common neural operations; normalization layers are the one case worth spelling out, and we do so where they occur, in \Cref{app:gmn-nggnn}.

\subsection{Modifications on the GMN of \texorpdfstring{\citet{lim2023graph}}{Lim et al.}}
\label{app:gmn-lim}

\paragraph{The backbone.}
The precise mathematical construction of the GMN of \citet{lim2023graph} on MLP is given in \citet{dayan2026expressive} (Definition D.10).
We follow their notation to describe the precise modifications needed.
The main departure from the simple GMN~\eqref{eq:gmn} is the following:
\begin{enumerate}[leftmargin=*,topsep=0pt,itemsep=0pt]
\item Biases are edges rather than node features: there is one bias node $\beta^{(\ell)}$ per layer, joined to every neuron of layer $\ell$ by an edge carrying $b^{(\ell)}_i$, so the edge set splits into weight edges and bias edges, but both are handled by the same message map, distinguished only by the parameter-type feature described below.
\item Nodes and edges carry additional initial features besides the weight or bias scalar. Additional node features include layer index and node type one-hot vectors, and additional edge features include layer index, direction, and parameter type one-hot vectors.
\item The model maintains a global feature $u$, and is updated at each round. 
\item The graph is undirected. Each edge has two different edge features, one for each direction, with the direction encoded as edge features.
Message passing occurs in both directions sharing the same message map.
\end{enumerate}

\paragraph{Duplication-compatible GMN.}
We now spell out the model obtained by applying the three modifications of \Cref{prop:dup-invariance} to this backbone.
We only state here the modifications needed on top of the formulation of Definition D.10 of \citet{dayan2026expressive}.

\emph{Graph.}
We need to modify the graph construction from an undirected graph to a directed one, since during message-passing, the messages in the forward and backward directions needed to be aggregated separately.

Concretely, the node set is $N=\bigcup_{\ell=0}^{L+1}\{v_i^{(\ell)}:i\in[n_\ell]\}\cup\bigcup_{\ell=1}^{L+1}\{\beta^{(\ell)}\}$, one neuron node per neuron and one bias node per layer, and the edge set is $E=E_{\mathrm w}\cup E_{\mathrm b}$ with
\[
\begin{gathered}
E_{\mathrm w}=\left\{(v_j^{(\ell-1)},v_i^{(\ell)}):\ell\in[L+1],i\in[n_\ell],j\in[n_{\ell-1}]\right\},
\\
E_{\mathrm b}=\left\{(\beta^{(\ell)},v_i^{(\ell)}):\ell\in[L+1],i\in[n_\ell]\right\}.
\end{gathered}
\]

\emph{Node and edge features.}
We modify the weight edge features to carry the fan-in rescaled weight $\Wlbar_{ij}=n_{\ell-1}\Wl_{ij}$; also, we remove the direction edge features, because now the directions are encoded in the graph structure.
The node features and global feature do not require any modification.
The layer-index and node-type one-hot features are admissible because they are compatible with duplication widening \eqref{eq:dup}: for example, all neurons of a hidden layer receive the same pair of one-hot vectors and hence unchanged when the input networks are widened.

Concretely, the edge features are modified to be
\[
e^{[0]}_{(v_j^{(\ell-1)},v_i^{(\ell)})}=\left(\Wlbar_{ij},\,\mathrm{layer}(\ell),\,\mathrm{ptype}(\mathrm w)\right),
\qquad
e^{[0]}_{(\beta^{(\ell)},v_i^{(\ell)})}=\left(b^{(\ell)}_i,\,\mathrm{layer}(\ell),\,\mathrm{ptype}(\mathrm b)\right).
\]

\emph{Message-passing layer update.}
In the update of node features, we modify the message passing to separately aggregate the forward and backward messages, and replace sum aggregation with mean aggregation.
The update of edge features is unchanged.
In the update of the global feature, we replace the sum over all nodes and edges with a mean over each layer separately.

Concretely, a layer is a tuple of MLPs $(\phi_m^{\rightarrow},\phi_m^{\leftarrow},\phi_h,\phi_e,\phi_u)$, shared across the graph, updating $(h^{[t]},e^{[t]},u^{[t]})\mapsto(h^{[t+1]},e^{[t+1]},u^{[t+1]})$ for $t=0,\dots,T-1$ by
\begin{align}
\label{eq:lim-bias-mean}
m^{[t]}_{v_j^{(\ell-1)}\to v_i^{(\ell)}}&=\phi_m^{\rightarrow}\left(h^{[t]}_{v_j^{(\ell-1)}},h^{[t]}_{v_i^{(\ell)}},e^{[t]}_{(v_j^{(\ell-1)},v_i^{(\ell)})},u^{[t]}\right),\notag\\
m^{[t]}_{v_i^{(\ell)}\leftarrow v_j^{(\ell+1)}}&=\phi_m^{\leftarrow}\left(h^{[t]}_{v_j^{(\ell+1)}},h^{[t]}_{v_i^{(\ell)}},e^{[t]}_{(v_i^{(\ell)},v_j^{(\ell+1)})},u^{[t]}\right),\notag\\
m^{[t]}_{\beta^{(\ell)}\to v_i^{(\ell)}}&=\phi_m^{\rightarrow}\left(h^{[t]}_{\beta^{(\ell)}},h^{[t]}_{v_i^{(\ell)}},e^{[t]}_{(\beta^{(\ell)},v_i^{(\ell)})},u^{[t]}\right),\notag\\
m^{[t]}_{\beta^{(\ell)}\leftarrow v_i^{(\ell)}}&=\phi_m^{\leftarrow}\left(h^{[t]}_{v_i^{(\ell)}},h^{[t]}_{\beta^{(\ell)}},e^{[t]}_{(\beta^{(\ell)},v_i^{(\ell)})},u^{[t]}\right),\notag\\
h^{[t+1]}_{v_i^{(\ell)}}&=\phi_h\Bigl(h^{[t]}_{v_i^{(\ell)}},\ \tfrac1{n_{\ell-1}}\!\sum_{j\in[n_{\ell-1}]}m^{[t]}_{v_j^{(\ell-1)}\to v_i^{(\ell)}},\notag\\
&\qquad\tfrac1{n_{\ell+1}}\!\sum_{j\in[n_{\ell+1}]}m^{[t]}_{v_i^{(\ell)}\leftarrow v_j^{(\ell+1)}},\ m^{[t]}_{\beta^{(\ell)}\to v_i^{(\ell)}},\ u^{[t]}\Bigr),\notag\\
h^{[t+1]}_{\beta^{(\ell)}}&=\phi_h\left(h^{[t]}_{\beta^{(\ell)}},\ \tfrac1{n_\ell}\!\sum_{i\in[n_\ell]}m^{[t]}_{\beta^{(\ell)}\leftarrow v_i^{(\ell)}},\ u^{[t]}\right),\notag\\
e^{[t+1]}_{ji}&=\phi_e\left(h^{[t]}_j,h^{[t]}_i,e^{[t]}_{ji},u^{[t]}\right),\notag\\
u^{[t+1]}&=\phi_u\Bigl(\sum_{\ell=0}^{L+1}\tfrac1{n_\ell}\!\sum_{i\in[n_\ell]}h^{[t]}_{v_i^{(\ell)}},\ \sum_{\ell=1}^{L+1} h^{[t]}_{\beta^{(\ell)}},\notag\\
&\qquad\sum_{\ell=1}^{L+1}\tfrac1{n_{\ell-1}n_\ell}\!\sum_{j,i}e^{[t]}_{(v_j^{(\ell-1)},v_i^{(\ell)})},\ \sum_{\ell=1}^{L+1}\tfrac1{n_\ell}\!\sum_{i\in[n_\ell]}e^{[t]}_{(\beta^{(\ell)},v_i^{(\ell)})},\ u^{[t]}\Bigr),
\end{align}
with the two sums in $h^{[t+1]}_{v_i^{(\ell)}}$ set to $0$ at $\ell=0$ and $\ell=L+1$ respectively, where the corresponding neighborhood is empty.
The two directions of message passing use different message maps, $\phi_m^{\rightarrow}$ along the edges and $\phi_m^{\leftarrow}$ against them, since the direction edge feature that distinguished them is gone.
Within each direction, however, we follow \citet{lim2023graph} and share one message map between the messages among neuron nodes and the messages between a bias node and its neuron nodes, and likewise share one update map $\phi_h$ between neuron nodes and bias nodes.
In principle these could instead be specialized per message type or per node type for extra expressiveness, at the cost of additional parameters.

\emph{Graph-level readout.}
The graph-level output is read off $u^{[T]}$ after the final round, optionally through a further learnable map; since $u$ is already assembled from layer-wise means summed above, no separate pooling stage is needed.

\paragraph{Matrix-product GMN.}
On top of the duplication-compatible model above, the second level lets each weight and each bias act on the adjacent node feature by a matrix product as in \eqref{eq:matprod}, and removes every remaining component that is neither linear in the weights of one layer nor row-wise in the node features.


\emph{Node and edge features.}
We drop all edge features beyond the raw scalars, together with the edge update $\phi_e$, so that a weight edge carries only $\Wlbar_{ij}$ and a bias edge only $b^{(\ell)}_i$, held fixed across rounds.
The node features and the global feature $u$ are left unchanged.

\emph{Message-passing layer update.}
We make the following changes to the MPNN update equations in~\eqref{eq:lim-bias-mean}.
First, as we have removed the edge features, the edge update $\phi_e$ is removed too, and the global update $\phi_u$ does not use edge features anymore.
Second, both message functions $\phi_m^{\rightarrow}$ and $\phi_m^{\leftarrow}$ are restricted to be linear in the edge scalar and in the source state alone, dropping their dependence on the destination state and on the global feature.

Concretely, $\phi_m^{\rightarrow}$ and $\phi_m^{\leftarrow}$ of \eqref{eq:lim-bias-mean} is restricted to the form below
\[
\phi_m^{\rightarrow}(h_{\mathrm{src}},h_{\mathrm{dst}},e,u)=eA^{\rightarrow}h_{\mathrm{src}},
\qquad
\phi_m^{\leftarrow}(h_{\mathrm{src}},h_{\mathrm{dst}},e,u)=eA^{\leftarrow}h_{\mathrm{src}},
\]
for learnable $A^{\rightarrow},A^{\leftarrow}\in\RR^{d_h\times d_h}$, shared across layers as in \eqref{eq:lim-bias-mean} above (unlike the per-layer $A^{\mathrm{in}}_\ell,A^{\mathrm{out}}_\ell$ of \eqref{eq:matprod}).
Explicitly, the four messages read
\begin{align}
\label{eq:lim-matprod-msg}
\begin{split}
m^{[t]}_{v_j^{(\ell-1)}\to v_i^{(\ell)}}&=\Wlbar_{ij}A^{\rightarrow}h^{[t]}_{v_j^{(\ell-1)}},
\qquad
m^{[t]}_{v_i^{(\ell)}\leftarrow v_j^{(\ell+1)}}=\overline W^{(\ell+1)}_{ji}A^{\leftarrow}h^{[t]}_{v_j^{(\ell+1)}},\\
m^{[t]}_{\beta^{(\ell)}\to v_i^{(\ell)}}&=b^{(\ell)}_iA^{\rightarrow}h^{[t]}_{\beta^{(\ell)}},
\qquad\ \,
m^{[t]}_{\beta^{(\ell)}\leftarrow v_i^{(\ell)}}=b^{(\ell)}_iA^{\leftarrow}h^{[t]}_{v_i^{(\ell)}}.
\end{split}
\end{align}
The global feature is updated by
\begin{align*}
u^{[t+1]}=\phi_u\left(\sum_{\ell=0}^{L+1}\tfrac1{n_\ell}\!\sum_{i\in[n_\ell]}h^{[t]}_{v_i^{(\ell)}},\ \sum_{\ell=1}^{L+1} h^{[t]}_{\beta^{(\ell)}},\ u^{[t]}\right).
\end{align*}

Equivalently this can be written in matrix form.
Write $H^{(\ell),[t]}\in\RR^{n_\ell\times d_h}$ for the matrix whose $i$th row is $h^{[t]}_{v^{(\ell)}_i}$.
The messages of \eqref{eq:lim-matprod-msg} can be written as matrix products:
\begin{align}
\label{eq:lim-matprod-mat}
\begin{split}
Z^{(\ell),\mathrm{in},[t]}&=\Wl H^{(\ell-1),[t]}A^{\rightarrow},
\qquad\ \
Z^{(\ell),\mathrm{out},[t]}=\frac{n_\ell}{n_{\ell+1}}\left(W^{(\ell+1)}\right)^\top H^{(\ell+1),[t]}A^{\leftarrow},\\
Z^{(\ell),\mathrm{b},[t]}&=b^{(\ell)}\left(A^{\rightarrow}h^{[t]}_{\beta^{(\ell)}}\right)^{\!\top},
\qquad\ \
z^{(\ell),\mathrm{b},[t]}=\frac{1}{n_\ell}\left(b^{(\ell)}\right)^\top H^{(\ell),[t]}A^{\leftarrow}.
\end{split}
\end{align}
The node and global updates are then, with $\phi_h$ applied row-wise,
\begin{align*}
\begin{split}
H^{(\ell),[t+1]}&=\phi_h\left(H^{(\ell),[t]},\ Z^{(\ell),\mathrm{in},[t]},\ Z^{(\ell),\mathrm{out},[t]},\ Z^{(\ell),\mathrm{b},[t]},\ u^{[t]}\right),\\
h^{[t+1]}_{\beta^{(\ell)}}&=\phi_h\left(h^{[t]}_{\beta^{(\ell)}},\ z^{(\ell),\mathrm{b},[t]},\ u^{[t]}\right),\\
u^{[t+1]}&=\phi_u\left(\sum_{\ell=0}^{L+1}\tfrac1{n_\ell}\one_{n_\ell}^\top H^{(\ell),[t]},\ \sum_{\ell=1}^{L+1} h^{[t]}_{\beta^{(\ell)}},\ u^{[t]}\right).
\end{split}
\end{align*}


\subsection{Modifications on the NG-GNN of \texorpdfstring{\citet{kofinas2024graph}}{Kofinas et al.}}
\label{app:gmn-nggnn}

\paragraph{The backbone.}
We now explicitly describe the NG-GNN of~\citet{kofinas2024graph} on MLP inputs.

\emph{Graph.}
The graph construction is exactly the same as the simple GMN in the main text.
That is, the node set is $N=\bigcup_{\ell=0}^{L+1}\{v_i^{(\ell)}:i\in[n_\ell]\}$ and the edge set is
$E=\left\{(v_j^{(\ell-1)},v_i^{(\ell)}):\ell\in[L+1],i\in[n_\ell],j\in[n_{\ell-1}]\right\}$.

\emph{Node and edge features.}
The raw node and edge features are standardized versions of the bias and weight scalars, respectively:
\begin{equation}
\label{eq:nggnn-raw-features}
\widetilde h_{v_i^{(\ell)}} =
\begin{cases}
0, & \ell=0,\\
\left(b_i^{(\ell)}-\mu_{b,\ell}\right)/s_{b,\ell}, & 1\leq \ell\leq L+1,
\end{cases}
\qquad
\widetilde e_{v_j^{(\ell-1)},v_i^{(\ell)}}=
\left(W_{ij}^{(\ell)}-\mu_{W,\ell}\right)/s_{W,\ell},
\end{equation}
where the mean $\mu_{W,\ell}$ and standard deviation $s_{W,\ell}$ are taken over all training dataset samples and all matrix entries of $W^{(\ell)}$, and likewise for $\mu_{b,\ell}$ and $s_{b,\ell}$.

Then the initial node and edge features are obtained by 
\begin{equation}
\label{eq:nggnn-preproc}
h_{v_i^{(\ell)}}^{[0]}=A_V\left(P_b(\widetilde h_{v_i^{(\ell)}})+r_{v_i^{(\ell)}}\right)+p_{\tau(v_i^{(\ell)})},
\qquad
e_{v_j^{(\ell-1)},v_i^{(\ell)}}^{[0]}=A_E P_W(\widetilde e_{v_j^{(\ell-1)},v_i^{(\ell)}}).
\end{equation}
Here $P_b:\RR\to\RR^{d_h}$ and $P_W:\RR\to\RR^{d_h}$ are either linear maps or Gaussian sinusoidal maps $x\mapsto C(\sin(Bx),\cos(Bx))$, with $B\in\RR^{m\times q_{\mathrm{in}}}$ Gaussian and $C\in\RR^{d_h\times 2m}$ learnable, where $q_{\mathrm{in}}=1$ here.
The maps $A_V,A_E\in\RR^{d_h\times d_h}$ are linear maps.
The vector $p_{\tau(v)}\in\RR^{d_h}$ is a learnable position encoding, with $\tau(v)$ distinguishes input coordinates, hidden layers, and output coordinates:
all nodes in a hidden layer $\ell\in\{1,\dots,L\}$ share one type ID, while the input nodes $v_i^{(0)}$ and output nodes $v_i^{(L+1)}$ each have coordinate-specific type IDs.

If $q$ learnable probe inputs are used, then the probe feature $r_v$ is obtained by taking the vector in $\RR^q$ of activations of node $v$ across those probes and applying a layer-specific linear map $\RR^q\to\RR^{d_h}$ followed by layer normalization.


\emph{Message-passing layer update.}
The message from node $v_j^{(\ell-1)}$ to node $v_i^{(\ell)}$ is computed in the following two possible ways.
In the FiLM variant, the edge features are used to produce scale and shift vectors applied on the messages:
\begin{equation}
\label{eq:nggnn-film}
\begin{split}
(\gamma_{v_j^{(\ell-1)},v_i^{(\ell)}}^{[t]},\beta_{v_j^{(\ell-1)},v_i^{(\ell)}}^{[t]})&=A_{\mathrm{film}}e_{v_j^{(\ell-1)},v_i^{(\ell)}}^{[t]},\\
m_{v_j^{(\ell-1)}\to v_i^{(\ell)}}^{[t]}&=
\gamma_{v_j^{(\ell-1)},v_i^{(\ell)}}^{[t]}\odot \eta_t\left(h_{v_i^{(\ell)}}^{[t]},h_{v_j^{(\ell-1)}}^{[t]}\right)+\beta_{v_j^{(\ell-1)},v_i^{(\ell)}}^{[t]},
\end{split}
\end{equation}
where $A_{\mathrm{film}}\in\RR^{2d_h\times d_h}$ is a linear map and
$\eta_t:\RR^{2d_h}\to\RR^{d_h}$ is an MLP.
In the non-FiLM variant, we instead use
\[m_{v_j^{(\ell-1)}\to v_i^{(\ell)}}^{[t]}=\eta_t\left(h_{v_i^{(\ell)}}^{[t]},h_{v_j^{(\ell-1)}}^{[t]},A_{\mathrm{edge}}e_{v_j^{(\ell-1)},v_i^{(\ell)}}^{[t]}\right),\]
where $A_{\mathrm{edge}}\in\RR^{d_h\times d_h}$ is a linear map and $\eta_t:\RR^{3d_h}\to\RR^{d_h}$ is an MLP.

The node features are updated by
\begin{equation}
\label{eq:nggnn-layer}
 h_{v_i^{(\ell)}}^{[t+1]}=\Gamma_t\left(h_{v_i^{(\ell)}}^{[t]},\AGGPNA\left(\{m_{v_j^{(\ell-1)}\to v_i^{(\ell)}}^{[t]}:j\in[n_{\ell-1}]\}\right)\right),
\qquad \ell\in[L+1],
\end{equation}
with the aggregate taken to be empty at $\ell=0$, since $v_i^{(0)}$ has no in-neighbors; $\AGGPNA$ is the PNA aggregation operator of \citet{corso2020principal}, which combines multiple different permutation-invariant aggregators, and $\Gamma_t$ is an MLP with layer normalization and dropout.

The edge features are updated as
\begin{equation}
\label{eq:nggnn-edge-update}
e_{v_j^{(\ell-1)},v_i^{(\ell)}}^{[t+1]}
=A_o^{\mathrm{upd}}\sigma
\left(
A_e^{\mathrm{upd}}e_{v_j^{(\ell-1)},v_i^{(\ell)}}^{[t]}
+A_s^{\mathrm{upd}}h_{v_j^{(\ell-1)}}^{[t+1]}
+A_t^{\mathrm{upd}}h_{v_i^{(\ell)}}^{[t+1]}\right),
\end{equation}
where $A_o^{\mathrm{upd}},A_e^{\mathrm{upd}},A_s^{\mathrm{upd}},A_t^{\mathrm{upd}}\in\RR^{d_h\times d_h}$ are linear maps, and $\sigma$ is an activation.

\emph{Graph-level readout.}
The final readout is an MLP $\rho$ applied to the concatenation of the output-layer node features:
\begin{equation}
\label{eq:nggnn-readout}
f^\Theta(\theta)
=
\rho\left(
\left(h_{v_i^{(L+1)}}^{[T]}\right)_{i\in[n_{L+1}]}
\right).
\end{equation}

We summarize the main departure from the simple GMN~\eqref{eq:gmn}:
\begin{enumerate}[leftmargin=*,topsep=0pt,itemsep=0pt]
\item The node and edge features use standardization on biases and weights, and optional sinusoidal encodings, positional encodings, and probe features.
\item The message functions take special forms that are either FiLM or non-FiLM.
\item The message passing aggregation is PNA rather than simple sum aggregation.
\item The readout uses only the last-layer node features, rather than all node features.
\end{enumerate}

\paragraph{Duplication-compatible NG-GNN.}
The modifications needed to make NG-GNN duplication-compatible are the following:

\emph{Node and edge features.}
We modify the raw weight edge feature of \eqref{eq:nggnn-raw-features} to standardize the fan-in rescaled weight $\Wlbar_{ij}=n_{\ell-1}\Wl_{ij}$ rather than $\Wl_{ij}$.
The standardization itself is admissible, being entrywise affine with per-layer constants.

The positional encoding $p_{\tau(v)}$ of \eqref{eq:nggnn-preproc} needs no modification because they are compatible with duplication widening \eqref{eq:dup}: all neurons of a hidden layer share one ID, while the coordinate-specific IDs occur only at the input and output layers, whose widths $n_0$ and $n_{L+1}$ do not grow.

The probe features need no modification either, since under the duplication widening \eqref{eq:dup} the widened activations are $\alup=a^{(\ell)}\otimes\one_{k_\ell}$, so all copies of a neuron carry the same activation at each of the $q$ probe inputs.

\emph{Message-passing layer update.}
We replace the PNA aggregation of \eqref{eq:nggnn-layer} by the restriction $\AGGPNA^{\mathrm{dup}}$ that keeps only the aggregators invariant under repeating every element of the aggregated multiset $k$ times --- the mean, min, max, and standard deviation, but not the sum --- and that keeps only the identity degree scaler of \citet{corso2020principal}, since other scalers are width-dependent.

Dropout can break duplication-compatibility during training, but compatibility holds at evaluation time, when dropout is turned off.

\emph{Graph-level readout.}
The readout \eqref{eq:nggnn-readout} is already duplication-compatible, since it reads only the output-layer node features and $n_{L+1}$ does not grow.

\paragraph{Matrix-product NG-GNN.}
The further modifications needed to make the matrix-product NG-GNN are also almost the same as those for the simple GMN.

\emph{Node and edge features.}
We drop the edge features altogether, together with the edge update \eqref{eq:nggnn-edge-update}, so that what is left on an edge is the raw scalar $\Wlbar_{ij}$, held fixed across rounds.
In particular the map $P_W$ of \eqref{eq:nggnn-preproc} has to be removed; the standardization of \eqref{eq:nggnn-raw-features} is entrywise affine with per-layer constants and could be kept, but we drop it as well and use the raw scalar.

The probe features can also be kept.
Fix the $q$ probe inputs and let $X^{(\ell)}\in\RR^{n_\ell\times q}$ be the matrix holding the layer-$\ell$ activations of the input network, one row per neuron and one column per probe input.
The ordinary forward pass computes them by
\[
X^{(\ell)}
=
\sigma\left(
\frac{1}{n_{\ell-1}}\Wlbar X^{(\ell-1)}
+b^{(\ell)}\one_q^\top
\right),
\qquad 1\le \ell\le L,
\]
and by the same formula without the nonlinearity at $\ell=L+1$.
So each weight matrix acts only by multiplying the previous activation matrix, and each bias only through the rank-one term $b^{(\ell)}\one_q^\top$, which is the matrix-product form we need.
The map that turns a row of $X^{(\ell)}$ into the feature $r_v$ is shared across nodes, so it is admissible.
In particular, the layer normalization inside this map can be kept unmodified, since the normalization is taken over the hidden dimension $d_h$ which does not grow with the width.
When regularized by an $\varepsilon$ inside the square root, it is also Lipschitz with a constant depending on its learned scale and shift $\varepsilon$ but not on the hidden width.

\emph{Message-passing layer update.}
We restrict $\AGGPNA^{\mathrm{dup}}$ further to the mean aggregator alone, the only one that is linear in the aggregated messages, and we restrict the message to
\[
m^{[t]}_{v_j^{(\ell-1)}\to v_i^{(\ell)}}=\Wlbar_{ij}A^{\mathrm{in}}_\ell h^{[t]}_{v_j^{(\ell-1)}},
\]
which specializes both \eqref{eq:nggnn-film} and its non-FiLM counterpart.
Equivalently, this can be written in matrix form.
Writing $H^{(\ell),[t]}\in\RR^{n_\ell\times d_h}$ for the matrix whose $i$th row is $h^{[t]}_{v^{(\ell)}_i}$, the mean aggregate and the node update are
\begin{equation*}
Z^{(\ell),\mathrm{in},[t]}=\Wl H^{(\ell-1),[t]}A^{\mathrm{in}}_\ell,
\qquad
H^{(\ell),[t+1]}=\Gamma_t\left(H^{(\ell),[t]},Z^{(\ell),\mathrm{in},[t]}\right),
\end{equation*}
with $\Gamma_t$ applied row-wise.

The layer normalization applied here needs to be modified.
In the original implementation, this normalization is not applied separately to each row in $\RR^{d_h}$; its mean and variance are computed from a width-growing collection of node features.
After the aggregate $Z^{(\ell),\mathrm{in},[t]}=\Wl H^{(\ell-1),[t]}A^{\mathrm{in}}_\ell$, those statistics include global averages and squared terms formed from $Z^{(\ell),\mathrm{in},[t]}$.
These terms are nonlinear functions of $\Wl$, so the weights no longer enter only by matrix multiplication.
We therefore replace this operation with layer normalization over the fixed dimension $d_h$, or omit it.

\subsection{Modifications on the ScaleGMN of \texorpdfstring{\citet{kalogeropoulos2024scale}}{Kalogeropoulos et al.}}
\label{app:gmn-scalegmn}

\paragraph{The backbone.}
The ScaleGMN of \citet{kalogeropoulos2024scale} further imposes scaling symmetries on top of the permutation symmetries.
We now explicitly describe the ScaleGMN construction on MLP inputs, and we follow the implementation of ScaleGMN in their experiments, since we will also use it in our experiments.

\emph{Graph.}
The graph construction is exactly the same as the simple GMN in the main text. 
That is, a directed graph with node set $N=\bigcup_{\ell=0}^{L+1}\{v_i^{(\ell)}:i\in[n_\ell]\}$ and the edge set $E=\left\{(v_j^{(\ell-1)},v_i^{(\ell)}):\ell\in[L+1],i\in[n_\ell],j\in[n_{\ell-1}]\right\}$.

\emph{Node and edge features.}
Node and edge features are initialized from the biases and weight entries as in the simple GMN, except that the scalars are first projected to the hidden dimension $d_h$ and then combined with learnable positional encoding in a way that preserves the intended scaling behavior:
\begin{equation}
\label{eq:scalegmn-preproc}
h_{v_i^{(\ell)}}^{[0]} =
\begin{cases}
W_V 1 + p_{\tau(v_i^{(0)})}, & \ell = 0,\\
\left(W_V b^{(\ell)}_i\right) \odot p_{\tau(v_i^{(\ell)})}, & 1 \le \ell \le L,\\
W_V b^{(L+1)}_i + p_{\tau(v_i^{(L+1)})}, & \ell=L+1,
\end{cases}
\qquad
e_{v_j^{(\ell-1)}, v_i^{(\ell)}}^{[0]} = \left(W_E W^{(\ell)}_{ij}\right) \odot q_{\tau(v_j^{(\ell-1)}, v_i^{(\ell)})},
\end{equation}
where $W_V, W_E \in \RR^{d_h}$ are learnable bias-free linear maps $\RR \to \RR^{d_h}$ (written as a single column vector ), and $p_{\tau(v)}, q_{\tau(u,v)} \in \RR^{d_h}$ are learnable embeddings indexed by node/edge type IDs $\tau(v)$ and $\tau(u,v)$.
Concretely, $\tau(v)$ distinguishes input coordinates, hidden layers, and output coordinates:
all nodes in a hidden layer $\ell\in\{1,\dots,L\}$ share one type ID, while the input nodes $v_i^{(0)}$ and output nodes $v_i^{(L+1)}$ each have coordinate-specific type IDs.
Likewise, $\tau(u,v)$ depends only on the layer pair and on whether the edge is attached to the input or output boundary:
edges $(v_j^{(0)}, v_i^{(1)})$ are typed by the input coordinate $j$, interior hidden-to-hidden edges $(v_j^{(\ell-1)}, v_i^{(\ell)})$ with $2 \le \ell \le L$ share one type ID per layer pair $(\ell-1,\ell)$, and final-layer edges $(v_j^{(L)}, v_i^{(L+1)})$ are typed by the output coordinate $i$.

The bidirectional variant of ScaleGMN, where message passing also occurs in the backward direction on the edges, uses a different backward edge feature for positive-scale symmetry.
In the formulation of \citet{kalogeropoulos2024scale}, the backward edge feature for the edge from $v_j^{(\ell-1)}$ to $v_i^{(\ell)}$ is initialized as
\begin{equation}
\label{eq:scalegmn-bw-preproc}
e_{v_j^{(\ell-1)},v_i^{(\ell)}}^{\leftarrow,[0]}
=
\left(W_E^{\leftarrow}\left(1/W^{(\ell)}_{ij}\right)\right)
\odot q_{\tau(v_j^{(\ell-1)},v_i^{(\ell)})}^{\leftarrow}.
\end{equation}
For sign symmetry this is unnecessary: the implementation can share the same edge feature as the forward direction, just like in the simple GMN.

\emph{Message-passing layer update.}
For $t = 0, \dots, T-1$, a layer sends a forward message along each edge and a backward message against it, aggregates the two separately, and updates the node feature,
\begin{align*}
\begin{split}
    m_{v_j^{(\ell-1)} \to v_i^{(\ell)}}^{[t]}
    &=
    \Psi_m\left(
    \left(A_e e_{v_j^{(\ell-1)}, v_i^{(\ell)}}^{[t]}\right)
    \odot
    \left(A_v h_{v_j^{(\ell-1)}}^{[t]}\right)
    \right)
    \in \mathbb{R}^{d_h}, \\
    m_{v_i^{(\ell)} \leftarrow v_j^{(\ell+1)}}^{[t]}
    &=
    \Psi_m^{\leftarrow}\left(
    \left(A_e^{\leftarrow} e_{v_i^{(\ell)}, v_j^{(\ell+1)}}^{\leftarrow,[t]}\right)
    \odot
    \left(A_v^{\leftarrow} h_{v_j^{(\ell+1)}}^{[t]}\right)
    \right)
    \in \mathbb{R}^{d_h}, \\
    \widetilde h_{v_i^{(\ell)}}^{[t+1]}
    &=
    \Psi_h\left(
    h_{v_i^{(\ell)}}^{[t]},\;
    \sum_{j \in [n_{\ell-1}]}
    m_{v_j^{(\ell-1)} \to v_i^{(\ell)}}^{[t]},\;
    \sum_{j \in [n_{\ell+1}]}
    m_{v_i^{(\ell)} \leftarrow v_j^{(\ell+1)}}^{[t]}
    \right)
    \in \mathbb{R}^{d_h},
\end{split}
\end{align*}
where $A_e, A_v, A_e^{\leftarrow}, A_v^{\leftarrow} \in \mathbb{R}^{d_h \times d_h}$ are learnable bias-free linear maps, written as left matrix multiplication on the column vectors $e^{[t]}, h^{[t]} \in \mathbb{R}^{d_h}$, and the two aggregates are set to $0$ at $\ell=0$ and $\ell=L+1$ respectively, where the corresponding neighborhood is empty.
The maps $\Psi_m,\Psi_m^{\leftarrow}$ and $\Psi_h$ are the implementation's \texttt{EquivariantNet} blocks for the non-permutation symmetry modes: on hidden nodes they apply equivariant transformations for the selected symmetry, while on the fixed input/output nodes they may fall back to ordinary MLP branches.
A learnable residual skip connection is then applied
\[
h_v^{[t+1]} = S_t h_v^{[t]} + \widetilde h_v^{[t+1]},
\]
with $S_t \in \mathbb{R}^{d_h \times d_h}$ learnable.
This is then followed by entrywise dropout on the node feature matrix.

Both edge features are updated after each GNN layer except the last:
\begin{align}
\label{eq:scalegmn-edge-upd}
\begin{split}
e_{v_j^{(\ell-1)}, v_i^{(\ell)}}^{[t+1]}
&=
\underbrace{A_E^{\mathrm{upd}} e_{v_j^{(\ell-1)}, v_i^{(\ell)}}^{[t]}}_{\in \mathbb{R}^{d_h}}
\odot
\underbrace{\phi_E\left(
\mathrm{inv}\left(h_{v_j^{(\ell-1)}}^{[t]}\right),\;
\mathrm{inv}\left(h_{v_i^{(\ell)}}^{[t]}\right)
\right)}_{\in \mathbb{R}^{d_h}},\\
e_{v_i^{(\ell)}, v_j^{(\ell+1)}}^{\leftarrow,[t+1]}
&=
\underbrace{A_E^{\leftarrow,\mathrm{upd}} e_{v_i^{(\ell)}, v_j^{(\ell+1)}}^{\leftarrow,[t]}}_{\in \mathbb{R}^{d_h}}
\odot
\underbrace{\phi_E^{\leftarrow}\left(
\mathrm{inv}^{\leftarrow}\left(h_{v_i^{(\ell)}}^{[t]}\right),\;
\mathrm{inv}^{\leftarrow}\left(h_{v_j^{(\ell+1)}}^{[t]}\right)
\right)}_{\in \mathbb{R}^{d_h}},
\end{split}
\end{align}
where $A_E^{\mathrm{upd}}, A_E^{\leftarrow,\mathrm{upd}} \in \mathbb{R}^{d_h \times d_h}$ are learnable bias-free linear maps, $\mathrm{inv}$ is the invariant canonicalization used by the implementation's \texttt{InvariantLayer}, and $\phi_E,\phi_E^{\leftarrow}$ are MLPs, again with independent parameters in the two directions.

Just like simple GMN, this defines the \emph{bidirectional ScaleGMN}, and the \emph{forward ScaleGMN} is the restriction $\Psi_m^{\leftarrow}=0$, in which the edge features \eqref{eq:scalegmn-bw-preproc}, the message parameters $A_e^{\leftarrow},A_v^{\leftarrow}$, and the backward edge update in \eqref{eq:scalegmn-edge-upd} are absent.

\emph{Graph-level readout.}
For the implementation's \texttt{full\_graph} readout, the map has the form
\begin{equation*}
f^{\Theta}(\theta)
=
\rho\left(
\sum_{\ell=1}^{L}\sum_{i\in[n_\ell]} g_{\mathrm{hid}}\left(h_{v_i^{(\ell)}}^{[T]}\right),
\;\left(h_{v_i^{(0)}}^{[T]}\right)_{i\in[n_0]},
\;\left(h_{v_i^{(L+1)}}^{[T]}\right)_{i\in[n_{L+1}]}
\right),
\end{equation*}
where $g_{\mathrm{hid}}$ is the invariant readout block for the selected symmetry and $\rho$ is the final readout MLP.

A different readout, named \texttt{last\_layer} in the implementation, reads only the ordered boundary-node features.
For the forward and bidirectional variants, it takes the following different forms
\begin{align*}
f^{\Theta,\mathrm{last}}_{\mathrm{fw}}(\theta)
&=
\rho\left(\left(h_{v_i^{(L+1)}}^{[T]}\right)_{i\in[n_{L+1}]}\right),
\\
f^{\Theta,\mathrm{last}}_{\mathrm{bidir}}(\theta)
&=
\rho\left(
W_{\mathrm{last}}\left(h_{v_i^{(L+1)}}^{[T]}\right)_{i\in[n_{L+1}]} + b_{\mathrm{last}},
\;
W_{\mathrm{first}}\left(h_{v_i^{(0)}}^{[T]}\right)_{i\in[n_0]} + b_{\mathrm{first}}
\right),
\end{align*}
where $W_{\mathrm{last}}\in\RR^{d_h\times n_{L+1} d_h}$, $W_{\mathrm{first}}\in\RR^{d_h\times n_0 d_h}$ (with matching biases) are learnable linear maps that project them down to $\RR^{d_h}$ before the final MLP $\rho$.

We summarize the main departure from the simple GMN~\eqref{eq:gmn}:
\begin{enumerate}[leftmargin=*,topsep=0pt,itemsep=0pt]
\item Positional encodings are incorporated into the initial node and edge features, in a way that imposes the scaling equivariance.
\item The message and node-update maps take special forms to impose the scaling equivariance.
\item Residual connections and dropouts are used.
\item The bidirectional version on positive-scaling symmetries requires separate backward edge features, which should be initialized with the reciprocal weights.
\end{enumerate}

\paragraph{Duplication-compatible ScaleGMN.}
The modifications needed to make ScaleGMN duplication-compatible are almost the same as those for the simple GMN.

\emph{Node and edge features.}
We modify the raw weight edge features to carry the fan-in rescaled weight $\Wlbar_{ij}=n_{\ell-1}\Wl_{ij}$, and, in the bidirectional variant, the backward edge features to carry its reciprocal $1/\Wlbar_{ij}$.
This does not affect scale equivariance: the fan-in factor $n_{\ell-1}$ is a fixed positive constant, which commutes with the scaling action on the weights.

The type embeddings $p_{\tau(v)}$, $q_{\tau(u,v)}$, and $q^{\leftarrow}_{\tau(u,v)}$ of \eqref{eq:scalegmn-preproc} and \eqref{eq:scalegmn-bw-preproc} need no modification because they are compatible with duplication widening: all neurons of a hidden layer share one ID and all interior hidden-to-hidden edges share one ID per layer pair, while the coordinate-specific IDs occur only at the input and output boundaries, whose widths $n_0$ and $n_{L+1}$ do not grow.

\emph{Message-passing layer update.}
We replace the sum aggregation in the node-feature update with mean aggregation.
Concretely, the update equation is changed to
\begin{align}
\label{eq:scalegmn-impl-layer}
\begin{split}
\widetilde h^{[t+1]}_{v^{(\ell)}_i}&=\Psi_h\left(h^{[t]}_{v^{(\ell)}_i},\ \tfrac1{n_{\ell-1}}\!\sum_{j\in[n_{\ell-1}]}m^{[t]}_{j\to i},\ \tfrac1{n_{\ell+1}}\!\sum_{j\in[n_{\ell+1}]}m^{[t]}_{i\leftarrow j}\right).
\end{split}
\end{align}
Again this does not affect scale equivariance.
Dropout can break duplication-compatibility during training, but compatibility holds at evaluation time, when dropout is turned off.

\emph{Graph-level readout.}
The \texttt{last\_layer} readout is already duplication-compatible.
The \texttt{full\_graph} readout is not, but it can be modified into a layer-wise mean readout, just as in the main text.
Specifically, the layer-wise mean readout can be of the form
\[f^{\Theta, \mathrm{layer}}(\theta)
=
\rho\left(
\sum_{\ell=1}^{L}\frac{1}{n_\ell}\sum_{i\in[n_\ell]} g_{\mathrm{hid}}\left(h_{v_i^{(\ell)}}^{[T]}\right),
\;\left(h_{v_i^{(0)}}^{[T]}\right)_{i\in[n_0]},
\;\left(h_{v_i^{(L+1)}}^{[T]}\right)_{i\in[n_{L+1}]}
\right).
\]
Scale invariance of the readout is unaffected.
Notice that it is fine to either take the sum over layers $\ell$, or to instead concatenate the layer-wise mean readouts, as in \eqref{eq:gmn-readout}.
The latter is more expressive, but it restricts the model to inputs with a fixed number of layers $L$ (inconsequential to us, since we vary only the width).

\paragraph{Matrix-product ScaleGMN.}
The further modifications needed to make the matrix-product forward ScaleGMN are also almost the same as those for the simple GMN.
It is impossible to construct a bidirectional matrix-product ScaleGMN for positive-scaling symmetries, because the backward edge features would need to carry reciprocal weights, which cannot be expressed as a matrix product; for sign symmetries, this obstruction does not arise.

\emph{Node and edge features.}
We drop the edge features altogether, so that what is left is the raw scalar $\Wlbar_{ij}$, held fixed across rounds, exactly as in \Cref{app:gmn-lim}.
This is again harmless for scale equivariance.

\emph{Message-passing layer update.}
We restrict the forward message of \eqref{eq:scalegmn-impl-layer} by the specialization $A_e=A_v=I_{d_h}$ and $\Psi_m(x)=A^{\mathrm{in}}_{\ell}x$.
Since this is only a specialization of the parameters, the resulting models form a subclass of the duplication-compatible ScaleGMNs above, so scale equivariance is inherited and the only cost is expressivity.
Equivalently, this can be written in matrix form.
Writing $H^{(\ell),[t]}\in\RR^{n_\ell\times d_h}$ for the matrix whose $i$th row is $h^{[t]}_{v^{(\ell)}_i}$, the mean aggregate and the node update are
\begin{equation}
\label{eq:matprod-scale-impl}
Z^{(\ell),\mathrm{in},[t]}=\Wl H^{(\ell-1),[t]}A^{\mathrm{in}}_\ell,
\qquad
\widetilde H^{(\ell),[t+1]}=\Psi_h\left(H^{(\ell),[t]},Z^{(\ell),\mathrm{in},[t]}\right).
\end{equation}

For the bidirectional version under positive-scaling symmetry, this is impossible to construct:
the resulting aggregate would be a matrix product in the entrywise reciprocal of $(W^{(\ell+1)})^\top$ rather than in $(W^{(\ell+1)})^\top$ itself.

\section{Experiments}\label{appen:experiments}
Our experiments are built on the ScaleGMN codebase of \citet{kalogeropoulos2024scale}.
We experiment with four base GMN architectures, all sharing the same graph construction and node/edge preprocessing, obtained by crossing two settings of this codebase, \texttt{symmetry} (\texttt{permutation} or \texttt{scale}) and \texttt{direction} (\texttt{forward} or \texttt{bidirectional}).
\texttt{symmetry: scale} gives the ScaleGMN exactly as formulated in \Cref{app:gmn-scalegmn} above.
\texttt{symmetry: permutation} gives the GMN without the scaling symmetries.
On top of each of these four base models, we use our recipe to construct a duplication-compatible modification and a matrix-product modification, whenever the latter is well-defined.
\Cref{tab:model-variants} summarizes these model variants and where each is defined.

\begin{table}[h]
\caption{The four base GMN architectures compared in our experiments, and their duplication-compatible and matrix-product variants.
The forward and bidirectional ScaleGMN are given explicitly in \Cref{app:gmn-scalegmn}, with the modifications specified.
The forward and bidirectional GMN are defined analogously.}
\label{tab:model-variants}
\begin{center}
\setlength{\heavyrulewidth}{1.2pt}
\setlength{\lightrulewidth}{1pt}
\setlength{\cmidrulewidth}{1pt}
\setlength{\arrayrulewidth}{1pt}
\setlength{\aboverulesep}{0pt}
\setlength{\belowrulesep}{0pt}
\renewcommand{\arraystretch}{1.4}
\resizebox{\textwidth}{!}{%
\begin{tabular}{|l|p{5.5cm}|>{\centering\arraybackslash}p{4.6cm}|>{\centering\arraybackslash}p{3.7cm}|}
\toprule
\multicolumn{1}{|c|}{\textbf{\large Base GMN model}} & \multicolumn{1}{c|}{\textbf{\large Code setting}} & \textbf{\large Duplication-compatible variant} & \textbf{\large Matrix-product variant} \\
\midrule
GMN (forward) &
{\footnotesize\texttt{symmetry: permutation}\newline\texttt{direction: forward}} &
\multirow{4}{4.6cm}{\centering\begin{enumerate}[leftmargin=*,itemsep=0pt,topsep=0pt,parsep=0pt]
\item fan-in rescaled edges
\item mean aggregation
\item layer-wise mean readout
\end{enumerate}} &
\multirow{3}{3.7cm}{\centering message $\Wlbar_{ij}A^{\mathrm{in}}_\ell h_j$.} \\
\cline{1-2}
GMN (bidirectional) &
{\footnotesize\texttt{symmetry: permutation}\newline\texttt{direction: bidirectional}} &
&
\\
\cline{1-2}
ScaleGMN (forward) &
{\footnotesize\texttt{symmetry: scale}\newline\texttt{direction: forward}} &
&
\\
\cline{1-2}\cline{4-4}
ScaleGMN (bidirectional) &
{\footnotesize\texttt{symmetry: scale}\newline\texttt{direction: bidirectional}} &
&
\emph{not defined} \\
\bottomrule
\end{tabular}%
}
\end{center}
\end{table}

We evaluate all model variants of \Cref{tab:model-variants} whenever possible, except that the matrix-product bidirectional ScaleGMN is not defined, as explained in \Cref{app:gmn-scalegmn}.
Moreover, we note that the entire row 4 is not run on the CNN-accuracy prediction task.
This is because the bidirectional ScaleGMN has a known numerical instability problem as reported in Appendix~A.2 of \citet{kalogeropoulos2024scale}.
A bidirectional positive-scale-equivariant ScaleGMN requires the backward edge feature $1/W$.
In these CNN zoos, the weights have very small nonzero entries, and hence their reciprocal values are very large, causing numerical overflows during training.

\subsection{Verification of GMN compatibility and continuity}\label{app:verify-compatibility}

We numerically verify that our implementation of the GMN modifications satisfies the compatibility and continuity properties claimed in \Cref{sec:transferable_gmn}, on the two kinds of input networks used throughout the experiments: MLP weights (MNIST INRs of layout $[2,n,n,1]$) and CNN weights (the small CIFAR-10 zoo CNN of \citet{unterthiner2020predicting}, three stride-$2$ convolutions followed by global average pooling and a dense head).
All checks use an untrained GMN with $d_h=32$ and $T=2$ rounds, in float32.
Each table entry is the maximum absolute difference over the sampled input networks, width multipliers, and coordinates; we count a check as passing when this maximum is below $10^{-6}$.

\paragraph{Duplication compatibility.}
We widen each input network by the duplication widening \eqref{eq:dup} and compare the GMN's output on the widened network against its output on the base network.
For the MLP case, we use $5$ width-$32$ INRs, each widened by a pair of hidden-layer multipliers drawn independently and uniformly at random from $k\in\{2,3,4\}$.
For the CNN case, we use $3$ width-$16$ CNNs, each widened by both $k=2$ and $k=3$ (a single multiplier applied to all three convolutional layers), for $6$ widened networks in total.
\Cref{tab:verify-dup} reports the results, which align with the theoretical prediction of \Cref{prop:dup-invariance}.

\begin{table}[t]
\caption{With all three modifications in place, the modified GMN become duplication-compatible; removing any one of them breaks this property.
Entries are the maximum GMN output differences between an input network and its duplication widening \eqref{eq:dup}, the maximum taken over the sampled input networks and over the coordinates of the model's output.}
\label{tab:verify-dup}
\begin{center}
\small
\resizebox{\textwidth}{!}{%
\begin{tabular}{@{}llcccc@{}}
\toprule
Model & Input & all three modifications & without fan-in rescaling & with sum aggregation & with sum readout \eqref{eq:gmn-readout} \\
\midrule
GMN (forward) & MLP & $1.8\times10^{-7}\;\checkmark$ & $4.3\times10^{-3}\;\times$ & $1.3\times10^{-2}\;\times$ & $1.8\times10^{-1}\;\times$ \\
GMN (bidirectional) & MLP & $2.4\times10^{-7}\;\checkmark$ & $9.9\times10^{-3}\;\times$ & $3.6\times10^{-2}\;\times$ & $1.2\times10^{-1}\;\times$ \\
ScaleGMN (forward) & MLP & $1.9\times10^{-7}\;\checkmark$ & $1.2\times10^{-3}\;\times$ & $6.7\times10^{-1}\;\times$ & $1.5\times10^{-1}\;\times$ \\
ScaleGMN (bidirectional) & MLP & $3.0\times10^{-7}\;\checkmark$ & $3.2\times10^{-3}\;\times$ & $5.2\times10^{-1}\;\times$ & $3.1\times10^{-2}\;\times$ \\
\midrule
GMN (forward) & CNN & $1.6\times10^{-7}\;\checkmark$ & $7.4\times10^{-4}\;\times$ & $2.2\times10^{-2}\;\times$ & $1.4\times10^{-2}\;\times$ \\
GMN (bidirectional) & CNN & $1.6\times10^{-7}\;\checkmark$ & $9.6\times10^{-4}\;\times$ & $3.3\times10^{-2}\;\times$ & $2.6\times10^{-2}\;\times$ \\
ScaleGMN (forward) & CNN & $1.2\times10^{-7}\;\checkmark$ & $3.7\times10^{-4}\;\times$ & $3.4\times10^{-2}\;\times$ & $1.1\times10^{-2}\;\times$ \\
ScaleGMN (bidirectional) & CNN & $2.6\times10^{-7}\;\checkmark$ & $4.6\times10^{-3}\;\times$ & $1.8\times10^{-2}\;\times$ & $3.4\times10^{-2}\;\times$ \\
\bottomrule
\end{tabular}%
}
\end{center}
\end{table}

\paragraph{Compatibility with block-wise widenings.}
We next consider the more general, function-preserving widening operations of \Cref{sec:duplication-equivalence}, generating wider networks from the following five families:
\begin{enumerate}[leftmargin=*,topsep=0pt,itemsep=0pt]
    \item Duplication widening \eqref{eq:dup}.
    \item Kronecker widening in~\Cref{prop:kronecker} with doubly stochastic $P^{(\ell)}$.
    \item Kronecker widening in~\Cref{prop:kronecker} with row-stochastic $P^{(\ell)}$.
    \item Block-wise widening \eqref{eq:blockwise} with balanced blocks (\Cref{prop:matprod}(ii)).
    \item General block-wise widening \eqref{eq:blockwise}.
\end{enumerate}
Table~\ref{tab:verify-blockwise} reports the results for the GMN and ScaleGMN families (subtables~\ref{tab:verify-blockwise-gmn} and~\ref{tab:verify-blockwise-scalegmn}), which again align with the theoretical prediction of \Cref{prop:matprod}(ii).

\begin{table}[t]
\caption{The forward matrix-product model is compatible with the general block-wise widening; the other four families are subclasses of it and are hence also compatible.
The bidirectional matrix-product model is compatible only with block-wise widenings that satisfy the column-balance condition of \Cref{prop:matprod}(ii); the duplication widening and the doubly stochastic Kronecker widening are subclasses of this condition and are hence also compatible.
The duplication-compatible model (forward or bidirectional) is compatible only with the duplication widening.
The same pattern holds for both the GMN (a) and ScaleGMN families (b).
Entries are the output difference between an input network and its widening, maximized over the sampled input networks and over the coordinates of the model's output.}
\label{tab:verify-blockwise}
\begin{center}

\begin{subtable}[t]{\textwidth}
\centering
\caption{GMN family.}
\label{tab:verify-blockwise-gmn}
\small
\resizebox{\textwidth}{!}{%
\begin{tabular}{@{}lcccc@{}}
\toprule
Widenings & \shortstack{Matrix-product\\(forward)} & \shortstack{Matrix-product\\(bidirectional)} & \shortstack{Duplication-compatible\\(forward)} & \shortstack{Duplication-compatible\\(bidirectional)} \\
\midrule
\multicolumn{5}{@{}l}{\emph{MLP weights}} \\
Duplication \eqref{eq:dup} & $2.4\times10^{-7}\;\checkmark$ & $3.3\times10^{-7}\;\checkmark$ & $1.8\times10^{-7}\;\checkmark$ & $2.7\times10^{-7}\;\checkmark$ \\
Kronecker, doubly-stoch. & $3.0\times10^{-7}\;\checkmark$ & $2.4\times10^{-7}\;\checkmark$ & $4.4\times10^{-3}\;\times$ & $9.8\times10^{-3}\;\times$ \\
Kronecker, row-stoch. & $2.4\times10^{-7}\;\checkmark$ & $1.6\times10^{-1}\;\times$ & $2.9\times10^{-2}\;\times$ & $2.5\times10^{-2}\;\times$ \\
Block-wise, balanced & $2.4\times10^{-7}\;\checkmark$ & $2.7\times10^{-7}\;\checkmark$ & $8.9\times10^{-3}\;\times$ & $2.0\times10^{-2}\;\times$ \\
Block-wise \eqref{eq:blockwise} & $2.4\times10^{-7}\;\checkmark$ & $1.4\times10^{-1}\;\times$ & $3.1\times10^{-2}\;\times$ & $3.0\times10^{-2}\;\times$ \\
\midrule
\multicolumn{5}{@{}l}{\emph{CNN weights}} \\
Duplication \eqref{eq:dup} & $1.3\times10^{-7}\;\checkmark$ & $1.9\times10^{-7}\;\checkmark$ & $1.5\times10^{-7}\;\checkmark$ & $4.5\times10^{-7}\;\checkmark$ \\
Kronecker, doubly-stoch.  & $1.2\times10^{-7}\;\checkmark$ & $2.4\times10^{-7}\;\checkmark$ & $3.1\times10^{-3}\;\times$ & $4.6\times10^{-2}\;\times$ \\
Kronecker, row-stoch. & $1.9\times10^{-7}\;\checkmark$ & $1.5\times10^{-2}\;\times$ & $4.6\times10^{-3}\;\times$ & $3.5\times10^{-2}\;\times$ \\
Block-wise, balanced & $6.7\times10^{-8}\;\checkmark$ & $2.4\times10^{-7}\;\checkmark$ & $3.4\times10^{-2}\;\times$ & $1.6\times10^{-1}\;\times$ \\
Block-wise \eqref{eq:blockwise} & $2.7\times10^{-7}\;\checkmark$ & $1.4\times10^{-1}\;\times$ & $4.8\times10^{-2}\;\times$ & $2.4\times10^{-1}\;\times$ \\
\bottomrule
\end{tabular}%
}
\end{subtable}

\vspace{1.5em}

\begin{subtable}[t]{\textwidth}
\centering
\caption{ScaleGMN family.}
\label{tab:verify-blockwise-scalegmn}
\small
\resizebox{\textwidth}{!}{%
\begin{tabular}{@{}lcccc@{}}
\toprule
Widenings & \shortstack{Matrix-product\\(forward)} & \shortstack{Matrix-product\\(bidirectional)} & \shortstack{Duplication-compatible\\(forward)} & \shortstack{Duplication-compatible\\(bidirectional)} \\
\midrule
\multicolumn{5}{@{}l}{\emph{MLP weights}} \\
Duplication \eqref{eq:dup} & $6.6\times10^{-7}\;\checkmark$ & $7.2\times10^{-7}\;\checkmark$ & $2.1\times10^{-7}\;\checkmark$ & $1.8\times10^{-7}\;\checkmark$ \\
Kronecker, doubly-stoch. & $4.7\times10^{-7}\;\checkmark$ & $6.6\times10^{-7}\;\checkmark$ & $7.6\times10^{-2}\;\times$ & $6.9\times10^{-2}\;\times$ \\
Kronecker, row-stoch. & $8.6\times10^{-7}\;\checkmark$ & $6.0\times10^{-1}\;\times$ & $8.8\times10^{-2}\;\times$ & $1.0\times10^{-1}\;\times$ \\
Block-wise, balanced & $4.8\times10^{-7}\;\checkmark$ & $7.5\times10^{-7}\;\checkmark$ & $1.0\times10^{-1}\;\times$ & $7.2\times10^{-2}\;\times$ \\
Block-wise \eqref{eq:blockwise} & $5.0\times10^{-7}\;\checkmark$ & $4.6\times10^{-1}\;\times$ & $1.0\times10^{-1}\;\times$ & $8.0\times10^{-2}\;\times$ \\
\midrule
\multicolumn{5}{@{}l}{\emph{CNN weights}} \\
Duplication \eqref{eq:dup} & $2.4\times10^{-7}\;\checkmark$ & $2.2\times10^{-7}\;\checkmark$ & $1.8\times10^{-7}\;\checkmark$ & $3.0\times10^{-7}\;\checkmark$ \\
Kronecker, doubly-stoch.  & $2.1\times10^{-7}\;\checkmark$ & $1.2\times10^{-7}\;\checkmark$ & $6.9\times10^{-4}\;\times$ & $1.6\times10^{-3}\;\times$ \\
Kronecker, row-stoch. & $1.2\times10^{-7}\;\checkmark$ & $1.7\times10^{-2}\;\times$ & $5.1\times10^{-3}\;\times$ & $8.7\times10^{-3}\;\times$ \\
Block-wise, balanced & $2.4\times10^{-7}\;\checkmark$ & $2.5\times10^{-7}\;\checkmark$ & $2.1\times10^{-2}\;\times$ & $1.8\times10^{-2}\;\times$ \\
Block-wise \eqref{eq:blockwise} & $4.8\times10^{-7}\;\checkmark$ & $7.6\times10^{-2}\;\times$ & $1.3\times10^{-2}\;\times$ & $1.6\times10^{-2}\;\times$ \\
\bottomrule
\end{tabular}%
}
\end{subtable}

\end{center}
\end{table}

\paragraph{Spectral continuity.}
We evaluate both models on the family used in the proof of \Cref{prop:misalignment}(ii).
For MLP weights, we set $\overline W^{(2)}=H_n$, the Sylvester Hadamard matrix, and every other weight and bias to zero; since $\norm[V_n,\op]{\theta_n}=1/\sqrt n\to0$, a model Lipschitz in the normalized spectral norm must satisfy $f(\theta_n)\to f(0)$.
For CNN weights, the same family is placed at the center of the second layer's $3\times3$ kernel, i.e.\ $\overline W^{(2)}[:,:,\mathrm{centre}]=H_n$ with all other weights zero, so that layer acts as a $1\times1$ convolution with matrix $H_n/n$.
\Cref{tab:verify-continuity} reports the results, which align with the theoretical prediction of \Cref{prop:matprod}(i) and \Cref{prop:misalignment}(ii).

\begin{table}[h]
\caption{The matrix-product GMN is continuous in the normalized spectral norm, whereas the duplication-compatible GMN is not.
Entries are $\abs{f(\theta_n)-f(0)}$ on the Hadamard family, which should converge to $0$ as $n\to\infty$ for continuous models.}
\label{tab:verify-continuity}
\begin{center}
\resizebox{\textwidth}{!}{%
\small
\begin{tabular}{@{}ccccccc@{}}
\toprule
& & \multicolumn{2}{c}{MLP weights} & \multicolumn{2}{c}{CNN weights} \\
\cmidrule(lr){3-4}\cmidrule(lr){5-6}
$n$ & $\norm[V_n,\op]{\theta_n}$ & Matrix-product GMN & Duplication-compatible GMN & Matrix-product GMN & Duplication-compatible GMN \\
\midrule
$16$ & $0.2500$ & $2.08\times10^{-3}$ & $4.99\times10^{-3}$ & $8.76\times10^{-4}$ & $6.68\times10^{-4}$ \\
$32$ & $0.1768$ & $1.04\times10^{-3}$ & $5.05\times10^{-3}$ & $4.38\times10^{-4}$ & $6.42\times10^{-4}$ \\
$64$ & $0.1250$ & $5.21\times10^{-4}$ & $5.08\times10^{-3}$ & $2.19\times10^{-4}$ & $6.30\times10^{-4}$ \\
$128$ & $0.0884$ & $2.60\times10^{-4}$ & $5.09\times10^{-3}$ & $1.10\times10^{-4}$ & $6.23\times10^{-4}$ \\
$256$ & $0.0625$ & $1.30\times10^{-4}$ & $5.10\times10^{-3}$ & --- & --- \\
\bottomrule
\end{tabular}%
}
\end{center}
\end{table}

\subsection{Size generalization on hand-crafted widenings}
\label{app:hand-crafted-widenings}

As a sanity check, we first consider size generalization on hand-crafted widenings of the input networks, whose outcome is predicted exactly by the design principles behind the GMN modifications.
We train the GMNs at the smallest width only, then evaluate the trained checkpoints on wider input networks obtained by two widening operations, duplication widening~\eqref{eq:dup} and block-wise widening~\eqref{eq:blockwise}, both of which by construction represent exactly the same function as the training-width network.

\paragraph{Dataset and training procedure.}
Every GMN is trained only on the smallest input width: $51\,000$ MNIST INRs at width $24$, $56\,000$ FMNIST INRs at width $32$, or $8000$ CNNs at width $c=16$ for each of the two accuracy-prediction zoos.
As test data, we use $1000$ held-out networks at the training width.
Test networks at larger widths are then obtained by applying the duplication widening~\eqref{eq:dup} or the block-wise widening~\eqref{eq:blockwise} independently at each target width.
Since both widenings preserve the represented function by construction, every widened copy retains its source network's original label.

Training of the GMNs uses AdamW, a batch size of $64$, $1000$ warmup steps, at most $200$ epochs, and early stopping with a patience of $50$ epochs.
On the INR tasks, the learning rate is chosen by a sweep over the seven values $6.25\times10^{-5},1.25\times10^{-4},\dots,4\times10^{-3}$, each candidate trained for $8600$ steps ($\approx10$ epochs).
On the accuracy-prediction tasks, no sweep is run and the configured learning rate of $10^{-3}$ is used throughout.
Both the learning rate and the evaluated checkpoint are selected on validation data at the training width alone.

\begin{figure}[t]
\begin{subfigure}{\textwidth}
\centering
\includegraphics[width=\textwidth]{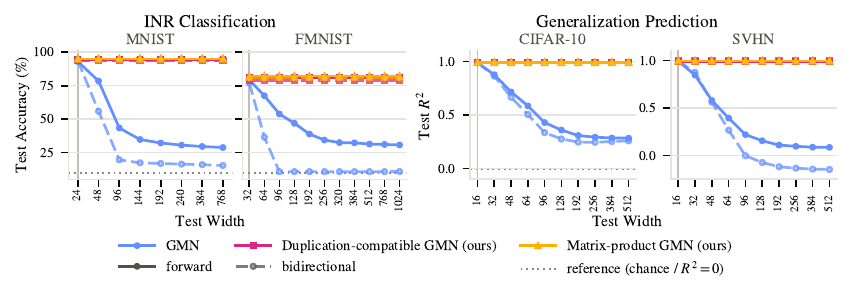}
\caption{GMN families.}
\label{fig:widen-dup-gmn}
\end{subfigure}
\\[1em]
\begin{subfigure}{\textwidth}
\centering
\includegraphics[width=\textwidth]{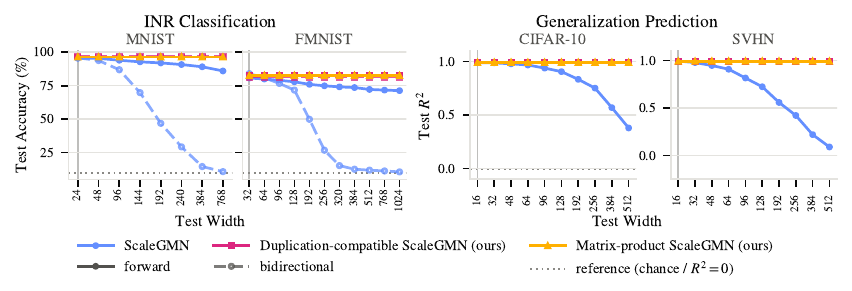}
\caption{ScaleGMN families.
(Bidirectional ScaleGMN is not evaluated on the CNN tasks because of training instability.)}
\label{fig:widen-dup-scalegmn}
\end{subfigure}
\caption{Size generalization on wider input networks generated with duplication widening~\eqref{eq:dup}.
Both duplication-compatible and matrix-product GMNs generalize in size perfectly, with flat error curves across all test widths, while the baseline GMNs, which are not duplication-compatible, degrade sharply on wider input networks.}
\label{fig:widen-dup}
\end{figure}

\paragraph{Result 1: Duplication widening.}
\Cref{fig:widen-dup-gmn,fig:widen-dup-scalegmn} show the results under the duplication widening~\eqref{eq:dup}.
As predicted by \Cref{prop:dup-invariance}, the duplication-compatible and matrix-product GMNs reproduce their in-distribution performance exactly at every test width, confirming that our modifications let the GMNs generalize to wider input networks formed by duplication widening.

\begin{figure}[t]
\begin{subfigure}{\textwidth}
\centering
\includegraphics[width=\textwidth]{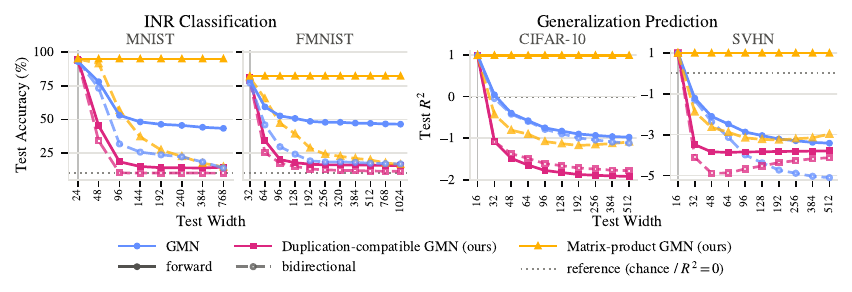}
\caption{GMN families.}
\label{fig:widen-block-gmn}
\end{subfigure}
\\[1em]
\begin{subfigure}{\textwidth}
\centering
\includegraphics[width=\textwidth]{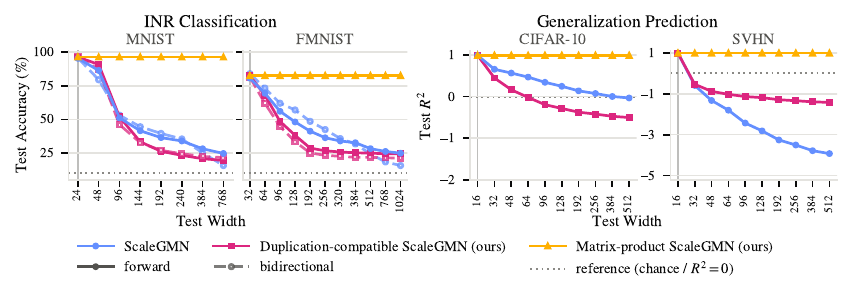}
\caption{ScaleGMN families.
(Bidirectional ScaleGMN is not evaluated on the CNN tasks because of training instability.)}
\label{fig:widen-block-scalegmn}
\end{subfigure}
\caption{Size generalization on wider input networks generated with block-wise widening~\eqref{eq:blockwise}.
Only the forward matrix-product GMNs/ScaleGMNs generalize in size perfectly, with flat error curves across all test widths, while all other models, which are not compatible with block-wise widenings, degrade sharply on wider input networks.}
\label{fig:widen-block}
\end{figure}

\paragraph{Result 2: Block-wise widening.}
\Cref{fig:widen-block-gmn,fig:widen-block-scalegmn} show the results under the block-wise widening~\eqref{eq:blockwise}.
As predicted by \Cref{prop:matprod}(ii), only the forward matrix-product GMNs are compatible with the block-wise widening, and accordingly only they reproduce their in-distribution performance exactly at every test width, confirming that our modifications let them generalize to wider input networks formed by block-wise widening.

\subsection{Size generalization on independently trained weights}
\label{app:sizegen-random-init}
Next, in the more realistic scenario, we consider size generalization on input networks of different widths that are independently trained from random initialization.

\subsubsection{Dataset generation}

\paragraph{INR datasets.}
For each of MNIST and FMNIST, we fit one two-hidden-layer ReLU INR $\RR^2\to\RR$ per image, independently at every width, and the metanetwork predicts the image's class from the INR weights alone.
The image identities are partitioned across widths before any INR is fitted, so no image contributes weights to more than one width.
MNIST uses $51\,000$ training and $1000$ validation INRs at the training width $24$, plus $1000$ test INRs per width at $13$ widths up to $1024$ ($42.7\times$ the training width).
FMNIST uses $56\,000$ training and $1000$ validation INRs at the training width $32$, plus $1000$ test INRs per width at $12$ widths up to $1024$ ($32\times$).

The training width is chosen as the smallest width at which the INR fits are of acceptable quality.
The reason is that a training width whose fits are markedly worse than those at the test widths would confound the results: a rise in test accuracy with width could then reflect improving input quality rather than genuine size generalization by the metanetwork.

Every width is fitted twice, once under SP and once under $\mu$P.
$\mu$P is set to coincide with SP at the training width, so the two parameterizations agree exactly in distribution and differ only out of distribution.
Under SP, the fitting learning rate must be retuned at every width: a learning rate that fits well at the training width tends to perform increasingly poorly at larger widths, yielding lower-quality inputs there.
The selected values decay from $10^{-2}$ at the training width to $5\times10^{-4}$ (MNIST) and $10^{-3}$ (FMNIST) at width $1024$.
Under $\mu$P, by contrast, the learning-rate sweep is run only once, at the base width, and the resulting value is held fixed at every other width: \citet{yang2021tuning} observed that the optimal learning rate is width-independent under $\mu$P, which is precisely what allows this hyperparameter transfer across widths.

To assess INR quality, \Cref{tab:inr-quality} reports the peak signal-to-noise ratio (PSNR, measured in decibels, dB; higher is better) and the failure rate ($\mathrm{MSE}>0.01$) at every fitted width.
Under SP, quality peaks a few widths past the training width and then degrades, most severely on FMNIST.
Under $\mu$P, quality stays essentially flat, with a low failure rate throughout.
This is consistent with the finding of~\citet{yang2021tuning} that under $\mu$P, wider is simply better, whereas under SP this is usually not the case.
This drift in input quality under SP should be taken into account when interpreting the size-generalization results: part of the accuracy decrease at larger widths may be attributable to the decline in input quality rather than to the metanetwork itself.
However, the considerable decrease in accuracy under SP at larger widths cannot be explained by input quality alone, so the metanetwork's size generalization can be concluded to be genuinely poor in that regime.

\begin{table}[h]
\caption{Reconstruction quality of the fitted INRs at every width: peak signal-to-noise ratio (PSNR, in dB; higher is better) and failure rate ($\mathrm{MSE}>0.01$, in \%).
Under $\mu$P, quality generally stays flat across all widths; under SP it degrades past its peak, most severely on FMNIST.}
\label{tab:inr-quality}
\begin{center}
\small
\resizebox{\textwidth}{!}{%
\begin{tabular}{@{}llccccccccccccc@{}}
\toprule
Dataset & Param. & $24$ & $32$ & $48$ & $64$ & $80$ & $96$ & $128$ & $192$ & $256$ & $384$ & $512$ & $768$ & $1024$ \\
\midrule
MNIST & SP & $30.2/0.7$ & $31.9/0.3$ & $34.5/0.2$ & $34.8/0.2$ & $35.4/0.4$ & $35.9/0.5$ & $35.0/0.4$ & $32.6/0.7$ & $33.4/1.1$ & $31.1/1.1$ & $31.7/1.0$ & $29.3/1.0$ & $29.5/3.2$ \\
MNIST & $\mu$P & $30.2/0.3$ & $31.3/0.1$ & $32.2/0.3$ & $32.8/0.0$ & $33.0/0.0$ & $33.0/0.3$ & $33.2/0.0$ & $33.1/0.1$ & $33.0/0.2$ & $32.8/0.0$ & $32.5/0.0$ & $32.8/0.0$ & $32.5/0.0$ \\
FMNIST & SP & --- & $25.5/5.9$ & $27.3/2.7$ & $27.7/2.8$ & $27.6/2.3$ & $27.9/2.2$ & $28.0/3.9$ & $27.1/4.3$ & $27.0/5.4$ & $25.6/7.5$ & $25.1/10.9$ & $24.1/13.6$ & $23.3/19.5$ \\
FMNIST & $\mu$P & --- & $25.5/4.8$ & $26.1/4.4$ & $26.6/3.4$ & $26.9/3.0$ & $27.2/2.3$ & $27.3/1.8$ & $27.3/2.4$ & $27.4/1.9$ & $27.5/1.5$ & $27.4/1.7$ & $27.4/2.2$ & $27.3/2.7$ \\
\bottomrule
\end{tabular}%
}
\end{center}
\end{table}

\paragraph{CNN datasets.}
The two accuracy prediction datasets are zoos of CNNs --- three $3\times3$ convolutions of $c$ channels each, global average pooling, and a linear classifier, $4970$ parameters at $c=16$ --- trained on grayscale CIFAR-10 and on grayscale SVHN, and the GMN's task is to predict a member's test accuracy from its weights.
Each member draws its hyperparameters log-uniformly (learning rate in $[5\times10^{-5},5\times10^{-2}]$, initialization multiplier in $[0.25,4]$, weight decay in $[10^{-6},10^{-1}]$) together with a dropout rate from $\{0,0.2,0.45\}$, a fraction of the training set from $\{0.1,0.25,0.5,1\}$, and an epoch budget from $\{5,10,20\}$.
Particularly, the hyperparameter distribution is identical at every width and the widths differ only in their number of channels.

We use $8000$ training and $1000$ validation networks at $c=16$, and $1000$ test networks per width at ten widths from $c=16$ to $c=512$ ($32\times$), with a separate seed per role and per width.
As on the INR tasks, every width is instantiated twice, under SP and under $\mu$P with base width $16$.
The $\mu$P networks are set to coincide with the SP networks at the training width, so the two parameterizations agree exactly in distribution and differ only out of distribution.

We further examine a property of the generated data that matters for the size-generalization evaluation below: how the prediction target itself drifts with width.
Wider networks have more capacity and should fit the training data better, so we expect test accuracy to generally increase with width.
A GMN trained on the smallest width alone never observes this trend, so extrapolating it is a genuinely out-of-distribution problem.
One route by which it could still succeed is if the function computed by a higher-accuracy wide network resembles the function computed by some lower-accuracy narrow network with different hyperparameters that the GMN did see during training, allowing the GMN to generalize by detecting the similarity in the represented function.
\Cref{tab:cnn-label-shift} reports, at every width, the fraction of that width's test accuracies lying inside the central $98\%$ range of the training width's accuracies, together with the shift of the median accuracy.
It is observed that under SP, there is a clear drift of the prediction target with width, while under $\mu$P the target is much more stable.
This motivates us to explore various evaluation criteria that are robust to the drift, as described next.

\begin{table}[h]
\caption{Drift of the prediction target with width on the two CNN zoos, relative to the training width $c=16$.
Each cell gives the fraction of that width's test accuracies falling inside the central $98\%$ range of the $c=16$ accuracies, and the shift of the median accuracy (both are $1.000$ and $+0.000$ at $c=16$ by definition).}
\label{tab:cnn-label-shift}
\begin{center}
\small
\resizebox{\textwidth}{!}{%
\begin{tabular}{@{}llccccccccc@{}}
\toprule
Dataset & Param. & $32$ & $48$ & $64$ & $96$ & $128$ & $192$ & $256$ & $384$ & $512$ \\
\midrule
CIFAR-10 & SP & $0.942/{+}0.031$ & $0.889/{+}0.063$ & $0.859/{+}0.079$ & $0.818/{+}0.106$ & $0.792/{+}0.112$ & $0.728/{+}0.128$ & $0.726/{+}0.136$ & $0.654/{+}0.161$ & $0.678/{+}0.154$ \\
CIFAR-10 & $\mu$P & $0.958/{+}0.022$ & $0.921/{+}0.033$ & $0.884/{+}0.042$ & $0.892/{+}0.054$ & $0.880/{+}0.053$ & $0.863/{+}0.062$ & $0.863/{+}0.060$ & $0.869/{+}0.063$ & $0.864/{+}0.062$ \\
SVHN & SP & $0.928/{+}0.000$ & $0.876/{+}0.048$ & $0.846/{+}0.079$ & $0.799/{+}0.113$ & $0.761/{+}0.130$ & $0.715/{+}0.159$ & $0.704/{+}0.174$ & $0.652/{+}0.224$ & $0.700/{+}0.197$ \\
SVHN & $\mu$P & $0.948/{+}0.000$ & $0.911/{+}0.000$ & $0.876/{+}0.000$ & $0.880/{+}0.009$ & $0.864/{+}0.000$ & $0.850/{+}0.013$ & $0.852/{+}0.011$ & $0.855/{+}0.024$ & $0.847/{+}0.002$ \\
\bottomrule
\end{tabular}%
}
\end{center}
\end{table}

\subsubsection{Evaluation and results}

\paragraph{Training procedure.}
The GMNs are trained exactly as in \Cref{app:hand-crafted-widenings}, except that the learning-rate sweep is always run before selecting the best GMN based on test performance at the training width.

\begin{figure}[p]
\begin{subfigure}{\textwidth}
\centering
\includegraphics[width=\textwidth]{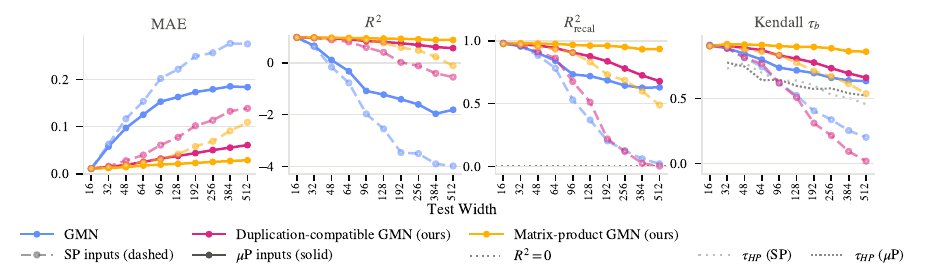}
\caption{GMN (forward).}
\label{fig:criteria-cnn-gmn-fw}
\end{subfigure}
\\[1em]
\begin{subfigure}{\textwidth}
\centering
\includegraphics[width=\textwidth]{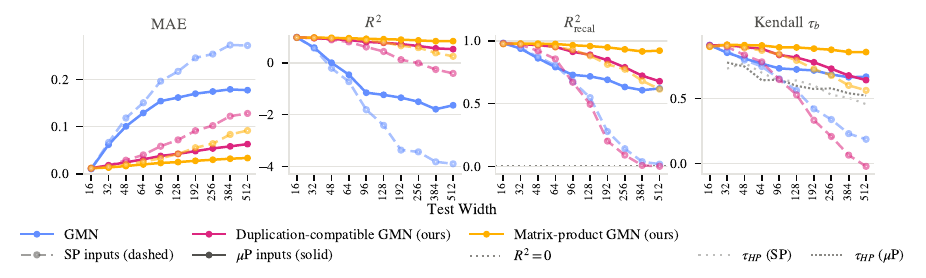}
\caption{GMN (bidirectional).}
\end{subfigure}
\\[1em]
\begin{subfigure}{\textwidth}
\centering
\includegraphics[width=\textwidth]{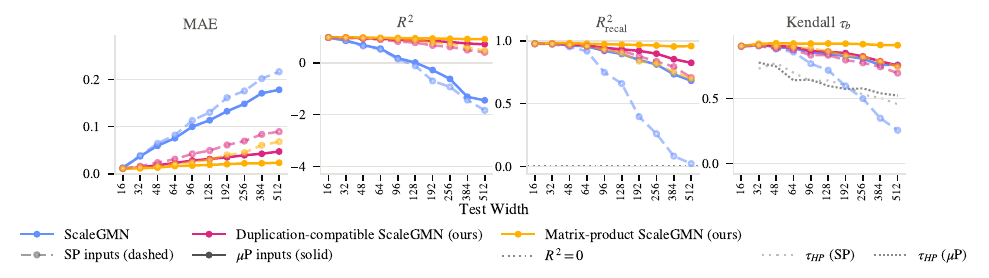}
\caption{ScaleGMN (forward).}
\label{fig:criteria-cnn-scalegmn-fw}
\end{subfigure}
\caption{The four evaluation criteria for accuracy prediction on the CIFAR-10 CNN zoo, one row of panels per setting.}
\label{fig:criteria-cifar10}
\end{figure}

\begin{figure}[p]
\begin{subfigure}{\textwidth}
\centering
\includegraphics[width=\textwidth]{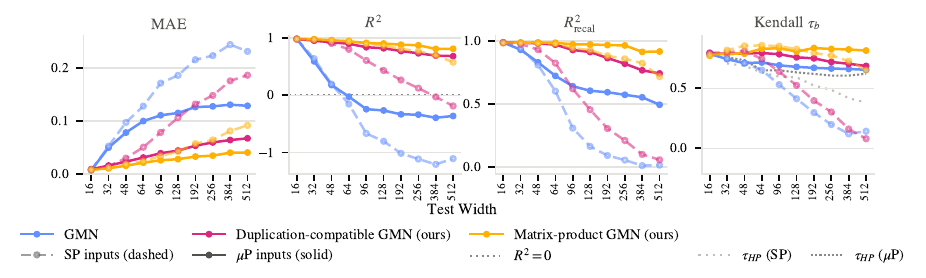}
\caption{GMN (forward).}
\label{fig:criteria-svhn-gmn-fw}
\end{subfigure}
\\[1em]
\begin{subfigure}{\textwidth}
\centering
\includegraphics[width=\textwidth]{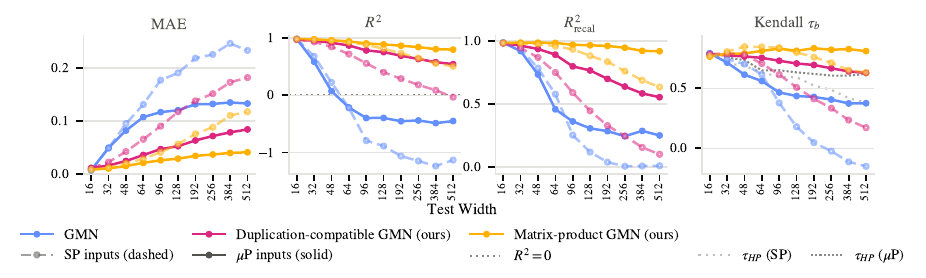}
\caption{GMN (bidirectional).}
\end{subfigure}
\\[1em]
\begin{subfigure}{\textwidth}
\centering
\includegraphics[width=\textwidth]{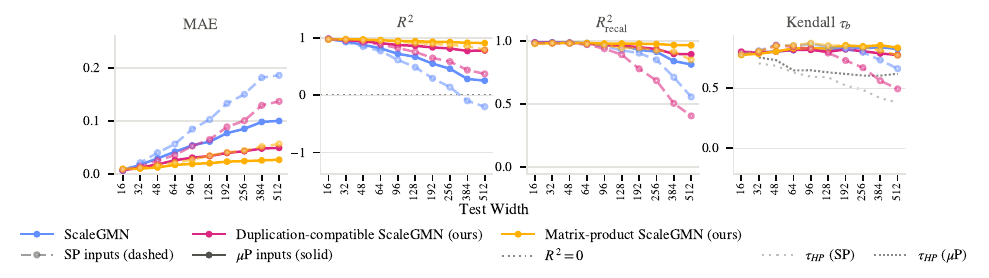}
\caption{ScaleGMN (forward).}
\label{fig:criteria-svhn-scalegmn-fw}
\end{subfigure}
\caption{The four evaluation criteria for accuracy prediction on the SVHN CNN zoo, one row of panels per setting.}
\label{fig:criteria-svhn}
\end{figure}

\paragraph{Evaluation criteria for accuracy prediction.}
Because the prediction target drifts with width, one criterion alone cannot tell a metanetwork that has learned nothing about a new width from one that ranks the networks well but predicts values on the wrong scale.
We therefore report four criteria for the CNN accuracy prediction task.
\begin{itemize}[leftmargin=*,topsep=0pt,itemsep=0pt]
\item The mean absolute error $\frac{1}{n_{\mathrm{test}}}\sum_i |y_i - \hat y_i|$, for labels $y_i$ and predictions $\hat y_i$ at a given width, is measured in units of accuracy and is not normalized.
\item $R^2 = 1 - \frac{\sum_i (y_i - \hat y_i)^2}{\sum_i (y_i - \bar y)^2}$, where $\bar y$ is the mean label at that width, compares the squared errors with the spread of the labels at that width: $R^2 = 1$ means the predictions match the labels exactly, $R^2 = 0$ means the predictor does no better than always guessing the mean label, and $R^2 < 0$ means it does worse than that constant baseline.
\item $R^2$ after an optimal affine recalibration, $\Rrecal = \max_{a,b} R^2(y, a\hat y + b)$, allows the predictions to be shifted and rescaled first, so a drift that is affine costs nothing.
Comparing it with $R^2$ tells apart the two ways $R^2$ can fail: low $R^2$ with high $\Rrecal$ means the predictor gets the networks in the right relative positions but on the wrong scale, while both low means it fails outright.
\item Kendall's $\tau_b$ \citep{kendall1938new} is a rank correlation: it asks only whether the predicted ordering of the networks matches the true ordering of their accuracies, ignoring the values on either side, so any monotone drift leaves it unchanged.
It is also the criterion best suited to model selection, where one only wants to pick the best network out of a pool and the predicted values themselves do not matter.
\end{itemize}
For the Kendall's $\tau_b$ criterion, we also report a baseline $\tau_{HP}$ that quantifies how much of the ranking at a test width is available without reading the wide weights at all: one may assume the best hyperparameters at the training width will also be the best hyperparameters at a test width, so one could simply rank the wide networks based on the hyperparameters that generated them.
We quantify this by generating $200$ networks per width that realizes the same hyperparameter draws at every width.
Then, we compute the Kendall's $\tau_b$ between the accuracies of the same draw at the training width $c_0 = 16$ and at width $c$.
The resulting $\tau_{HP}$ is included as a baseline in the $\tau_b$ panels of \Cref{fig:criteria-cifar10,fig:criteria-svhn} to indicate how much of the ranking is available without reading the wide weights at all.

\begin{figure}[p]
\begin{subfigure}{\textwidth}
\centering
\includegraphics[width=\textwidth]{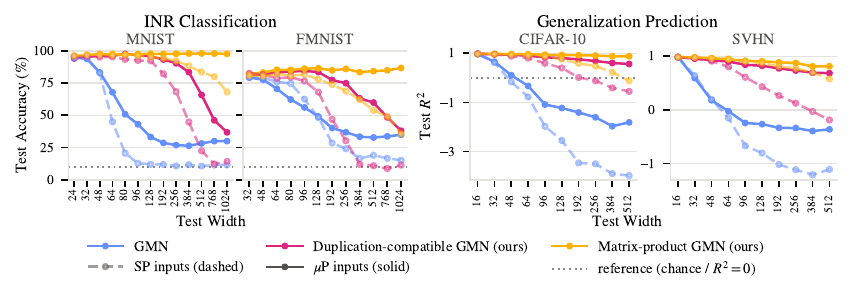}
\caption{GMN (forward).}
\label{fig:sizegen-gmn-fw}
\end{subfigure}
\\[0pt]
\ifdefined\arxivversion\else
\begin{subfigure}{\textwidth}
\centering
\includegraphics[width=\textwidth]{figures/fig_sizegen_gmn_bd.pdf}
\caption{GMN (bidirectional).}
\label{fig:sizegen-gmn-bd}
\end{subfigure}
\\[0pt]
\begin{subfigure}{\textwidth}
\centering
\includegraphics[width=\textwidth]{figures/fig_sizegen_scalegmn_fw.pdf}
\caption{ScaleGMN (forward).}
\label{fig:sizegen-scalegmn-fw}
\end{subfigure}
\\[0pt]
\fi
\begin{subfigure}{\textwidth}
\centering
\includegraphics[width=0.72\textwidth]{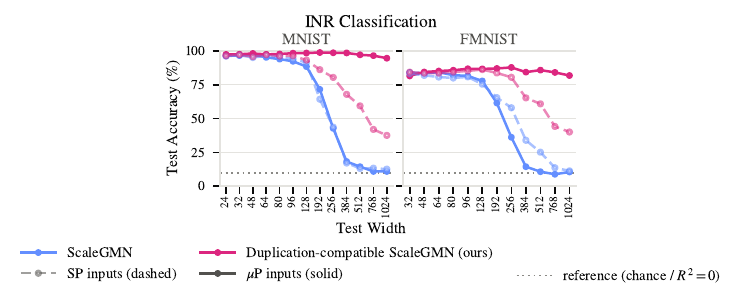}
\caption{ScaleGMN (bidirectional).
(Bidirectional ScaleGMN does not have a valid matrix-product variant; Also, bidirectional ScaleGMN is not evaluated on the CNN tasks because of training instability.)}
\label{fig:sizegen-scalegmn-bd}
\end{subfigure}
\ifdefined\arxivversion
\caption{Size generalization on wider input networks trained independently from random initialization, under either $\mu$P or SP, for the two architectures not shown in the main text (\Cref{fig:sizegen}).}
\else
\caption{Size generalization on wider input networks trained independently from random initialization, under either $\mu$P or SP.
\Cref{fig:sizegen-gmn-bd} repeats the main-text figure (\Cref{fig:sizegen}) for completeness.}
\fi
\label{fig:sizegen-appendix}
\end{figure}

\paragraph{Results.}
\ifdefined\arxivversion
The main-text figure (\Cref{fig:sizegen}) shows the bidirectional GMN and the forward ScaleGMN.
\Cref{fig:sizegen-appendix} gives the two remaining GMN variants (\Cref{fig:sizegen-gmn-fw,fig:sizegen-scalegmn-bd}) for completeness.
All four variants use the same four datasets, the same training widths, and the same test-width ranges.
\else
The main-text figure (\Cref{fig:sizegen}) shows the plain GMN, bidirectional.
\Cref{fig:sizegen-appendix} repeats it as \Cref{fig:sizegen-gmn-bd} alongside the other GMN variants (\Cref{fig:sizegen-gmn-fw,fig:sizegen-scalegmn-fw,fig:sizegen-scalegmn-bd}) for completeness.
All four panels use the same four datasets, the same training widths, and the same test-width ranges.
\fi

For the CNN accuracy-prediction task, we further report the results forall four evaluation criteria side-by-side in \Cref{fig:criteria-cifar10,fig:criteria-svhn} for the CIFAR-10 and SVHN datasets respectively.
We use the CIFAR-10 results as an example to show how these plots give a fine-grained understanding of the GMN's size generalization performance; similar patterns arise in the SVHN results, which we omit for brevity.

For example, compare the baseline ScaleGMN (forward) with its duplication-compatible counterpart under $\mu$P at width $c=512$ (\Cref{fig:criteria-cnn-scalegmn-fw}).
The baseline ScaleGMN still achieves $\tau_b=0.761$ but has $R^2=-1.450$ with $\Rrecal=+0.682$: it still orders the wide networks well, and its predictions are merely on the wrong scale.
Its duplication-compatible counterpart has an indistinguishable $\tau_b=0.756$, so the modification does not improve the ranking, but its $R^2=+0.714$ shows that it predicts the actual accuracy values much better.
Because $\tau_b$ and $\Rrecal$ barely change between the two variants while $R^2$ swings by nearly $2$, the gain is specifically in tracking the accuracy drift with width, not in ranking or in affine-corrected fit.

Under SP, both the baseline GMN (forward) and the duplication-compatible GMN (forward) perform badly on all four criteria (\Cref{fig:criteria-cnn-gmn-fw}), indicating a genuine failure: neither the actual accuracy values nor the ranking of the networks is recovered.

By contrast, the matrix-product GMN and the matrix-product ScaleGMN under $\mu$P (\Cref{fig:criteria-cnn-gmn-fw,fig:criteria-cnn-scalegmn-fw}) are where all four criteria agree that size generalization is reasonably good.

\section{Theory of size generalization on trained weights}
\label{appen:size-generalization-theory}
Throughout this section, we use the following data model to describe the distribution of trained neural networks at each width.
We first fix a training rule.
By this we mean that the following are shared across all widths: a dataset, a loss, an architecture specified up to its width, an optimization algorithm, and a parameterization (i.e., the scaling of initialization variances and learning rates with the model's width as in \Cref{tab:width-scalings}).
For each $n$, these determine a deterministic map $\mathsf{Train}_n$, which sends the hyperparameters, the auxiliary randomness, and the initialization to the weights at the end of training.

The training procedure for the width-$n$ network is driven by three sources of randomness, which are mutually independent.
\begin{itemize}[leftmargin=*,topsep=0pt,itemsep=0pt]
\item $\xi_n$: the random initialization of the width-$n$ network. We assume the $\xi_n$ are independent across $n$.
\item $\Gamma$: the hyperparameters (learning rate, weight decay, batch size, training horizon, etc.).
We assume $\operatorname{Law}(\Gamma)$ does not depend on $n$:
every width draws its hyperparameters from the same distribution.
\item $S$: the remaining random choices, such as the minibatch sequence, data augmentations or dropout masks, and optimizer noise.
We assume $\operatorname{Law}(S)$ does not depend on $n$.
\end{itemize}

For width $n$, define
\[
\theta_n=\mathsf{Train}_n(\Gamma,S,\xi_n),
\qquad
\mu_n\colonequals \operatorname{Law}(\theta_n).
\]

\paragraph{Scope of the data model.}
The data model above is fairly flexible, but it does exclude some practically relevant cases.

The assumption of a fixed training rule across widths, and the requirement that $\operatorname{Law}(S)$ be width-independent, are both mild.
Dropping either would mean asking a GMN trained on narrow weights to generalize to wide weights produced by a different optimizer or a different data pipeline, a setting in which no size-generalization guarantee should be expected in the first place.

Width-independence of $\operatorname{Law}(\Gamma)$ excludes width-dependent learning rates.
In practice, the optimal learning rate typically differs across widths, so a family of weights at different widths that all perform reasonably well requires the learning-rate distribution to vary with width (e.g., in our INR-classification experiment, we constructed the SP-trained weight dataset via a per-width hyperparameter search).
However, this is not much of a problem under $\mu$P, the parameterization we assume in the theoretical guarantees stated later: under $\mu$P, the optimal learning rate is observed to stay close across widths~\citep{yang2021tuning}; see~\Cref{fig:mup-hp-transfer}.

Width-independence of $\operatorname{Law}(\Gamma)$ also forces the training horizon to follow one law across widths, since the horizon is a coordinate of $\Gamma$, so the number of epochs cannot grow with the network's width.
This is typically violated in practice as well, since larger models usually take longer to converge and are therefore trained for more epochs.

Furthermore, the independence of $S,\Gamma,\xi_n$ forbids any feedback from a run into its own hyperparameters, such as a validation-selected stopping time or a restart after a diverged run.

We first review, in \Cref{sec:mup-preliminaries}, the Tensor Programs machinery that we use to analyze the limiting behavior of $\mu_n$ as $n\to\infty$.
We then state and prove our two size-generalization guarantees: \Cref{thm:guarantee-1} is proved in \Cref{sec:one-layer-mup}, and \Cref{thm:guarantee-2} is proved in \Cref{app:guarantee-2}.

\subsection{Preliminary: Tensor Programs and \texorpdfstring{$\mu$P}{muP}}
\label{sec:mup-preliminaries}
\paragraph{Tensor programs.}
We briefly review the Tensor Programs framework developed in~\citet{yang2023tensorprogramsivb}.
This framework gives a rigorous way to characterize infinite-width limits for a broad class of neural-network computations, including training dynamics of all standard architectures using standard entry-wise optimizers.

The framework applies to the following setting:
The dataset is fixed, and the number of training steps $T$ is fixed independently of the width $n$.
Randomness enters only through the initialization and then propagates to every downstream quantity, including activations, gradients, and weight updates.
We are interested in the limiting behavior of these quantities (as random variables) as the width $n\to\infty$.
To invoke this framework in our setting, in later analysis we condition on $S=s$ and $\Gamma=\gamma$ and regard $\theta_n=\mathsf{Train}_n(\gamma,s,\xi_n)$, for fixed $(\gamma,s)$, as the width-$n$ instantiation of a single tensor program whose only randomness is $\xi_n$.
This is exactly why the data model above required $\operatorname{Law}(S)$ and $\operatorname{Law}(\Gamma)$ to be independent of $n$, and $S,\Gamma$ to be independent of $\xi_n$: these ensure that conditioning on $(\gamma,s)$ gives, for every $n$, the same training rule $\mathsf{Train}_n(\gamma,s,\cdot)$ applied to $\xi_n$ with its unconditional law.
Without them, conditioning on $(S,\Gamma)$ could change the training rule with $n$ or distort the law of $\xi_n$, and the Tensor Programs machinery above would not apply.
Conditioning alone is not enough, however: the runs at different widths form a single tensor program only if the parameterization is \emph{stable}, a requirement we discuss at the end of this subsection.

The high-level idea of the Tensor Program framework is the following.
The basic object is a \emph{tensor program}, an abstract programming language for computations built from random initialization, matrix multiplication, averaging over the width coordinate, and entrywise nonlinearities.
See \Cref{def:tensor-program} below.
The Tensor Programs papers show, first, that many neural computations used in practice, including convolutions, attention, batch and layer normalizations, and residual connections, can be expressed as tensor programs; see Appendix~A of~\cite{yang2019wide}.
They also show that if an architecture's forward pass can be written as a tensor program, then its backward pass and a fixed finite number of weight updates can be written as a tensor program; see Section~2.9.3 of~\cite{yang2023tensorprogramsivb}.
Thus, to understand finite-time infinite-width training dynamics, it suffices to characterize the infinite-width limits of tensor programs.
For this purpose, one assigns to each tensor program a recursively defined collection of random variables and deterministic scalars that describe the limiting objects; see Definition~2.6.5 of~\cite{yang2023tensorprogramsivb}.
The master theorem then states the precise mode in which the finite-width program converges to these limits.

We now give the concrete form of the tensor programs that we need.

\begin{definition}[Tensor program, Definition~2.6.1 of \cite{yang2023tensorprogramsivb}]
\label{def:tensor-program}
A \emph{tensor program} of width $n$ generates a sequence $\bm x$ of vectors in $\RR^n$ and a sequence $\bm c$ of scalars in $\RR$, starting from an initial set $\bm c^0\subseteq\bm c$ of random scalars, an initial set $\bm x^0\subseteq\bm x$ of random vectors in $\RR^n$, and an initial set $\mc W$ of random $n\times n$ matrices.
At each step, one of the following three instructions is applied.
\begin{description}
\item[\textsc{Avg}] Choose a vector $x\in\bm x$ and append to $\bm c$ the scalar
\[
\frac1n\sum_{\alpha=1}^n x_\alpha\in\RR.
\]
\item[\textsc{MatMul}] Choose a matrix $W\in\mc W$ and a vector $x\in\bm x$, and append to $\bm x$ the vector
\[
Wx\in\RR^n\quad\text{or}\quad W^\top x\in\RR^n.
\]
\item[\textsc{OuterNonlin}] Choose an integer $r\ge0$ and a function $\psi\colon\RR^{\abs{\bm x}(r+1)+\abs{\bm c}}\to\RR$, and append to $\bm x$ the vector $y\in\RR^n$ with
\[
y_\alpha=\frac1{n^r}\sum_{\beta_1,\dots,\beta_r=1}^n\psi(\bm x_\alpha;\bm x_{\beta_1};\dots;\bm x_{\beta_r};\bm c),
\]
where $\bm x_\gamma\in\RR^{\abs{\bm x}}$ is the $\gamma$th row of $\bm x$, viewed as the $n\times\abs{\bm x}$ matrix whose columns are the vectors in $\bm x$.
\end{description}
\end{definition}

The vector variables record quantities with one dimension that scales with the width $n$, such as preactivations, activations, and their gradients.
The scalar variables record quantities whose dimension does not scale with the width, such as network outputs and losses.

Every vector in $\bm x$ has an associated limiting random variable, and every scalar $\theta\in\bm c$ has a deterministic limit $\mathring\theta$.
Both are defined by the recursion of Definition~2.6.5 of~\cite{yang2023tensorprogramsivb}.
We now state the master theorem of \citet{yang2023tensorprogramsivb} that we will use later for our analysis, which characterizes the precise sense in which the scalars $\bm c$ of a tensor program converge to their infinite-width limits $\mathring{\bm c}$ as $n\to\infty$.
This first requires some mild regularity assumptions on the random initialization and the nonlinearities used in the tensor program.

\begin{assumption}[Setup~2.6.3 of \cite{yang2023tensorprogramsivb}]
\label{ass:tp-setup}
In the tensor program of \Cref{def:tensor-program},
\begin{enumerate}[label=(\alph*),leftmargin=*,topsep=0pt,itemsep=0pt]
\item every entry of every initial matrix $W\in\mc W$ is i.i.d.\ $\mc N(0,1/n)$;
\item every entry of every initial vector in $\bm x^0$ is i.i.d.\ $\mc N(0,1)$;
\item the initial scalars satisfy $\bm c^0\asto0$ as $n\to\infty$; and
\item every $\psi\colon\RR^k\to\RR$ used in \textsc{OuterNonlin} is \emph{pseudo-Lipschitz}: there are constants $C,d<\infty$ such that
\[
\abs{\psi(x)-\psi(y)}
\le
C\norm[2]{x-y}
\left(
1+\sum_{i=1}^k\lvert x_i\rvert^{d}+\sum_{i=1}^k\lvert y_i\rvert^{d}
\right)
\qquad\text{for all }x,y\in\RR^k.
\]
\end{enumerate}
\end{assumption}

\begin{theorem}[Master theorem, Theorem~2.6.10 of \cite{yang2023tensorprogramsivb}]
\label{lem:tp-master}
Let $\bm c=\bm c(n)$ be any finite collection of scalars produced by a tensor program satisfying \Cref{ass:tp-setup}, with the size of this collection independent of $n$.
Let $\mathring{\bm c}$ be the deterministic vector assigned to these scalars by Definition~2.6.5 of~\cite{yang2023tensorprogramsivb}.
Then, as $n\to\infty$,
\[
\bm c\asto\mathring{\bm c}.
\]
\end{theorem}

We now state a corollary of the master theorem that shows empirical measures of vectors produced by a tensor program converge in Wasserstein distance to the law of their limiting random variables.
This puts the master theorem in a form that is more convenient for our later analysis of size generalization.

\begin{corollary}
\label{cor:tp-vector-tests}
Consider a tensor program satisfying \Cref{ass:tp-setup}.
Let $v^1_n,\dots,v^K_n\in\RR^n$ be any fixed finite collection of vectors produced by the program, with associated limiting random variables $V^1,\dots,V^K$.
Then the following hold.
\begin{enumerate}[label=(\roman*),leftmargin=*,topsep=0pt,itemsep=0pt]
\item For every pseudo-Lipschitz $\psi\colon\RR^K\to\RR$,
\[
\frac1n\sum_{i=1}^n
\psi\left(v^1_{n,i},\dots,v^K_{n,i}\right)
\xrightarrow{\mathrm{a.s.}}
\Ex\left[\psi(V^1,\dots,V^K)\right]
\in\RR.
\]
\item Define $\xi_{n,i}\colonequals(v^1_{n,i},\dots,v^K_{n,i})$ with empirical measure
$\widehat\nu_n
\colonequals
\frac1n\sum_{i=1}^n\delta_{\xi_{n,i}}$, and let 
$\nu_\infty
\colonequals
\operatorname{Law}(V^1,\dots,V^K)$.
Then for every $p\ge1$,
\[
W_p(\widehat\nu_n,\nu_\infty)\to0
\qquad\text{almost surely}.
\]
\end{enumerate}
\end{corollary}

\begin{proof}
For part (i), extend the program by applying \textsc{OuterNonlin} with $r=0$ to the selected vectors.
This produces the vector with entries $\psi(v^1_{n,i},\dots,v^K_{n,i})$.
By the \textsc{OuterNonlin} rule in Definition~2.6.5 of~\cite{yang2023tensorprogramsivb}, its limiting random variable is $\psi(V^1,\dots,V^K)$.
Applying \textsc{Avg} to this vector produces the scalar $\frac1n\sum_{i=1}^n\psi\left(v^1_{n,i},\dots,v^K_{n,i}\right)$, and the \textsc{Avg} rule assigns to it the deterministic limit $\Ex\psi(V^1,\dots,V^K)$.
Since $\psi$ is pseudo-Lipschitz, the extended program still satisfies \Cref{ass:tp-setup}.
\Cref{lem:tp-master} therefore applies to this scalar and gives the desired convergence.

For part (ii), part (i) implies that for every pseudo-Lipschitz $\psi\colon\RR^K\to\RR$,
\begin{equation}
\label{eq:tp-feature-test}
\int \psi\,\mathrm{d}\widehat\nu_n
=
\frac1n\sum_{i=1}^n\psi(\xi_{n,i})
\xrightarrow{\mathrm{a.s.}}
\Ex\psi(V^1,\dots,V^K)
=
\int \psi\,\mathrm{d}\nu_\infty,
\end{equation}
with finite limit.
Take a countable pseudo-Lipschitz convergence-determining class $\{\phi_j\}_{j\ge1}$ for weak convergence of probability measures on $\RR^K$; for example, the proof of Proposition~4.4 in~\cite{ethier2009markov} gives an explicit construction.
For every fixed $j$, the function $\phi_j$ is pseudo-Lipschitz, so \eqref{eq:tp-feature-test} holds with $\psi=\phi_j$ on an event of probability one.
Intersecting these countably many events gives an event of probability one on which
\[
\int\phi_j\,\mathrm{d}\widehat\nu_n
\to
\int\phi_j\,\mathrm{d}\nu_\infty
\qquad\text{for every }j.
\]
Therefore, $\widehat\nu_n\Rightarrow\nu_\infty$ almost surely, where $\Rightarrow$ denotes weak convergence of probability measures.

We now upgrade this weak convergence to Wasserstein convergence.
For each positive integer $q$, the function $\psi_q(x)=\norm[2]{x}^q$ is pseudo-Lipschitz.
Applying \eqref{eq:tp-feature-test} to these countably many functions and intersecting the corresponding probability-one events gives
\[
\int\norm[2]{x}^q\,\mathrm{d}\widehat\nu_n(x)
\to
\int\norm[2]{x}^q\,\mathrm{d}\nu_\infty(x)
\qquad\text{for every positive integer }q
\]
almost surely.
On the intersection with the weak-convergence event, Theorem~6.9 of~\cite{villani2009optimal} gives $W_q(\widehat\nu_n,\nu_\infty)\to0$ for every positive integer $q$.

Finally, given any real number $p\ge1$, choose an integer $q\ge p$.
Then, $W_p(\widehat\nu_n,\nu_\infty) \leq W_q(\widehat\nu_n,\nu_\infty)\to0$ almost surely.
\end{proof}

Beyond almost sure convergence, we now prove uniform moment bounds so that later this can be upgraded to convergence in $L^p$.
Notice that this lemma focuses on the case of matrix-free tensor programs, which is sufficient for our purposes later.
For general tensor programs, this requires stronger assumptions on the nonlinearities to be polynomially bounded (see e.g. Proposition 4.3.4 of \cite{yang2023tensorprogramsivb}).

\begin{lemma}[Uniform moment bound for matrix-free tensor programs]
\label{lem:tp-matrix-free-moments}
Consider a tensor program satisfying \Cref{ass:tp-setup}, and suppose that its matrix set $\mc W$ is empty.
Suppose additionally that the initial scalars have moments of all orders bounded uniformly in $n$.
Then, for every positive integer $p$, every vector $x=x(n)$ produced by the program, and every scalar $c=c(n)$ produced by the program,
\[
\sup_n
\Ex\left[\frac1n\sum_{\alpha=1}^n\abs{x_\alpha}^p\right]
<\infty
\qquad\text{and}\qquad
\sup_n\Ex\abs{c}^p<\infty.
\]
\end{lemma}

\begin{proof}
We argue by induction over the finite sequence of program instructions, proving the two bounds simultaneously for every positive integer $p$.

\emph{Base case.}
They hold for the initial vectors because their entries are standard Gaussian, and they hold for the initial scalars by the additional moment assumption.

\emph{Induction step.}
Consider first that an \textsc{Avg} instruction produces $c=\frac1n\sum_{\alpha=1}^n x_\alpha$.
Since the function $u\mapsto\abs{u}^p$ is convex, Jensen's inequality gives
$\Ex\abs{c}^p
\le
\Ex\left[\frac1n\sum_{\alpha=1}^n\abs{x_\alpha}^p\right],$
which is bounded uniformly in $n$ by the induction hypothesis.

Consider next that an \textsc{OuterNonlin} instruction produces
\[y_\alpha
=
\frac1{n^r}
\sum_{\beta_1,\dots,\beta_r=1}^n
\psi(\bm x_\alpha;\bm x_{\beta_1};\dots;\bm x_{\beta_r};\bm c).\]
Since $\psi$ is pseudo-Lipschitz (Assumption~\ref{ass:tp-setup}), the pseudo-Lipschitz inequality with $y=0$ gives, for every $z\in\RR^k$,
$
\abs{\psi(z)}
\le
\abs{\psi(0)}
+ C\norm[2]{z}\left(1+\sum_{i=1}^k\lvert z_i\rvert^{d}\right).
$
Since $k$ is fixed, we have $\norm[2]{z}\le \norm[1]{z}\le k t$, where $t=\max_j\abs{z_j}$, and $\sum_{i=1}^k\abs{z_i}^d\le k t^d$.
It follows that, for a constant $C_1<\infty$ independent of $z$,
\[
\abs{\psi(z)}
\le
\abs{\psi(0)}+Ckt(1+kt^d)
\le
C_1(1+t^{d+1})
\le
C_1\left(1+\sum_{j=1}^k\abs{z_j}^{d+1}\right).
\]
After raising this bound to the $p$th power and absorbing constants, for each positive integer $p$ there are a constant $C_p<\infty$ and a positive integer $q_p=p(d+1)$ such that for every $z\in\RR^k$,
$
\abs{\psi(z)}^p
\le
C_1^p\left(1+\sum_{j=1}^k\abs{z_j}^{d+1}\right)^p
\le
C_p\left(1+\sum_{j=1}^k\abs{z_j}^{q_p}\right).$

Applying Jensen's inequality yields
\begin{align*}
\Ex\left[\frac1n\sum_{\alpha=1}^n\abs{y_\alpha}^p\right]
&\le
\frac1{n^{r+1}}
\sum_{\alpha,\beta_1,\dots,\beta_r=1}^n
\Ex\abs{\psi(\bm x_\alpha;\bm x_{\beta_1};\dots;\bm x_{\beta_r};\bm c)}^p.
\end{align*}
Write the previously produced vectors and scalars that enter this instruction as $\bm x=(x^1,\dots,x^m)$ and $\bm c=(c^1,\dots,c^s)$, so that $\bm x_\alpha=(x^1_\alpha,\dots,x^m_\alpha)$.
Substituting the bound on $\abs{\psi(z)}^p$ into the right-hand side above gives
\begin{align*}
&
\frac1{n^{r+1}}
\sum_{\alpha,\beta_1,\dots,\beta_r=1}^n
\Ex\abs{\psi(\bm x_\alpha;\bm x_{\beta_1};\dots;\bm x_{\beta_r};\bm c)}^p
\\
&\le
C_p
\Biggl(
1
+
\sum_{a=1}^m
\frac1{n}
\sum_{\alpha=1}^n
\Ex\abs{x^a_\alpha}^{q_p}
+
\sum_{j=1}^r
\sum_{a=1}^m
\frac1{n}
\sum_{\beta_j=1}^n
\Ex\abs{x^a_{\beta_j}}^{q_p}
+
\sum_{b=1}^s
\Ex\abs{c^b}^{q_p}
\Biggr).
\end{align*}
Hence there is a constant $C'_p<\infty$, independent of $n$, such that the right-hand side above is bounded by
\[
C'_p
\left(
1
+
\sum_{a=1}^m
\Ex\left[\frac1n\sum_{\gamma=1}^n\abs{x^a_\gamma}^{q_p}\right]
+
\sum_{b=1}^s
\Ex\abs{c^b}^{q_p}
\right).
\]
The induction hypothesis bounds each of these finitely many quantities uniformly in $n$.

These are all possible instructions because $\mc W$ is empty, so the induction proves the claim.
\end{proof}

\paragraph{$\mu$P and hyperparameter transfer.}
Using the Tensor Programs framework, \citet{yang2023tensorprogramsivb} proves that $\mu$P is the unique parameterization in a large class of width-dependent parameterizations that satisfies a range of desirable properties.
In this sense, it is the ``optimal''  parameterization in the infinite-width limit.
\Cref{tab:width-scalings} summarizes the concrete scalings for SP and $\mu$P.

\ifdefined\arxivversion\else
\begin{center}
\captionsetup{type=table,hypcap=false, font=small,skip=2pt}
\caption{Initialization variance and example hyperparameters for different MLP parameters under SP and {\color{purple}$\mu$P} trained with SGD and Adam~\citep{yang2021tuning}.
$\mu$P is also defined for any standard architecture and prescribe scalings for weight-decay coefficients. See~\cite{yang2023tensorprogramsivb} for details. In both cases, the output bias $b^{(L+1)}$ is initialized to zero with constant learning rate. For a weight matrix $W$ taking inputs $x$ as $Wx$, fan\_in and fan\_out are the number of columns and rows respectively.}
\label{tab:width-scalings}
\small
\newcommand{\paramcell}[2]{${\color{black}#1}\quad{\color{purple}#2}$}
\resizebox{\textwidth}{!}{%
\begin{tabular}{@{}c c c c@{}}
\toprule
& \shortstack[c]{Init variance}
& \shortstack[c]{SGD learning rate}
& \shortstack[c]{AdamW learning rate}\\
\midrule
Input weights $W^{(1)}$; Biases $(b^{(\ell)})_{\ell=1}^{L}$
& \paramcell{1/\text{fan\_in}}{1/\text{fan\_in}}
& \paramcell{1}{\text{fan\_out}}
& \paramcell{1}{1}\\
Hidden weights $(\Wl)_{\ell=2}^{L}$
& \paramcell{1/\text{fan\_in}}{1/\text{fan\_in}}
& \paramcell{1}{1}
& \paramcell{1}{1/\text{fan\_in}}\\
Output weights $W^{(L+1)}$
& \paramcell{1/\text{fan\_in}}{1/\text{fan\_in}^2}
& \paramcell{1}{1/\text{fan\_in}}
& \paramcell{1}{1/\text{fan\_in}}\\
\bottomrule
\end{tabular}%
}
\vspace{-6pt}
\end{center}
\fi

\begin{figure}[h]
\centering
\includegraphics[width=0.9\textwidth]{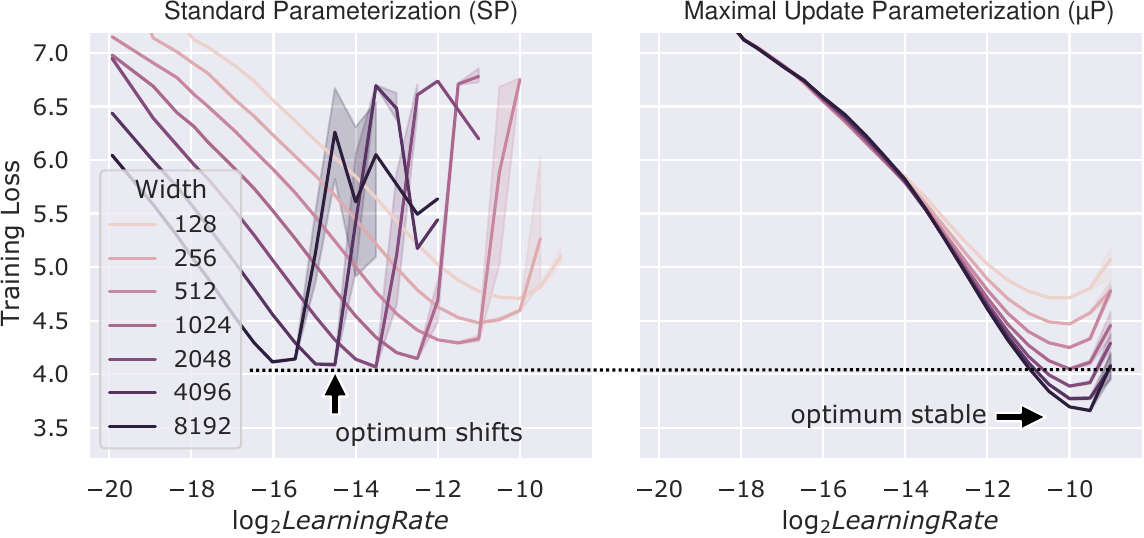}
\caption[Hyperparameter transfer under muP.]{Hyperparameter transfer under $\mu$P, Figure~1 of \citet{yang2021tuning}.
Training loss against learning rate for Transformers of width ranging over $128,\dots,8192$, trained with Adam.}
\label{fig:mup-hp-transfer}
\end{figure}

Furthermore, \citet{yang2021tuning} empirically observed that $\mu$P allows hyperparameters tuned at a small width to be transferred to larger widths without retuning, a phenomenon called \emph{hyperparameter transfer}.
This is shown in \Cref{fig:mup-hp-transfer}, where the optimal learning rate is stable across a wide range of widths under $\mu$P, but moves considerably under SP.
This is precisely because under $\mu$P the training dynamics of the networks of different widths are similar to each other, as they converge to the same well-behaved limit with controlled finite approximation error.

\subsection{Proof of \Cref{thm:guarantee-1}}
\label{sec:one-layer-mup}

Throughout this subsection the input MLP has $L=1$ layer, its input and output dimensions $\din,\dout$ are fixed, and its single hidden width is $n$, so its forward pass is
\begin{equation*}
f_n^\theta(u)=W^{(2)}\sigma\left(W^{(1)}u+b^{(1)}\right)+b^{(2)},
\end{equation*}
with $W^{(1)}\in\RR^{n\times\din}$, $b^{(1)}\in\RR^n$, $W^{(2)}\in\RR^{\dout\times n}$, and $b^{(2)}\in\RR^{\dout}$.
Recall that the fan-in rescaled coordinates are $\overline W^{(1)}=\din W^{(1)}$ and $\overline W^{(2)}=nW^{(2)}$.

In this 1-layer MLP setting, $\mu$P is exactly the classical mean-field scaling.
Hence the results in this part could be proved directly using mean-field limit results such as~\citep{mei2018meanfield,chizat2018global} instead of the Tensor Programs framework.
We nevertheless use the Tensor Programs formulation below to keep the proof strategy consistent with the later size-generalization result, where the same machinery is needed.

We first give a high-level explanation in \Cref{sec:one-layer-mup-intuition}, where we also compare $\mu$P with SP and explain why the parameterization matters.
It is followed by the detailed proof in \Cref{sec:one-layer-mup-proof}.
Finally, \Cref{sec:obstruction-larger-depths} shows the obstruction to extending this result to MLPs with $\geq 2$ layers.
\subsubsection{A high-level explanation}\label{sec:one-layer-mup-intuition}
We first explain intuitively why we expect $W_1(\mu_n, \mu_m; d_\infty^{\op})\to 0 $ if the training of the networks is done under $\mu$P, and why this is not expected to hold under SP.

\paragraph{What $\mu$P prescribes, and why it makes the widths agree.}
What $\mu$P (\Cref{tab:width-scalings}) prescribes for the 1-layer MLP is summarized below:
\[
\begin{array}{r|cccc}
\mu\text{P} & W^{(1)} & b^{(1)} & W^{(2)} & b^{(2)}\\
\hline
\text{initial variance} & 1/\din & 1/\din & 1/n^2 & 0\\
\text{SGD rate} & n & n & 1/n & 1\\
\text{AdamW rate} & 1 & 1 & 1/n & 1
\end{array}
\]
At initialization, $\overline W^{(1)}=\din W^{(1)}$, $b^{(1)}$, and $\overline W^{(2)}=nW^{(2)}$ all have i.i.d.\ entries with distributions independent of $n$.
Grouping the entries attached to one and the same hidden neuron, the $n$ vectors
\begin{equation}
\label{eq:zeta-coords}
\zeta_{n,i}(\theta)
\colonequals
\left(
\frac{\overline W^{(1)}_{i,:}}{\sqrt{\din}},\,
b^{(1)}_i,\,
\frac{\overline W^{(2)}_{:,i}}{\sqrt{\dout}}
\right)
\in\RR^{\din+1+\dout}
\qquad(i\in[n])
\end{equation}
are i.i.d.\ draws from one Gaussian law that does not depend on $n$.
Meanwhile, the output bias $b^{(2)}\in\RR^{\dout}$ is initialized to zero.
A freshly initialized width-$n$ network is therefore nothing more than $n$ independent samples from one fixed law, together with a single $\dout$-dimensional zero-vector.
A wider network draws more samples from the same law; it does not change the law.

The learning rate scalings prescribed by $\mu$P are exactly what is needed to preserve this picture through training, rather than only at initialization.
At every fixed step $t$, the $n$ vectors $\zeta_{n,i}(\theta_n(t))$ attached to the hidden neurons $i\in[n]$ still ``behave like'' i.i.d.\ draws from a law that depends on $t$ but not on $n$.
The precise sense of ``behave like'' is that the empirical distribution of the $n$ vectors converges almost surely as $n\to\infty$: there exists some limiting distribution $\nu_\infty(t)$ on $\RR^{\din+1+\dout}$ such that
\[W_2\left(\frac1n\sum_{i=1}^n\delta_{\zeta_{n,i}(\theta_n(t))},\,\nu_\infty(t)\right)\asto0\]
as $n\to\infty$, where $W_2$ is the Wasserstein-2 distance.
Moreover, the behavior of the output bias during the training is also predictable: there exists some limiting $b^{(2)}_\infty(t)\in\RR^{\dout}$ such that $b^{(2)}_n(t)\asto b^{(2)}_\infty(t)$ as $n\to\infty$.
The Tensor Programs framework allows us to precisely describe the limiting law $\nu_\infty(t)$ and the limiting bias $b^{(2)}_\infty(t)$ at every step $t$, and \Cref{cor:tp-vector-tests} gives the rigorous proof of the convergence.

It remains to relate this convergence to the closeness we actually want, namely closeness of two trained networks of different widths $n$ and $m$ in $d_\infty^{\op}$, the symmetrized distance induced by the normalized spectral norm.
It turns out that in the 1-layer MLP case the distance in $d_\infty^{\op}$ is controlled by the Wasserstein-2 distance between the two empirical distributions of the vectors $\zeta_{n,i}$ together with the difference between the two output biases.
Both of these are small, since by the above each of the two empirical distributions is close to the common $\nu_\infty(t)$ and each output bias is close to the common $b^{(2)}_\infty(t)$.
Hence the trained networks of widths $n$ and $m$ are close in $d_\infty^{\op}$, which in turn gives the desired closeness of $\mu_n$ and $\mu_m$ in $W_1(\cdot; d_\infty^{\op})$.

\paragraph{What SP prescribes, and why the widths disagree already at initialization.}
The same specialization of \Cref{tab:width-scalings} gives
\[
\begin{array}{r|cccc}
\text{SP} & W^{(1)} & b^{(1)} & W^{(2)} & b^{(2)}\\
\hline
\text{initial variance} & 1/\din & 1/\din & 1/n & 0\\
\text{SGD rate} & 1 & 1 & 1 & 1\\
\text{AdamW rate} & 1 & 1 & 1 & 1
\end{array}.
\]

Let $\mu_n(0)\colonequals\operatorname{Law}(\theta_n(0))$ denote the initialization law at width $n$.
We will show that, already at initialization, the sequence $(\mu_n(0))_n$ is not Cauchy in $W_1(\,\cdot\,,\,\cdot\,;d_\infty^{\op})$ under SP.
With $\dout$ fixed, the standard random matrix estimate for the largest singular value gives $\Ex\norm{W^{(2)}(0)}\to1$, and in particular $\Ex\norm{W^{(2)}(0)}=\Theta(1)$.
As such,
\begin{align*}
\Ex\norm[V_n,\op]{\theta_n(0)}= \sqrt{\frac n{\dout}}\Ex\norm{W^{(2)}(0)}+O(1)= \Theta(\sqrt n),
\end{align*}
because the first-layer and bias contributions are $O(1)$.
In other words, the initialized weights of different widths are not expected to be close, as they are farther and farther away from the origin in the normalized spectral norm as $n\to\infty$.

As a result, for every coupling $\pi$ of $\mu_n(0)$ and $\mu_{2n}(0)$, if $(\theta_n(0),\theta_{2n}(0))$ denotes the pair sampled from $\pi$, then the reverse triangle inequality gives
\begin{align*}
\Ex_{\pi}d_\infty^{\op}(\theta_n(0),\theta_{2n}(0))
\ge
\abs{\Ex\norm[V_{2n},\op]{\theta_{2n}(0)}-\Ex\norm[V_n,\op]{\theta_n(0)}}
=\Theta(\sqrt n),
\end{align*}
where we used that $d_\infty^{\op}(\theta,0)=\norm[V_r,\op]{\theta}$ for $\theta\in V_r$.
Hence,
\[
W_1\left(\mu_n(0),\mu_{2n}(0);d_\infty^{\op}\right)\to \infty.
\]

The same initialization argument applies at every depth $L\ge1$, showing that SP-initialized weights are generally far apart in $d_\infty^{\op}$.
The training process is unlikely to remedy this, as we numerically checked in various experiments.

\subsubsection{Proofs of the 1-layer results}\label{sec:one-layer-mup-proof}
We first state the additional regularity assumptions on the 1-layer MLP model and its training.
These assumptions ensure that the entire training process is a tensor program satisfying \Cref{ass:tp-setup}, so that \Cref{lem:tp-master,cor:tp-vector-tests} apply.

\begin{assumption}
\label{ass:tp-regularity}
\begin{enumerate}[label=(\roman*),leftmargin=*,topsep=0pt,itemsep=0pt]
\item The weights and biases are initialized with i.i.d. Gaussian entries.
\item Conditional on every $(\Gamma,S)=(\gamma,s)$ in the support of $(\Gamma,S)$, the training horizon $T=T(\gamma)$ is finite and independent of the width.
\item The optimizer is an entrywise optimizer with update function $Q_t\colon\RR^{t+1}\to\RR$ as defined in Section~2.1 of~\cite{yang2023tensorprogramsivb}, which includes SGD and Adam as special cases.
\item The activation $\sigma$, its derivative $\sigma'$, the derivative of the training loss, and the optimizer maps $Q_t$ for $t<T$ are pseudo-Lipschitz in the sense of \Cref{ass:tp-setup}(d).
\end{enumerate}
\end{assumption}

These assumptions are inherited from the Tensor Programs framework.
They are for technical convenience rather than a fundamental restriction, and are likely to be removed or relaxed as noted in~\cite{yang2023tensorprogramsivb}.

\begin{proposition}[Trained weights at large widths agree in the normalized spectral distance]
\label{prop:sizegen-w1-convergence}
Assume that \Cref{ass:tp-regularity} holds.
Suppose that there exists $q>1$ such that
\[
\sup_{(\gamma,s)\in \operatorname{supp}(\Gamma,S)}
\sup_{n\ge1}
\Ex_{\xi_n}\|\theta_n\|_{V_n,\fro}^q
<\infty.
\]
Then, for every $p\in[1,q)$,
\begin{equation}
\label{eq:normalized-spectral-norm-convergence}
\sup_{m\ge n}W_p(\mu_n,\mu_m;d_\infty^{\op})\longrightarrow 0
\qquad\text{as }n\to\infty.
\end{equation}
\end{proposition}

\begin{proof}
We first fix $(\gamma,s)$ in the support of $(\Gamma,S)$ and condition on $(\Gamma,S)=(\gamma,s)$.
This fixes the hyperparameters $\gamma$ and the auxiliary training randomness $s$, so that the initialization is the only remaining source of randomness.
Throughout the conditional argument, $\theta_n$ denotes $\mathsf{Train}_n(\gamma,s,\xi_n)$.
According to our set-up, after conditioning on $(\Gamma,S)=(\gamma,s)$, the family $(\xi_n)_{n\ge1}$ remains independent, and all almost-sure statements below are interpreted with respect to this conditional coupling.

\emph{The run as a tensor program.}
Under \Cref{ass:tp-regularity}, training the 1-layer MLP under $\mu$P is a tensor program satisfying \Cref{ass:tp-setup}.
See Section~2.1 of~\cite{yang2020feature} for the construction for the bias-free case trained with SGD.
Notice that this tensor program requires no random $n\times n$ matrix.
The $\mu$P scalings ensure that $\overline W^{(1)}$, $b^{(1)}$, and $\overline W^{(2)}$ have entries of order one throughout every fixed finite training run, and Definition~2.6.5 of~\cite{yang2023tensorprogramsivb} assigns limiting random variables to these entries.

Append to the tensor program the vectors $\overline W^{(1)}_{:,a}(T)/\sqrt{\din}$ for $a\in[\din]$ and $\overline W^{(2)}_{b,:}(T)/\sqrt{\dout}$ for $b\in[\dout]$.
Each is produced from an existing vector by a linear \textsc{OuterNonlin} instruction, so the extended program still satisfies \Cref{ass:tp-setup}.
Together with $b^{(1)}(T)$, these vectors have the joint coordinates $\zeta_{n,i}(\theta_n(T))$ defined in \eqref{eq:zeta-coords}.
Applying \Cref{cor:tp-vector-tests}(ii) with $K=\din+1+\dout$ gives
\[
W_2\left(\frac{1}{n}\sum_{i=1}^n \delta_{\zeta_{n,i}(\theta_n(T))},\nu_\infty(T)\right)\asto 0,
\]
where $\nu_\infty(T)$ is the conditional limiting probability measure on $\RR^{\din+1+\dout}$.
The entries of $b^{(2)}(T)$ are scalars of the program, so \Cref{lem:tp-master} gives $b^{(2)}_n(T)\asto b^{(2)}_\infty(T)$ for a limiting output bias $b^{(2)}_\infty(T) \in \RR^{\dout}$.

\emph{Almost-sure convergence in $d_\infty^{\fro}$, and hence in $d_\infty^{\op}$.}
The width-$n$ weight space is $V_n\colonequals \RR^{n\times \din} \oplus \RR^n\oplus \RR^{\dout\times n}\oplus\RR^{\dout}$.
Since the hidden-neuron permutations act on all the $n$-dimensional coordinates simultaneously, it is equivalent to regard $V_n$ as $\RR^{n(\din+1+\dout)}\oplus\RR^{\dout}$.
Recall that the normalized Frobenius norm \eqref{eq:norm-fro} at $L=1$ reads
\[
\|\theta\|_{V_n,\fro}^2=\frac{1}{n} \|\zeta_{n}(\theta)\|_\fro^2 + \frac{1}{\dout}\|b_\theta^{(2)}\|_2^2.
\]
The corresponding symmetrized distance $d_\infty^{\fro}$ on the limit space coincides with the Wasserstein-$2$ distance between the empirical measures of the stacked hidden-neuron features, combined in $\ell_2$ with the output-bias distance (Appendix~E.1 of \cite{levin2025transferring}): for $\theta\in V_n$ and $\widetilde\theta\in V_m$,
\[
\left(d_\infty^{\fro}(\theta,\widetilde\theta)\right)^2
=
W_2^2\left(\frac1n\sum_{i=1}^n \delta_{\zeta_{n,i}(\theta)},\frac1m\sum_{i=1}^m \delta_{\zeta_{m,i}(\widetilde\theta)}\right)
 + \frac{1}{\dout}\|b_\theta^{(2)}-b_{\widetilde\theta}^{(2)}\|_2^2,
\]
where $W_2$ is computed on $\RR^{\din+1+\dout}$ with ground metric $d_\zeta(u,v)=\norm[2]{u-v}$.

Combining with the Tensor Programs convergence results above, we have
\[
\begin{aligned}
d_\infty^{\fro}(\theta_n(T),\theta_m(T))
&\leq
W_2\left(\frac{1}{n}\sum_{i=1}^n \delta_{\zeta_{n,i}(\theta_n(T))},\nu_\infty(T)\right)
 + W_2\left(\frac{1}{m}\sum_{i=1}^m \delta_{\zeta_{m,i}(\theta_m(T))},\nu_\infty(T)\right)\\
&\quad
 + \frac{1}{\sqrt{\dout}}\|b_n^{(2)}(T)-b^{(2)}_\infty(T)\|_2
 + \frac{1}{\sqrt{\dout}}\|b_m^{(2)}(T)-b^{(2)}_\infty(T)\|_2
\ \asto\ 0
\end{aligned}
\]
as $\min\{n,m\}\to\infty$.
Then, by \Cref{prop:op-fro-comparison} which relates $d_\infty^{\op}$ and $d_\infty^{\fro}$, we have
\[
d_\infty^{\op}(\theta_n(T),\theta_m(T))
\le 2d_\infty^{\fro}(\theta_n(T),\theta_m(T))\asto 0.
\]

\emph{Convergence in $L^p$.}
Again by \Cref{prop:op-fro-comparison} which relates $d_\infty^{\op}$ and $d_\infty^{\fro}$, we have
\[
d_\infty^{\op}(\theta_n(T),\theta_m(T))
\le 2d_\infty^{\fro}(\theta_n(T),\theta_m(T))
\le
2\left(\norm[V_n,\fro]{\theta_n(T)}+\norm[V_m,\fro]{\theta_m(T)}\right).
\]
Raising this to the $q$th power and using the bound $(a+b)^q\le 2^{q-1}(a^q+b^q)$ for $a,b\ge0$ and $q>1$, we have
\begin{equation}
\label{eq:sizegen-qmoment-bound}
\sup_{(\gamma,s)\in\operatorname{supp}(\Gamma,S)}
\sup_{n,m\ge1}
\Ex_{\xi}
d_\infty^{\op}(\theta_n(T),\theta_m(T))^q
\le
2^{2q}
\sup_{(\gamma,s)\in\operatorname{supp}(\Gamma,S)}
\sup_{n\ge1}
\Ex_{\xi_n}\norm[V_n,\fro]{\theta_n}^q
<\infty.
\end{equation}
Fix $p\in[1,q)$.
The preceding display and the de la Vallée--Poussin criterion show that the family
\[
\left\{
d_\infty^{\op}(\theta_n(T),\theta_m(T))^p
:
n,m\ge1
\right\}
\]
is uniformly integrable, because its $(q/p)$th moments are uniformly bounded and $q/p>1$.
The almost-sure convergence established above and the Vitali convergence theorem therefore give
\[
\Ex_{\xi}
d_\infty^{\op}(\theta_n(T),\theta_m(T))^p
\longrightarrow0
\]
as $\min\{n,m\}\to\infty$, conditional on $(\Gamma,S)=(\gamma,s)$.

\emph{Conclusion.}
Couple $\mu_n$ and $\mu_m$ as follows: draw a single pair $(\Gamma,S)$, and then, conditionally on it, draw the two initializations $\xi_n$ and $\xi_m$ independently.
Since $W_p$ is an infimum over couplings,
\[
W_p(\mu_n,\mu_m;d_\infty^{\op})^p
\le
\Ex_{\Gamma,S}\Ex_{\xi}
d_\infty^{\op}(\theta_n,\theta_m)^p.
\]
The previous step shows that the inner expectation tends to $0$ as $\min\{n,m\}\to\infty$, for every fixed $(\gamma,s)$ in the support of $(\Gamma,S)$.
We now bound the inner expectation uniformly in $(\gamma,s)$ and in $n,m$.
Let $C<\infty$ denote the left-hand side of \eqref{eq:sizegen-qmoment-bound}.
Since $p<q$, Jensen's inequality applied to the concave map $t\mapsto t^{p/q}$ on $[0,\infty)$ gives
\[
\Ex_{\xi}
d_\infty^{\op}(\theta_n,\theta_m)^p
\le
\left(\Ex_{\xi}d_\infty^{\op}(\theta_n,\theta_m)^q\right)^{p/q}
\le
C^{p/q},
\]
a bound independent of $(\gamma,s)$ and of $n,m$.
So the dominated convergence theorem applies and yields
\[
\Ex_{\Gamma,S}\Ex_{\xi}
d_\infty^{\op}(\theta_n,\theta_m)^p
\longrightarrow0,
\]
and hence $W_p(\mu_n,\mu_m;d_\infty^{\op})\to0$, whenever $\min\{n,m\}\to\infty$.
Therefore, for every $\varepsilon>0$, there exists $N\ge1$ such that
\[
W_p(\mu_n,\mu_m;d_\infty^{\op})<\varepsilon
\]
whenever $n,m\ge N$.
If $n\ge N$, this bound holds for every $m\ge n$.
Taking the supremum over $m\ge n$ proves \eqref{eq:normalized-spectral-norm-convergence}.
\end{proof}

\begin{corollary}[Size generalization on $\mu$P-trained 1-layer MLPs]
\label{cor:sizegen-one-layer}
Assume the hypotheses of \Cref{prop:sizegen-w1-convergence} for some $q>2$.
Let the activation $\sigma$ and all components of the duplication-compatible GMN $f^\Theta$ be Lipschitz.
Let $X\subseteq\RR^{\din}$ be compact, let the target functional $F^*\colon\mathcal C(X,\RR^{\dout})\to\RR^k$ be Lipschitz with respect to the uniform norm, and let the loss $\ell\colon\RR^k\times\RR^k\to\RR$ be Lipschitz in each argument.
Define the risk
\[
\mc R_n(f^\Theta)
\colonequals
\Ex_{\theta\sim\mu_n}
\ell\left(f^\Theta(\theta),F^*(\mc R(\theta))\right).
\footnote{Note the slight abuse of notation here: $\mc R_n$ (subscripted, the risk at width $n$) and $\mc R$ (unsubscripted, the realization map of \Cref{prop:misalignment}) denote different objects sharing the letter $\mc R$.}
\]
Then
\[
\sup_{m\ge n}\left|\mc R_n(f^\Theta)-\mc R_m(f^\Theta)\right|\longrightarrow0
\qquad\text{as }n\to\infty .
\]
\end{corollary}

\begin{proof}
\emph{Pointwise bound.}
The proof of \Cref{prop:misalignment}(i) shows that the realization map $\mc R$ is $L_R$-Lipschitz on every ball of radius $R$ with respect to $\norm[\Vinf,\op]{\cdot}$.
For $L=1$, the recursions defining $C_\ell$ and $C'_\ell$ in that proof show that $C_0$ and $C'_1$ are independent of $R$, that $C_1$ grows at most linearly in $R$, and hence that $L_R=\sqrt{\dout}C'_2$ grows at most linearly in $R$.
Applying this ballwise estimate with $R=\max\{\norm[\Vinf,\op]{\theta},\norm[\Vinf,\op]{\widetilde\theta}\}$ shows that there is a constant $C_{\mc R}<\infty$ such that
\[
\norm[\infty]{\mc R(\theta)-\mc R(\widetilde\theta)}
\le
C_{\mc R}
\left(
1+\norm[\Vinf,\op]{\theta}+\norm[\Vinf,\op]{\widetilde\theta}
\right)
\norm[\Vinf,\op]{\theta-\widetilde\theta}
\]
for every $\theta,\widetilde\theta\in\Vinf$.
Both $\mc R$ and the normalized spectral norm are invariant under hidden-neuron permutations, so applying the preceding inequality to $g\cdot\theta$ and $\widetilde\theta$, and then taking the infimum over $g\in\Ginf$, gives
\[
\norm[\infty]{\mc R(\theta)-\mc R(\widetilde\theta)}
\le
C_{\mc R}
\left(
1+\norm[\Vinf,\op]{\theta}+\norm[\Vinf,\op]{\widetilde\theta}
\right)
d_\infty^{\op}(\theta,\widetilde\theta).
\]

By \Cref{prop:misalignment}(ii), the duplication-compatible GMN is $L_\Theta$-Lipschitz with respect to $\norm[\Vinf,1]{\cdot}$.
For 1-layer input MLPs, \Cref{prop:one-layer-op-domination} gives $\norm[\Vinf,1]{\theta'}\le\kappa\norm[\Vinf,\op]{\theta'}$ for every $\theta'\in\Vinf$, where $\kappa\colonequals\max\{1,\sqrt{\din},\sqrt{\dout}\}$.
Combining this domination with the invariance of the GMN under hidden-neuron permutations gives
\[
\norm[2]{f^\Theta(\theta)-f^\Theta(\widetilde\theta)}
\le
\kappa L_\Theta d_\infty^{\op}(\theta,\widetilde\theta).
\]

Write $L_{\mathrm{loss}}$ for a Lipschitz constant of $\ell$ in each argument, $L_*$ for one of $F^*$, and set $h(\theta)\colonequals\ell\left(f^\Theta(\theta),F^*(\mc R(\theta))\right)$.
The two preceding bounds and the triangle inequality give a constant $C<\infty$ such that
\[
\abs{h(\theta)-h(\widetilde\theta)}
\le
C
\left(
1+\norm[\Vinf,\op]{\theta}+\norm[\Vinf,\op]{\widetilde\theta}
\right)
d_\infty^{\op}(\theta,\widetilde\theta).
\]
Taking $\widetilde\theta=0$ and using the assumed $q$th-moment bound shows that $h$ is integrable under every $\mu_n$.

\emph{Taking expectations.}
Set $p_*\colonequals q/(q-1)$, so that $q$ and $p_*$ are conjugate exponents.
Because $q>2$, we have $1<p_*<2<q$, so \Cref{prop:sizegen-w1-convergence} applies with $p=p_*$.

We have $\sup_{n\ge1}\Ex_{\theta\sim\mu_n}\norm[\Vinf,\fro]{\theta}^q<\sup_{(\gamma,s)\in\operatorname{supp}(\Gamma,S)}\sup_{n\geq 1}\Ex_{\xi_n}\norm[V_n,\fro]{\theta_n}^q<\infty$.
Combined with the domination $\norm[\Vinf,\op]{\theta}\le\sqrt{2(L+1)}\norm[\Vinf,\fro]{\theta}$ of \Cref{prop:op-fro-comparison}, this gives $\sup_{n\ge1}\Ex_{\theta\sim\mu_n}\norm[\Vinf,\op]{\theta}^q<\infty$.
Consequently, if $\pi$ is any coupling of $\mu_n$ and $\mu_m$, then
\[
\Ex_\pi
\left(
1+\norm[\Vinf,\op]{\theta}+\norm[\Vinf,\op]{\widetilde\theta}
\right)^q
\le
3^{q-1}
\left(
1
+\Ex_{\theta\sim\mu_n}\norm[\Vinf,\op]{\theta}^q
+\Ex_{\widetilde\theta\sim\mu_m}\norm[\Vinf,\op]{\widetilde\theta}^q
\right)
\]
is bounded by a constant independent of $n$, $m$, and $\pi$.

For every coupling $\pi$ of $\mu_n$ and $\mu_m$, the coupling identity and the triangle inequality give $\abs{\mc R_n(f^\Theta)-\mc R_m(f^\Theta)}\le\Ex_\pi\abs{h(\theta)-h(\widetilde\theta)}$.
Applying the pointwise bound on $\abs{h(\theta)-h(\widetilde\theta)}$ established above, and then H\"{o}lder's inequality with conjugate exponents $q$ and $p_*$, gives
\[
\begin{aligned}
\abs{\mc R_n(f^\Theta)-\mc R_m(f^\Theta)}
&\le
C
\Ex_\pi
\left[
\left(
1+\norm[\Vinf,\op]{\theta}+\norm[\Vinf,\op]{\widetilde\theta}
\right)
d_\infty^{\op}(\theta,\widetilde\theta)
\right] \\
&\le
C
\left(
\Ex_\pi
\left(
1+\norm[\Vinf,\op]{\theta}+\norm[\Vinf,\op]{\widetilde\theta}
\right)^q
\right)^{1/q}
\left(
\Ex_\pi
d_\infty^{\op}(\theta,\widetilde\theta)^{p_*}
\right)^{1/p_*} \\
&\le
C'
\left(
\Ex_\pi
d_\infty^{\op}(\theta,\widetilde\theta)^{p_*}
\right)^{1/p_*},
\end{aligned}
\]
where $C'<\infty$ is independent of $n,m$ and $\pi$.
Taking the infimum over $\pi$ gives $\abs{\mc R_n(f^\Theta)-\mc R_m(f^\Theta)}\le C'W_{p_*}(\mu_n,\mu_m;d_\infty^{\op})$.
The uniform convergence in \eqref{eq:normalized-spectral-norm-convergence} now proves the result.
\end{proof}

\subsubsection{Obstruction on MLP with larger depths}\label{sec:obstruction-larger-depths}
The proof in \Cref{sec:one-layer-mup-proof} crucially relies on the fact that the MLP is 1-layer, and in particular that there is no hidden-to-hidden weight matrix.
We now show the obstruction to generalize the result to MLPs with $\geq 2$ layers.
We can actually prove that $\mu$P-trained MLPs with $\geq 2$ layers, $W_1(\mu_n,\mu_{2n};d_\infty^{\op})\not\to 0$ at initialization.

For simplicity, consider randomly initialized $n$-width and $2n$-width MLPs with $\geq 2$ layers, and let $\mu_n$ and $\mu_{2n}$ denote the corresponding weight distributions.
Focus on one hidden-to-hidden layer, whose width-$n$ and width-$2n$ matrices are respectively $n\times n$ and $2n\times 2n$.
We first map the width-$n$ weights to get an equivalent width-$2n$ model: $W_n\mapsto W_n\otimes \frac12\one_2\one_2^\top$.
The resultant matrix has rank at most $n$, so its $(n+1)$-th singular value is zero.

The width-$2n$ matrix $W_{2n}$ has i.i.d.\ entries with variance $1/(2n)$ at initialization.
Standard random matrix theory results show that the empirical distribution of its squared singular values converges to the Marchenko--Pastur distribution supported on $[0,4]$.
Choose $c>0$ small enough so that this limiting distribution assigns mass greater than $1/2$ to $(c^2,\infty)$.
Then, with high probability, more than $n$ singular values of the $2n\times 2n$ matrix exceed $c$.
Equivalently, the $(n+1)$-st singular value is bounded below by $c$ with high probability.
Finally, Weyl's inequality gives, for every hidden-unit permutation $g$ at width $2n$,
\[
\left\|W_{2n}-g\cdot\left(W_n\otimes \tfrac12\one_2\one_2^\top\right)\right\|
\ge
s_{n+1}\left(W_{2n}\right)-0
\ge c>0
\]
on this high-probability event.
This lower bound is independent of the coupling of the two initialized networks, because it depends only on the width-$2n$ marginal.
Hence, for every coupling $\pi$ of $\mu_n$ and $\mu_{2n}$,
\[
\Ex_\pi d_\infty^{\op}(\theta_n,\theta_{2n})
\ge
c\mathbb{P}\left(s_{n+1}(W_{2n})\ge c\right).
\]
Taking the infimum over couplings gives
\[
\liminf_{n\to\infty}W_1(\mu_n,\mu_{2n};d_\infty^{\op})
\ge
c>0,
\]
so the convergence in~\eqref{eq:normalized-spectral-norm-convergence} fails at initialization.

For trained weights at fixed positive time, it is unknown whether the same obstruction persists.
On the one hand, empirically it is observed that sometimes, the trained weights have a low-rank structure, which may remove the obstruction;
On the other hand, the existing $\mu$P theory does not seem to give any guarantee of such closeness.

\subsection{Proof of \Cref{thm:guarantee-2}}
\label{app:guarantee-2}
The previous subsection compared independently trained weights through the distance $d_\infty^{\op}$.
That strategy works for 1-layer MLPs under $\mu$P, where trained weights of different widths are provably close in $d_\infty^{\op}$, but it breaks down as soon as the input MLP has a hidden-to-hidden layer: by \Cref{sec:obstruction-larger-depths}, weights of different widths stay separated in $d_\infty^{\op}$ already at initialization.

This does not mean that size generalization itself fails at depth $\ge2$; it only means that this particular route to it is unavailable.
The reason is that $d_\infty^{\op}$ is not a faithful measure of the distance that the GMN class prescribes.
We proved that the matrix-product GMNs (or duplication-compatible GMNs in the 1-layer MLP case) are Lipschitz with respect to $d_\infty^{\op}$, so that closeness in $d_\infty^{\op}$ implies closeness of the GMN outputs.
But the converse is not true: weights can produce close GMN outputs while remaining far apart in $d_\infty^{\op}$.
Consequently $W_1(\mu_n,\mu_m;d_\infty^{\op})\to0$ is a sufficient but not a necessary condition for the size-generalization gap $\mathcal{R}_n(f^\Theta)-\mathcal{R}_m(f^\Theta)$ to vanish.

This suggests measuring the two weight distributions in a distance generated by the GMN class itself.
Concretely, let $\mathcal A\subset\RR^q$ be a compact parameter set for the matrix-product GMN, and for $\Theta\in\mathcal A$ write $f_n^\Theta$ for its width-$n$ instantiation.
We consider the following distance between the trained weight distributions $\mu_n$ and $\mu_m$:
\begin{equation*}
d_{\mathcal{A}}(\mu_n, \mu_m)
\colonequals
\sup_{\Theta\in\mathcal{A}}
\abs{
\Ex_{\theta\sim \mu_n} \ell(f_n^\Theta(\theta),f_n^*(\theta))
-
\Ex_{\theta'\sim \mu_m} \ell(f_m^\Theta(\theta'),f_m^*(\theta'))
} .
\end{equation*}
This is the \emph{integral probability metric} (IPM) generated by the width-indexed test class of composed maps $\theta\mapsto\ell(f_n^\Theta(\theta),f_n^*(\theta))$, $\Theta\in\mathcal A$.
IPMs are a standard way of comparing probability measures, taking the supremum of $\abs{\Ex_\mu h-\Ex_\nu h}$ over a class of test functions $h$. 
Total variation, the Wasserstein-$1$ distance, and maximum mean discrepancy are examples of IPM with different choices of the test class~\citep{muller1997integral}.
It defines a pseudometric on the measures for which the integrals exist.

By construction $d_{\mathcal A}$ dominates the size-generalization gap, since $\abs{\mc R_n(f^\Theta)-\mc R_m(f^\Theta)}\le d_{\mathcal A}(\mu_n,\mu_m)$ for every fixed $\Theta\in\mathcal A$.
The goal of this subsection is therefore to show that $d_{\mathcal A}(\mu_n,\mu_m)\to0$ as $n,m\to\infty$.

\paragraph{High-level idea of the proof.}
In~\Cref{sec:mup-preliminaries}, we reviewed that the training of MLP under $\mu$P of fixed number of training steps can be written with a tensor program, with the infinite-width limit precisely described. 
The key observation we use here is that, evaluating the matrix-product GMN on those weights does not leave this class.
For a fixed parameter $\Theta$, the only way $f^\Theta$ touches a width-$n$ weight matrix is by multiplying it, or its transpose, against a vector assembled from node features with fixed-dimensional coefficient matrices.
Every such operation is a legal tensor-program instruction.
Appending these instructions, message-passing round by message-passing round, to the program that produced $\theta_n$ therefore realizes the computation $f_n^\Theta(\theta_n)$ inside a single tensor program.
Assuming that the target summary $f_n^*$ and the loss $\ell$ are likewise expressible as tensor-program operations, the scalar $\ell(f_n^\Theta(\theta_n),f_n^*(\theta_n))$ is an output of one finite tensor program.
The tensor program machinery then gives a deterministic limit for this scalar as $n\to\infty$.

The remaining steps are technical.
We upgrade this convergence to that of the expected risk $\Ex_{\theta_n\sim \mu_n}[\ell(f_n^\Theta(\theta_n),f_n^*(\theta_n))]$, and furthermore, to uniform convergence over $\Theta\in\mathcal A$.
We now carry out this proof technique in detail.

\paragraph{Evaluation of matrix-product GMN on trained weights as a tensor program.}
We first spell out why evaluating a matrix-product GMN (with fixed parameter $\Theta$) on the trained weights can be appended to the same tensor program that produced those weights.

Following the original tensor program convention, we assume the MLP has homogeneous hidden-layer widths ($n_1=\cdots=n_L=n$).
This is purely for simplified notation, and extending to heterogeneous widths is straightforward.
Throughout, the GMN parameter $\Theta$ is fixed, so its entries can be hardcoded into the \textsc{OuterNonlin} instructions, and no separate constant inputs are needed.

We append the GMN evaluation by induction over message-passing rounds.

\emph{Initialization of node features at $t=0$.}
Each hidden-layer node feature coordinate
$H^{(\ell),[0]}_{:,a} \in \RR^n$ for $1\le \ell\le L,\ 1\le a\le d_h$
is produced as a program vector by a single \textsc{OuterNonlin} instruction with $r=0$, applying entrywise to the trained bias $b^{(\ell)}(T)\in\RR^n$, which is an existing program vector.
Indeed, for the initialization $h^{(\ell),[0]}_i=\psi_v\left(b^{(\ell)}_i\right)$ of the simple GMN, the instruction is the one whose nonlinearity $\bm x_j\mapsto(\psi_v)_a\left(\left(b^{(\ell)}(T)\right)_j\right)$ reads off the coordinate of the row $\bm x_j$ holding $b^{(\ell)}(T)$, applies the $a$th output coordinate of $\psi_v$ to it, and ignores all other arguments.
Since $\psi_v$ is an MLP, it is Lipschitz, hence pseudo-Lipschitz, and the extended program still satisfies \Cref{ass:tp-setup}.

The input node features $H^{(0),[0]} \in \RR^{\din\times d_h}$ form a constant matrix.
The output node features $H^{(L+1),[0]}\in\RR^{\dout\times d_h}$ are fixed-dimensional; their entries are produced as program scalars from the existing program scalars $b_i^{(L+1)}(T)$, $i\in[\dout]$, using \textsc{OuterNonlin} and \textsc{Avg}.

\emph{Message-passing round $t\ge0$.}
Assume that the columns of $H^{(\ell),[t]}$ have already been produced as program vectors for every hidden layer $\ell\in[1,L]$, and the entries of $H^{(0),[t]}$ and $H^{(L+1),[t]}$ have been produced as program scalars.

We first produce, for $\ell\in[1,L+1], 1\le a\le d_h$,
\[
U^{(\ell),\mathrm{in},[t]}_{:,a}
=
\sum_{b=1}^{d_h} H^{(\ell-1),[t]}_{:,b}(A^{\mathrm{in}}_\ell)_{ba},
\]
which is a program vector in $\RR^n$ for $\ell\neq 1$, and a set of program scalars for $\ell=1$.
For $\ell\ge2$, this is produced by a \textsc{OuterNonlin} instruction on existing program vectors $H^{(\ell-1),[t]}_{:,b}$, with the nonlinearity $\bm x_j\mapsto\sum_{b=1}^{d_h} (\bm x_j)_b(A^{\mathrm{in}}_\ell)_{ba}$.
The scalar case $\ell=1$ is the same finite linear combination at the level of program scalars.
The entries $(A^{\mathrm{in}}_\ell)_{ba}$ are fixed numbers that can be hardcoded into the \textsc{OuterNonlin} instruction.

We then produce the aggregate of incoming messages at each node.
When $2\le\ell\le L$, we produce a program vector
\[
Z^{(\ell),\mathrm{in},[t]}_{:,a}
=
W^{(\ell)}(T)U^{(\ell),\mathrm{in},[t]}_{:,a} \in \RR^n,
\]
for $a\in [1,d_h]$, by applying \textsc{MatMul} to the initial i.i.d.\ matrix part and using \textsc{OuterNonlin} and \textsc{Avg} for the finite outer-product terms representing the gradient updates during training.
When $\ell=1$, produce
\[
Z^{(1),\mathrm{in},[t]}_{:,a}
=
\sum_{j=1}^{\din} W^{(1)}_{:,j}(T)U^{(1),\mathrm{in},[t]}_{j,a},
\]
for $a\in [1,d_h]$, which is a \textsc{OuterNonlin} instruction applied to existing program vectors $W^{(1)}_{:,j}(T)$ and program scalars $U^{(1),\mathrm{in},[t]}_{j,a}$.
When $\ell=L+1$, first produce a program vector
\[
Y^{(L+1),\mathrm{in},[t]}_{i,:,a}
=
\overline W^{(L+1)}_{i,:}(T)\odot U^{(L+1),\mathrm{in},[t]}_{:,a},
\]
for each $i\in[\dout]$ and $a\in[d_h]$,
by \textsc{OuterNonlin} with $r=0$, using that $\overline W^{(L+1)}_{i,:}(T)$ and $U^{(L+1),\mathrm{in},[t]}_{:,a}$ are existing program vectors.
We highlight here that the $\mu$P scaling ensures that the rescaled weight $\overline W^{(L+1)}_{i,:}(T)$ has order-one entries throughout training and hence can be appended to the tensor program as a program vector.
Then, apply \textsc{Avg} to obtain
\[
Z^{(L+1),\mathrm{in},[t]}_{i,a}
=
\frac1n\sum_{\alpha=1}^n
\overline W^{(L+1)}_{i\alpha}(T)U^{(L+1),\mathrm{in},[t]}_{\alpha,a}
=
\sum_{\alpha=1}^n
W^{(L+1)}_{i\alpha}(T)U^{(L+1),\mathrm{in},[t]}_{\alpha,a}.
\]

The outgoing aggregation to form $Z^{(\ell),\mathrm{out},[t]}$ can be performed in the same way, so we leave it out for brevity.

Finally, the node update is row-wise.
For each $\ell\in[1,L], 1\le a\le d_h$, produce
\[
H^{(\ell),[t+1]}_{:,a}
=
\psi_{h,a}\left(
H^{(\ell),[t]},
Z^{(\ell),\mathrm{in},[t]},
Z^{(\ell),\mathrm{out},[t]}
\right),
\]
by \textsc{OuterNonlin} with $r=0$ applied to existing program vectors $H^{(\ell),[t]}_{:,b}$, $Z^{(\ell),\mathrm{in},[t]}_{:,b}$, and $Z^{(\ell),\mathrm{out},[t]}_{:,b}$ for $b\in[d_h]$.
$\psi_h$ is an MLP, hence pseudo-Lipschitz, so the extended program still satisfies \Cref{ass:tp-setup}.
The node update for $\ell=0$ and $\ell=L+1$ is similarly produced as program scalars by \textsc{OuterNonlin} and \textsc{Avg}.

This completes the induction for one message-passing round, and repeating it for the fixed number of rounds gives all final node features.

\emph{Graph-level readout.}
For each $\ell\in[1,L]$ and $1\le a\le d_h$, produce a program vector
\[
R^{(\ell)}_{:,a}
=
\psi_{r,a}\left(H^{(\ell),[T]}\right)\in \RR^n,
\]
by \textsc{OuterNonlin} with $r=0$, and then apply \textsc{Avg} to obtain
\[
r^{(\ell)}_a
=
\frac1n\sum_{\alpha=1}^n R^{(\ell)}_{\alpha,a}.
\]
The case $\ell=0$ and $\ell=L+1$ is similarly produced as program scalars by \textsc{OuterNonlin} and \textsc{Avg}.

The final readout is produced by applying $\rho$ to the program scalars $r^{(\ell)}_a$ for $\ell\in[0,L+1]$ and $a\in[d_h]$.
$\rho$ is an MLP and hence again pseudo-Lipschitz, so the extended program still satisfies \Cref{ass:tp-setup}.

\paragraph{Evaluation of the target and loss as a tensor program.}
Next, we want to further append the evaluation of the loss $\ell(f_n^\Theta(\theta_n),f_n^*(\theta_n))$ to the tensor program.
With the above, this will ensure the full computation ``train the input network, evaluate the matrix-product GMN, evaluate the target, and evaluate the loss'' is altogether a finite tensor program.

This requires the following assumption on the target functions $(f_n^*)_n$ and the loss $\ell$.
\begin{assumption}
\label{ass:mup-matprod-gmn-risk}
\begin{enumerate}[label=(\roman*),leftmargin=*,topsep=0pt,itemsep=0pt]
\item For each width $n$, the target function $f_n^*:V_n\to \RR^k$ is deterministic and its computation can be expressed with finitely many instructions of \textsc{Avg}, \textsc{OuterNonlin}, and \textsc{MatMul}, on the following inputs:
\begin{itemize}[leftmargin=*,topsep=0pt,itemsep=0pt]
    \item Matrix inputs: the trained hidden-to-hidden matrices $W^{(\ell)}(T)\in \RR^{n\times n}$ for $2\le\ell\le L$.
    \item Vector inputs:  the columns of input weights $W^{(1)}(T)\in \RR^{n\times \din}$, and the rows of scaled output weights  $\overline W^{(L+1)}(T)\in \RR^{\dout\times n}$, and the trained hidden biases $b^{(\ell)}(T) \in \RR^{n}$ for $1\le\ell\le L$.
    \item Scalar inputs: the trained output bias $b^{(L+1)}(T)\in \RR^{\dout}$, and any other fixed, width-independent scalars.
\end{itemize}
Every nonlinearity used in a \textsc{OuterNonlin} instruction in this computation is assumed to be pseudo-Lipschitz in the sense of \Cref{ass:tp-setup}(d), and fixed independently of the width.

\item The loss $\ell\colon\RR^k\times\RR^k\to\RR$ is Lipschitz.
\item For some $q>1$, the loss scalars have a width-uniform $L^q$ bound,
\[
\sup_{\Theta\in\mathcal A}
\sup_{(\gamma,s)\in\operatorname{supp}(\Gamma,S)}
\sup_{n\ge1}
\Ex_{\xi_n}
\abs{\ell\left(f_n^\Theta(\theta_n),f_n^*(\theta_n)\right)}^q
<\infty.
\]
\end{enumerate}
\end{assumption}

\Cref{ass:mup-matprod-gmn-risk}(i) is stated in this form so that it is clear that evaluating the target function can be appended to the tensor program.
It is less clear which target functions $(f_n^*)_n$ this condition actually includes.
We discuss when and when not it is satisfied.
\begin{rmk}[Satisfaction of the target function assumption]
    A convenient sufficient condition is \emph{exact uniform representability} of the target family $(f_n^*)_n$ by the matrix-product GMN.
    In that case, the evaluation of $f_n^*$ can be replaced by the evaluation of the matrix-product GMN that represents it at width $n$, which we proved above is a tensor program.
    With some additional work, the result might extend to target families that can be \emph{uniformly approximated} by the matrix-product GMN across widths.
    The expressivity of the matrix-product GMN on MLPs of different widths is not yet well understood, and remains an important open question.

    The assumption is also satisfied by targets obtained from finitely many evaluations of the input MLP itself.
    For example, fix width-independent inputs $u_1,\dots,u_m\in\RR^{\din}$ and let
    \[
    f_n^*(\theta)
    =
    \Phi\left(\mc{R}(\theta)(u_1),\dots,\mc{R}(\theta)(u_m)\right),
    \]
    where $\Phi\colon(\RR^{\dout})^m\to\RR^k$ is a fixed Lipschitz function.
    This covers, for instance, finite-validation-set generalization prediction for a Lipschitz objective.

    By contrast, the assumption does not cover an arbitrary functional $F^*\colon\mc{C}(X,\RR^{\dout})\to\RR^k$ applied to the realized function $\mc{R}(\theta)$.
    Such a functional may depend on the values of $\mc{R}(\theta)$ on infinitely many inputs, or on features that are not reducible to finitely many evaluations and width-independent nonlinearities.
    This is the main difference from the other size generalization result in~\Cref{cor:sizegen-one-layer}, which only assumes that the function-space functional is Lipschitz with respect to the uniform norm.
\end{rmk}

Finally, we give the proof of the main result.

We first give a technical lemma that upgrades pointwise convergence to uniform convergence on compact sets.
Say that a sequence of functions $(g_n)$ is \emph{equicontinuous} if for every $\varepsilon>0$ there exists $\delta>0$ such that $d(x,y)<\delta$ implies $\abs{g_n(x)-g_n(y)}<\varepsilon$ for all $n$.
\begin{lemma}[Equicontinuous pointwise convergence on compact sets]
\label{lem:equicontinuous-pointwise-uniform}
Let $(K,d)$ be a compact metric space, and let $g_n\colon K\to\RR$ be a sequence of functions.
If $(g_n)$ is equicontinuous and $g_n(x)\to g(x)$ for every $x\in K$, then $g$ is continuous and $g_n\to g$ uniformly on $K$.
\end{lemma}

\begin{proof}
Let $\varepsilon>0$.
By equicontinuity, choose $\delta>0$ such that $d(x,y)<\delta$ implies
\[
\abs{g_n(x)-g_n(y)}
<
\varepsilon/3
\qquad\text{for all }n.
\]
Taking $n\to\infty$ in this inequality gives
\[
\abs{g(x)-g(y)}
\le
\varepsilon/3
\qquad\text{whenever }d(x,y)<\delta.
\]
This proves continuity of $g$.
Since $K$ is compact, choose a finite $\delta$-net $\{x_1,\dots,x_R\}$ of $K$.
For each $r$, pointwise convergence gives $N_r$ such that
\[
\abs{g_n(x_r)-g(x_r)}
<
\varepsilon/3
\qquad\text{for all }n\ge N_r.
\]
Let $N\colonequals\max_{1\le r\le R}N_r$.
For any $n\ge N$ and any $x\in K$, choose $r$ with $d(x,x_r)<\delta$.
The triangle inequality gives
\[
\abs{g_n(x)-g(x)}
\le
\abs{g_n(x)-g_n(x_r)}
+
\abs{g_n(x_r)-g(x_r)}
+
\abs{g(x_r)-g(x)}
<
\varepsilon.
\]
Taking the supremum over $x\in K$ proves uniform convergence.
\end{proof}

\begin{proposition}
\label{thm:mup-modified-gmn-ipm}
Assume \Cref{ass:mup-matprod-gmn-risk}, and that the matrix-product GMN satisfies the hypotheses of \Cref{lem:matprod-param-sensitivity}.
Assume also that the trained weights have a width-uniform moment of order $T+1$ in the normalized spectral norm, $T$ being the number of message-passing rounds:
\begin{equation}
\label{eq:mup-matprod-weight-moment}
\sup_{n\ge1}
\Ex_{\Gamma,S}\Ex_{\xi_n}
\norm[V_n,\op]{\theta_n}^{T+1}
<\infty.
\end{equation}
Then
\[
\sup_{m\ge n}d_{\mathcal A}(\mu_n,\mu_m)
\to 0
\qquad\text{as }n\to\infty.
\]
\end{proposition}

\begin{proof}
We prove convergence through a common infinite-width limit.

\emph{Almost sure convergence of loss scalars.}
We first fix $(\gamma,s)$ in the support of $(\Gamma,S)$ and condition on $(\Gamma,S)=(\gamma,s)$.
This fixes the hyperparameters $\gamma$ and the auxiliary training randomness $s$, so that the initialization is the only remaining source of randomness.
Throughout the conditional argument, $\theta_n$ denotes $\mathsf{Train}_n(\gamma,s,\xi_n)$.

Fix $\Theta\in\mathcal A$.
We have shown that the evaluation of the matrix-product GMN on the trained weights can be appended to the tensor program that produced those weights; furthermore, under \Cref{ass:mup-matprod-gmn-risk}, the evaluation of the target function on the trained weights, and the computation of the loss can all be appended to the same tensor program.
Consequently, we have a finite tensor program where $\ell\left(f_n^\Theta(\theta_n),f_n^*(\theta_n)\right)$ is a scalar produced by it.
The master theorem~\Cref{lem:tp-master} therefore gives a deterministic conditional limit $L_\infty(\Theta;\gamma,s)$ such that
\[
\ell\left(f_n^\Theta(\theta_n),f_n^*(\theta_n)\right)
\xrightarrow[n\to\infty]{\mathrm{a.s.}}
L_\infty(\Theta;\gamma,s).
\]

\emph{Convergence of expected risks.}
By assumption \Cref{ass:mup-matprod-gmn-risk}(iii), the $q$th absolute moment satisfies
\[
\Ex_{\xi_n}
\abs{\ell\left(f_n^\Theta(\theta_n),f_n^*(\theta_n)\right)}^q
< C
\]
for some constant $C$ independent of $(\gamma,s)$ and uniformly in $n$.
Since $q>1$, the de la Vallée--Poussin criterion shows that the family
$\{\ell\left(f_n^\Theta(\theta_n),f_n^*(\theta_n)\right):n\ge1\}$
is uniformly integrable with respect to $\xi_n$.
The almost-sure convergence above and the Vitali convergence theorem therefore give
\[
\Ex_{\xi_n}\ell\left(f_n^\Theta(\theta_n),f_n^*(\theta_n)\right)
\longrightarrow
L_\infty(\Theta;\gamma,s).
\]
We next pass from the conditional expectation to the expectation over $(\Gamma,S)$.
Jensen's inequality applied to the concave map $t\mapsto t^{1/q}$, gives
\[
\abs{\Ex_{\xi_n}\ell\left(f_n^\Theta(\theta_n),f_n^*(\theta_n)\right)}
\le
\Ex_{\xi_n}\abs{\ell\left(f_n^\Theta(\theta_n),f_n^*(\theta_n)\right)}
\le
\left(\Ex_{\xi_n}\abs{\ell\left(f_n^\Theta(\theta_n),f_n^*(\theta_n)\right)}^q\right)^{1/q}
\le
C^{1/q},
\]
a bound independent of $(\gamma,s)$ and $n$.
The bounded convergence theorem over $(\Gamma,S)$ yields
\[
a_n(\Theta)
\colonequals
\Ex_{\Gamma,S}\Ex_{\xi_n}\ell\left(f_n^\Theta(\theta_n),f_n^*(\theta_n)\right)
\longrightarrow
a_\infty(\Theta)
\colonequals
\Ex_{\Gamma,S}L_\infty(\Theta;\Gamma,S).
\]

\emph{Uniformity over $\Theta$.}
We first prove that the family $\{a_n:n\ge1\}$ is Lipschitz on $\mathcal A$ with a common Lipschitz constant.

By \Cref{lem:matprod-param-sensitivity} and the Lipschitzness of $\ell$, for all $\Theta,\widetilde\Theta\in\mathcal A$,
\[
\left|
\ell\left(f_n^\Theta(\theta_n),f_n^*(\theta_n)\right)
-
\ell\left(f_n^{\widetilde\Theta}(\theta_n),f_n^*(\theta_n)\right)
\right|
\le
M_n\norm[2]{\Theta-\widetilde\Theta},
\]
where $M_n\colonequals\Lambda c_{\mathcal A,T}\left(1+\norm[V_n,\op]{\theta_n}\right)^{T+1}$, $\Lambda$ is the Lipschitz constant for the loss in \Cref{ass:mup-matprod-gmn-risk}(ii), and $c_{\mathcal A,T}$ is the constant from \Cref{lem:matprod-param-sensitivity}.

Since $(1+x)^{T+1}\le2^{T+1}(1+x^{T+1})$ for $x\ge0$, the assumed moment bound \eqref{eq:mup-matprod-weight-moment} gives
\[
\sup_{n\ge1}
\Ex_{\Gamma,S}\Ex_{\xi_n}
M_n
\le
2^{T+1}\Lambda c_{\mathcal A,T}
\left(
1+\sup_{n\ge1}\Ex_{\Gamma,S}\Ex_{\xi_n}\norm[V_n,\op]{\theta_n}^{T+1}
\right)
\colonequals
C_{\mathcal A}<\infty.
\]
Taking expectations in the bound above gives, for all $\Theta,\widetilde\Theta\in\mathcal A$,
\[
\abs{a_n(\Theta)-a_n(\widetilde\Theta)}
\le
C_{\mathcal A}\norm[2]{\Theta-\widetilde\Theta}
\qquad\text{for all }n.
\]
By the pointwise convergence already proved, the same Lipschitz bound holds for $a_\infty$ after taking $n\to\infty$.

Since $\mathcal A$ is compact, the common Lipschitz bound and \Cref{lem:equicontinuous-pointwise-uniform} upgrade the pointwise convergence $a_n(\Theta)\to a_\infty(\Theta)$ to
\[
\sup_{\Theta\in\mathcal A}
\abs{a_n(\Theta)-a_\infty(\Theta)}
\longrightarrow 0.
\]
Finally,
\begin{align*}
\sup_{m\ge n}d_{\mathcal A}(\mu_n,\mu_m)
&=
\sup_{m\ge n}\sup_{\Theta\in\mathcal A}\abs{a_n(\Theta)-a_m(\Theta)}\\
&\le
\sup_{\Theta\in\mathcal A}\abs{a_n(\Theta)-a_\infty(\Theta)}
+
\sup_{m\ge n}\sup_{\Theta\in\mathcal A}\abs{a_m(\Theta)-a_\infty(\Theta)}
\longrightarrow0.
\end{align*}
\end{proof}

\begin{corollary}[Size generalization for matrix-product GMNs under $\mu$P]
\label{cor:sizegen-matprod-mup}
Assume \Cref{ass:mup-matprod-gmn-risk} and the remaining hypotheses of \Cref{thm:mup-modified-gmn-ipm}.
For each fixed $\Theta\in\mathcal A$, define
\[
\mc R_n(f^\Theta)
\colonequals
\Ex_{\theta\sim\mu_n}
\ell\left(f_n^\Theta(\theta),f_n^*(\theta)\right).
\]
Then
\[
\sup_{m\ge n}\abs{\mc R_n(f^\Theta)-\mc R_m(f^\Theta)}
\longrightarrow0
\qquad\text{as }n\to\infty .
\]
\end{corollary}

\begin{proof}
For every $m\ge n$ and every fixed $\Theta\in\mathcal A$,
\[
\abs{\mc R_n(f^\Theta)-\mc R_m(f^\Theta)}
\le
d_{\mathcal A}(\mu_n,\mu_m).
\]
Taking the supremum over $m\ge n$ and applying \Cref{thm:mup-modified-gmn-ipm} proves the claim.
\end{proof}

\end{document}